\documentclass[12pt,final,twoside]{report}
\newcommand{\trtitle}{Benchmarking Faithfulness:\\ Towards Accurate Natural Language Explanations in Vision-Language Tasks}
\newcommand{\trtype}{Master Thesis} 
\newcommand{\trcourseofstudies}{Informatik} 
\newcommand{\trauthor}{Jakob Ambsdorf}
\newcommand{\trauthortitle}{} 
\newcommand{\tremail}{jakob.ambsdorf@gmail.com}
\newcommand{\trmatrikelnummer}{6919840}
\newcommand{\trstrasse}{Koppel 14}
\newcommand{\trort}{20099 Hamburg}
\newcommand{\trgutachterA}{\href{mailto:wermter@informatik.uni-hamburg.de}{Prof. Dr. Stefan Wermter}}
\newcommand{\trgutachterB}{\href{mailto:lee@informatik.uni-hamburg.de}{Dr. Jae Hee Lee}}

\newcommand{\trfach}{Knowledge Technology, WTM}
\newcommand{\trdate}{28.08.2022}
\newcommand{\trkeywords}{Learning, Language, Demonstration}

\usepackage[english]{babel}                         
\usepackage{acronym}                    
\usepackage{algpseudocode}                
\usepackage{algorithm}                  
\usepackage{amsfonts}                   
\usepackage{amsmath}                    
\usepackage{amssymb}                    
\usepackage{amsthm}
\usepackage{adjustbox}
\usepackage{booktabs}                   
\usepackage{color}                      
\definecolor{uhhRed}{RGB}{226,0,26}     
\definecolor{uhhGrey}{RGB}{136,136,136} 
\definecolor{uhhLightGrey}{RGB}{220, 220, 220}
\usepackage{fancybox}                   
\usepackage{fancyhdr}                   
\usepackage{float}
\usepackage[body={5.8in,9in}]{geometry} 

\usepackage{graphicx}                   
\usepackage{caption}
\usepackage{subcaption}
\usepackage{longtable} 
\usepackage{tabulary}
\usepackage{listings}                   
\usepackage{multicol}                   
\usepackage{multirow}                   
\usepackage{rotating}                   
\usepackage{tabularx}                   
\usepackage{url,xspace,boxedminipage}   
\usepackage{varwidth}
\usepackage{centernot}

\author{\trauthor}

\ifx\pdftexversion\undefined
\usepackage{hyperref}
\else
\usepackage[colorlinks=false,           
            linkcolor=uhhRed,           
            urlcolor=uhhRed,
            citecolor=uhhRed,
            bookmarks,                  
            bookmarksopen=true,         
            pdfpagemode=UseOutlines,    
            bookmarksopenlevel=1,       
            bookmarksnumbered,          
            pdftitle={\trtitle},
            pdfsubject={\trtype},
            pdfkeywords={\trkeywords},
            pdfauthor={\trauthor},
            plainpages=false
            ]{hyperref}
\fi

\ifx\pdftexversion\undefined
\else
\pdfoutput=1                            
\fi

\DeclareGraphicsExtensions{.pdf,.svg,.jpg,.png,.eps} 
\graphicspath{{./src/}}                 
\theoremstyle{plain}

\theoremstyle{definition}
\newtheorem{definition}{Definition}[chapter]

\theoremstyle{definition}
\newtheorem{assumption}{Assumption}[chapter]

\theoremstyle{definition}
\newtheorem{conjecture}{Conjecture}[chapter]

\theoremstyle{definition}
\newtheorem{implication}{Implication}[chapter]

\theoremstyle{remark}

\begin{document}

\pagenumbering{Roman}                   
\renewcommand{\headheight}{14.5pt}      

\thispagestyle{empty}
\fancyhead[LO,RE]{}                     

\begin{titlepage}
    \begin{flushleft}
        \includegraphics[width=85mm]{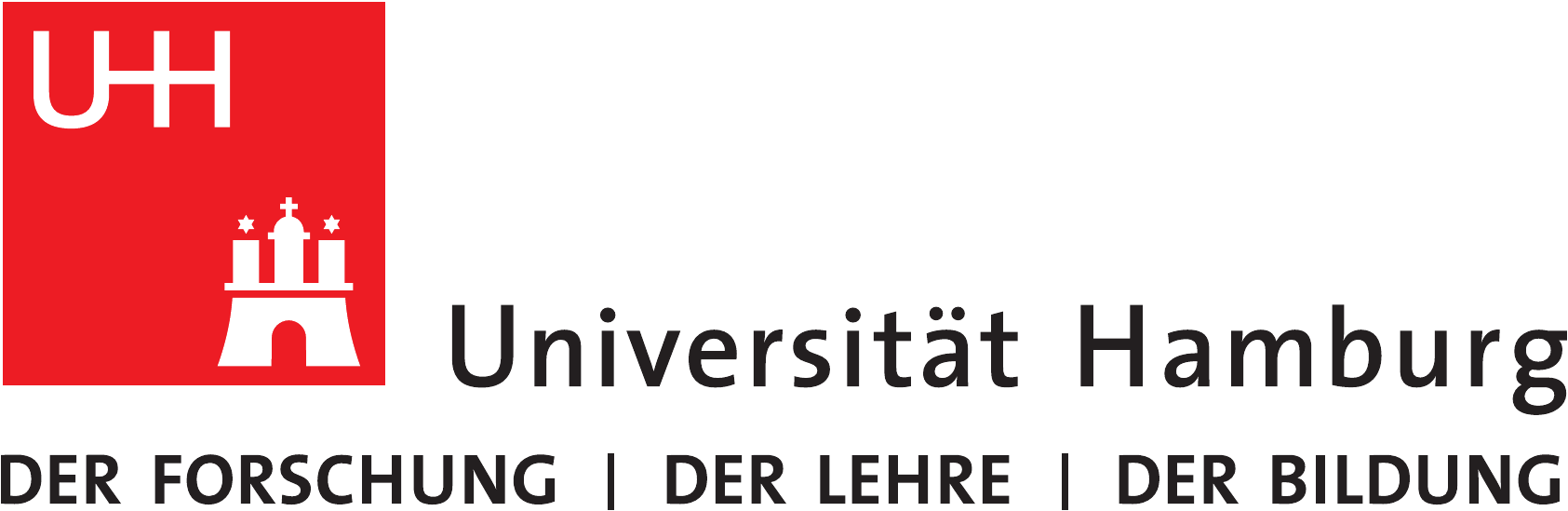}\\
    \end{flushleft}
    \rule{\textwidth}{0.4pt}
        \newline
        \vspace{2.0cm}
        \begin{center}
          \LARGE \textbf{\trtitle}
        \end{center}
    \vspace{2.0cm}
    \begin{center}
      \textbf{\trtype}\\
      im Arbeitsbereich \trfach\\
      \trgutachterA\medskip\\
      Department Informatik\\
      MIN-Fakult\"at\\
      Universit\"at Hamburg \\[0.5cm]
      vorgelegt von \\
      \textbf{\trauthortitle\href{mailto:\tremail}{\trauthor}}\\
      am\\
      \trdate
    \end{center}
    \vspace{1cm}
    \begin{center}
    \begin{tabular}{ll}
    Gutachter: & \trgutachterA \\
                   & \trgutachterB \\
    \end{tabular}
    \end{center}
    \vfill
    \begin{tabular}{l}
    \trauthor \\
    Matrikelnummer:  \trmatrikelnummer \\
    \trstrasse \\
    \trort
    \end{tabular}
    \newline
    \rule{\textwidth}{0.4pt}
    \newpage 
\end{titlepage}

\thispagestyle{empty}
\hspace{1cm}
\newpage

\section*{Abstract}\label{sec:abstract}
With deep neural models increasingly permeating our daily lives comes a need for transparent and comprehensible explanations of their decision-making. However, most explanation methods that have been developed so far are not intuitively understandable for lay users. In contrast, natural language explanations (NLEs) promise to enable the communication of a model's decision-making in an easily intelligible way. While current models successfully generate convincing explanations, it is an open question how well the NLEs actually represent the reasoning process of the models -- a property called \textit{faithfulness}. Although the development of metrics to measure faithfulness is crucial to designing more faithful models, current metrics are either not applicable to NLEs or are not designed to compare different model architectures across multiple modalities. 

Building on prior research on faithfulness measures and based on a detailed rationale, we address this issue by proposing three faithfulness metrics: \textit{Attribution-Similarity}, \textit{NLE-Sufficiency}, and \textit{NLE-Comprehensiveness}. The efficacy of the metrics is evaluated on the VQA-X and e-SNLI-VE datasets of the e-ViL benchmark for vision-language task NLE generation~\cite{kayser2021vil} by systematically applying modifications to the performant e-UG model for which we expect changes in the measured explanation faithfulness. We show on the e-SNLI-VE dataset that the removal of redundant inputs to the explanation-generation module of e-UG successively increases the model's faithfulness on the linguistic modality as measured by \textit{Attribution-Similarity}. We discuss reasons why the expected increase in faithfulness was not measured on the visual modality of the e-SNLI-VE dataset or on both modalities of the VQA-X dataset and derive conclusions for designing more faithful NLE models going forward. Further, our analysis demonstrates that \textit{NLE-Sufficiency} and \textit{-Comprehensiveness} are not necessarily correlated to \textit{Attribution-Similarity}, and we discuss how the two metrics can be utilized to gain further insights into the explanation generation process.

\vfill
\section*{Zusammenfassung}\label{sec:zusammenfassung}

Mit der zunehmenden Bedeutung von Deep Learning Systemen f\"ur unser t\"agliches Leben besteht ein Bedarf an transparenten und verständlichen Erklärungen für ihre Entscheidungsfindung. Die meisten bisher entwickelten Ans\"atze zur Erkl\"arung der Modelle sind jedoch für Laien nicht intuitiv verständlich. Im Gegensatz dazu, versprechen natürlichsprachliche Erklärungen (Natural Language Explanations - NLEs), die Entscheidungsfindung eines Modells auf leicht verständliche Weise zu vermitteln. Während aktuelle Modelle bereits erfolgreich überzeugende Erklärungen generieren, ist es eine offene Frage, wie gut die NLEs den Entscheidungsprozess dieser Modelle tatsächlich abbilden -- eine Eigenschaft, die als \textit{Faithfulness} bezeichnet wird. Obwohl die Entwicklung von Methoden zur Messung der Faithfulness entscheidend für das Design von diesbez\"uglich besseren Modellen ist, sind die derzeitigen Messmethoden entweder nicht auf NLEs anwendbar, oder nicht für den Vergleich verschiedener Modellarchitekturen über mehrere Modalitäten hinweg ausgelegt. 

Aufbauend auf vorherigen Forschungen zur Messung der Faithfulness, gehen wir dieses Problem an, indem wir drei Metriken zur Messung der Faithfulness vorschlagen: \textit{Attribution-Similarity}, \textit{NLE-Sufficiency} und \textit{NLE-Comprehensiveness}. Die Wirksamkeit der Metriken wird an den VQA-X und e-SNLI-VE Datensätzen des e-ViL Benchmarks für die NLE-Generierung in visuell-sprachlichen Aufgaben~\cite{kayser2021vil} durch systematische Modifikationen des leistungsfähigen e-UG Modells evaluiert, f\"ur welceh wir bestimmte Änderungen der Faithfulness erwarten. Wir zeigen anhand des e-SNLI-VE Datensatzes, dass die Entfernung redundanter Eingaben in das Erklärungsmodul von e-UG die Faithfulness in Bezug auf die linguistische Modalität erhöht, gemessen durch \textit{Attribution-Similarity}. Wir erörtern die m\"oglichen Gründe, warum wider Erwartens eine Erhöhung der Faithfulness auf der visuellen Modalität im e-SNLI-VE Datensatz, sowie auf beiden Modalitäten im VQA-X Datensatz nicht gemessen wurde und leiten daraus Erkenntnisse über die Gestaltung von zuk\"unftigen NLE-Modellen ab, die eine h\"ohere Faithfulness aufweisen sollen.
Darüber hinaus zeigt unsere Analyse, dass \textit{NLE-Sufficiency} und \textit{-Comprehensiveness} nicht notwendigerweise mit \textit{Attribution-Similarity} korreliert sind, und wir diskutieren wie die beiden Metriken genutzt werden können, um weitere Einblicke in die Erklärungsgenerierung zu gewinnen.
\fancyhead[LE,RO]{\it Abstract}

\cleardoublepage

\fancyhead[LE,RO]{\it Contents}
\tableofcontents
\cleardoublepage

\fancyhead[LE]{\it \leftmark}           
\fancyhead[RO]{\it \rightmark}          
\fancyhead[LO,RE]{}                     

\pagenumbering{arabic}

\chapter{Introduction}\label{chapter:introduction}

Despite the enormous success of deep learning in countless disciplines, widespread adoption of the technology is still inhibited by its lack of transparency: It is generally difficult to explain the decision-making of a deep neural model, especially to a user of the system~\cite{holzinger2017we}. The requirement of interpretability has led to the emergence of various methods to explain these models, popularized under the term \textit{eXplainable AI} (XAI)~\cite{arrieta2020explainable}. However, most methods are focused on mathematical approaches to finding accurate explanation models, while the view of the explainee, who ultimately needs to understand and interpret the information conveyed is less often considered~\cite{miller2019explanation, jacovi2022diagnosing}. One promising research direction for developing more readily comprehensible explanations is the generation of natural language explanations (NLEs), which provide an intuitively understandable rationalization of a neural network's decision~\cite{ehsan2018rationalization}.

Recently, the e-ViL benchmark~\cite{kayser2021vil} was proposed as a tool to evaluate models that generate NLEs in visual-language tasks (VL-NLE). It assigns a score to a model's prediction accuracy (the answer of the VL task), and another score to the quality of the explanations provided by the model on three different datasets. The explanations of the models are trained via supervised learning on provided ground-truth explanations in the datasets. Evaluating the leading models on the benchmark via participant ratings reveals their ability to produce generally convincing explanations. However, it is an open question in how far the given explanations accurately describe the decision-making process of the model, a property known as \textit{faithfulness}~\cite{jacovi2020towards}. In the extreme case, a non-faithful explanation could be convincing and plausible, while describing factors that are entirely irrelevant to the model's prediction. Such persuasive and misleading explanations are especially dangerous in applications where an accurate understanding is crucial, such as in legal cases or in medical diagnosis~\cite{holzinger2017we}. Nevertheless, a faithfulness score is not yet a component of the e-ViL benchmark, and current faithfulness metrics are either not adaptable to NLEs, or are not designed for comparing model architectures across multiple modalities.

To address this issue, we build on previous research on explanation faithfulness to introduce three metrics for measuring the faithfulness of VL-NLE models: \textit{Attribution-Similarity}, \textit{NLE-Sufficiency} and \textit{NLE-Comprehensiveness}. These metrics are designed to extend the existing evaluation methods of VL-NLE models, such as the e-ViL benchmark, by a faithfulness dimension.  

To evaluate the efficacy of the proposed metrics, we investigate the impact of different ablations of the recent VL-NLE model e-UG on the metrics, for which we assume different changes to the model's faithfulness. Like other VL-NLE models, e-UG consists of a task- and an explanation-generation module. For our study, we are successively removing redundant inputs to e-UG's explanation generation module, so that it is required to generate the explanation from features supplied by the task module, instead of creating unfaithful rationales primarily from the input question and predicted label. Such rationales are less dependent on the actual reasoning process that led to the prediction of the task answer. We hypothesize that models, which rely solely on the task module features, achieve superior scores on the proposed metrics compared to the standard e-UG configuration, and that we observe a further deterioration in faithfulness scores for baseline models that don't use any task-module features, but instead learn unfaithful rationales based on the input question and answer label alone. In addition, we investigate whether the more faithful model variants are still able to generate convincing explanations and provide conclusions on designing more faithful VL-NLE models based on our findings.

\subsubsection{Acknowledgement}
We gratefully acknowledge GPU computing resources provided by ZAL Center for Applied Aeronautical Research GmbH\footnote{\url{https://zal.aero}}.
\cleardoublepage

\chapter{Related Work}\label{chapter:related_work}

In this chapter, we review a selection of relevant related work. First, we present definitions of explanations faithfulness, and metrics that have been proposed to measure it. Afterward, we introduce current research on natural language explanations (NLEs) and finally discuss methods for designing and training faithful NLE-generating models.

\section{Explanation Faithfulness}\label{sec:faithfulness}

Current explanation techniques are primarily evaluated by assessing the degree to which they are understandable by users. However, multiple authors have argued that these human evaluations only cover one aspect of explanation quality, that can be described as \textit{persuasiveness} or \textit{plausibility}, while the degree to which an explanation accurately represents the reasoning process of a model is difficult to assess in participant studies~\cite{herman2017promise, jacovi2020towards}. This latter property of explanations is known as \textit{faithfulness}.

Camburu defines faithfulness as follows: ``Faithfulness of an explanation refers to the accuracy with which the
explanation describes the decision-making process of the target model. Faithfulness of
an explanation should not be confused with the property of an explanation to provide
ground-truth argumentation for solving the task at hand, which is independent of
a model decision-making process''~\cite{camburu2020explaining}. This definition highlights that plausible and objectively correct explanations are not necessarily faithful if the model's decision-making process follows a line of reasoning that is different from some plausible ground-truth explanation but still leads to the same conclusion.

Similarly, Jacovi and Goldberg differentiate the two properties as follows: ``Two particularly notable criteria [for explanations], each useful for different purposes, are plausibility and faithfulness. \textit{Plausibility} refers to how convincing the
interpretation is to humans, while \textit{faithfulness} refers to how accurately it reflects the true reasoning process of the model.''~\cite{jacovi2020towards} Similar distinctions between these two properties have been made by other authors~\cite{herman2017promise, wiegreffe2020attention}. In their positional paper~\cite{jacovi2020towards}, Jacovi and Goldberg argued that explanation faithfulness should be viewed as a continuum, rather than a binary property where an explanation model would be judged as unfaithful on the basis of a single or few counter-examples.

\section{Faithfulness Metrics}
\label{subsec:faithfulness-metrics}
As laid out in the previous section, the faithfulness of an explanation cannot be assessed by recording explanation ratings from human participants which rather captures the aspect of explanation plausibility or persuasiveness. To address the problem of measuring explanation faithfulness, several approaches have been proposed for various explanation methods. In this section, we summarize and discuss six methods for measuring faithfulness that vary in their approach and/or in the kind of explanations they are addressing. 

\subsection{SelPredVerif Framework}
Camburu et al.~\cite{blunsom2019can} proposed the \textsc{SelPredVerif} framework for assessing the faithfulness of feature relevance attribution methods (See Section~\ref{sec:ig-feat-attr}). The framework is designed around identifying a subset of a dataset, where for each example a set of features that are irrelevant to a model prediction can be determined. This subset of examples is created by training a selector-predictor architecture on the dataset, which consists of a selector module performing a hard selection of a minimum subset of input features and a predictor module that generates the actual prediction~\cite{camburu2020explaining}. Next, relevance attribution methods can be evaluated on this subset that is determined by the selector module of the trained model. The framework is intended as a sanity check for attribution methods, where they should minimize the relevance that is attributed to the irrelevant features. However, due to its design, the framework is limited to the evaluation of selector-predictor models and does not check the faithfulness in terms of the relative importance given to features that \textit{do} contribute to a prediction.

\subsection{Sufficiency and Comrpehensiveness}
De Young et al.~\cite{deyoung2020eraser} followed a different approach for assessing the faithfulness of extractive rationales in natural language processing (NLP), which are highlighted parts of a sentence that are particularly important for a certain prediction, such as important words for sentiment analysis or document classification. In their \textit{ERASER} benchmark, De Young et al. propose two metrics for measuring faithfulness: \textit{sufficiency} and \textit{comprehensiveness}. Given a set of tokens that are selected for a rationale, the \textit{sufficiency} is calculated as the change in the model's confidence for a given class, when only the selected rationale is used as an input. On the other hand, the \textit{comprehensiveness} of a rationale is assessed as the change in the model's confidence when the prediction is made on the basis of all the inputs that are not selected for the rationale. A faithful rationale should therefore display low \textit{sufficiency} values, as the prediction should be highly influenced by the tokens selected for the rationale, but high \textit{comprehensiveness} values, as all the other tokens should be relatively unimportant for the prediction. We discuss these two measures in further detail and extend them to natural language explanations in Section~\ref{subsec:sufficiency-comprehensiveness}.

\subsection{Cosine-Similarity of Answer and NLE Attribution}
\label{subsec:cos-sim-rel-wu-mooney}
To assess the faithfulness of Natural Language Explanations (NLEs) in visual-question answering (VQA), Wu and Mooney developed a faithful VQA model and a score to assess the degree of faithfulness of textual explanations~\cite{wu2018faithful}. Their approach to building a faithful VQA model is discussed in Section~\ref{sec:faithful-architectures}. The authors used GradCAM~\cite{selvaraju2017grad} to identify visual features that are relevant to the answer and the textual explanation, respectively. The resulting explanation vectors are then compared via cosine similarity, resulting in a score from -1 to 1. 

Wu and Mooney's approach is closely related to the proposed \textit{Attribution-Similarity} metric in Section~\ref{subsec:cosine-sim-rel-attr}, with the following two differences: (i) The authors use GradCAM, while we are proposing to use Integrated Gradients due to GradCAMs gradient saturation limitations (see Section~\ref{sec:ig-feat-attr}). (ii) Wu and Mooney assess the faithfulness only on an intermediate representation of the visual modality, while we extend the metric to the input question and input image representations. These two changes enable the assessment of faithfulness in both modalities and facilitate the comparison of different models on the faithfulness metric.

\subsection{Similarity of Relevant Image Segments and Referenced Objects in the Explanation}
In addition to the metric mentioned above, Wu and Mooney proposed another method for evaluating the faithfulness of NLE-rationalized VQA models. Their VQA model received image segmentation features as visual input features. Therefore, the authors were able to match the segments that are relevant to the answer prediction to words in the explanation that are likely referencing these segments. The relevant image segments are identified using LIME~\cite{ribeiro2016should} and then matched to common nouns in the explanation by selecting the most salient image segments according to attention weights in the model when this noun was generated in the explanation generation module. The faithfulness score is now calculated by multiplying the corresponding LIME weight (relevance) of each identified LIME segment with the cosine similarity of the closest matching referenced object, as follows:

\begin{equation*}
    s_{LIME} = \frac{\sum^{|V|}_{i=1} |w_i| max_{j\in\mathcal{L}} cos(v_i, v_j)}{\sum^{|V|}_{i=1} |w_i|}
\end{equation*}

With the visual features $V$, where $v_i$ is the visual feature of the i-th segment with the corresponding LIME relevance weights $w_i$, and $\mathcal{L}$ is the set of image segments linked to explanation nouns.

Due to the specific design of the score, which relies on the implementation details of the model that the authors used, this metric is not readily applicable to other models or even tasks.

\subsection{Robustness Equivalence}

Wiegreffe et al.~\cite{wiegreffe2021measuring} proposed two measurements designed to assess necessary conditions for faithful natural language explanations in NLP tasks. One of them is \textit{Robustness Equivalence}, where a zero-mean Gaussian noise with different $\sigma$ values is applied to the inputs of a model that is predicting jointly a task label and an NLE. The authors argue that for the explanation to be faithful, the explanation output must be tied to the label prediction. Therefore, the label and the explanations should be similarly susceptible to noise. More specifically, Wiegreffe et al. are measuring if the explanation and task output are similarly unstable at the same noise levels ($\sigma$ values). For the task prediction, the stability is assessed by measuring the label stability (how many labels across the dataset are flipped) and the overall accuracy. The explanation generation stability is determined by plotting the simulatability of the rationales using an automated metric: Traditionally, the simulatability metric represented the overall change in human task performance (simulating a model on the dataset) when subjects received rationales in addition to the task input. Therefore, simulatability is a measure of explanation quality. The automated version of this metric is realized by training a model on task inputs \textit{and} ground-truth rationales and observing the difference in prediction accuracy on a test dataset when using the rationales generated by the evaluated model, in comparison to a second model that just predicts the test dataset \textit{without} being exposed to rationales during training or inference.

\subsection{Feature Importance Agreement}

\textit{Feature Importance Agreement} is another necessary condition for faithfulness proposed by Wiegreffe et al.~\cite{wiegreffe2021measuring}. This measurement is taken by removing the top k (e.g $k \in \{10\%, 20\%, 30\%\}$) features of the label prediction and explanation generation according to a feature importance attribution method (the authors used Gradient Attribution~\cite{baehrens2010explain}) and subsequently observing the impact on the output label accuracy and explanation simulatability. This is done in such a way that the impact on the label accuracy was measured after removing the $k$ most important explanation features, and vice versa. Crucially, after the removal of the features, a new model is trained on only the remaining features and the difference in performance to the original model using all features is recorded, not only for calculating the simulatability but also for the label accuracy~\footnote{This method is known as ROAR (\textbf{R}em\textbf{o}ve \textbf{a}nd \textbf{R}etrain)~\cite{47088}}. According to Wieggreffe et al, a necessary condition for faithfulness is that removing the important explanation features should lead to a similarly decreased label accuracy and the removal of important label features should correspondingly lead to a similar decrease in explanation simulatability.

\section{Natural Language Explanations}

A significant amount of approaches in the field of explainable AI are focused around assigning a relevance weight to a set of input features. Usually, the resulting feature attribution vectors are then visualized, for example using a heatmap (e.g. \cite{selvaraju2017grad, bach2015pixel}). While these approaches can deliver crucial insights into the functioning of a model, a major drawback is the difficulty of interpreting them, especially for end-users that are not machine learning experts, since the way in which humans typically communicate is drastically different. For example, humans are using contrastive explanations by referring to counterfactuals, or social attribution by ascribing intentionality to actors~\cite{miller2019explanation, mittelstadt2019explaining}. These types of explanations and more complex chains of individual reasoning steps can be expressed in natural language. To bring AI explanations closer to their human counterparts, annotated datasets of explanations in natural language have been proposed recently. For example, the e-SNLI dataset~\cite{camburu2018snli} contains human-annotated rationales for a series of language entailment tasks of the SNLI dataset~\cite{maccartney-manning-2008-modeling}. The dataset was later merged with the visual-entailment dataset SNLI-VE~\cite{xie2019visual} into e-SNLI-VE~\cite{kayser2021vil}, which is used in the e-ViL benchmark alongside VQA-X~\cite{park2018multimodal} and VCR~\cite{zellers2019recognition} that are described in Section~\ref{subsec:evil-datasets}. Another vision-language dataset annotated with natural language rationales that is not included in the e-ViL benchmark is ACT-X, where, based on images of people carrying out different activities,  a sentence of the form ``I can tell the person is doing [action] because..." needs to be completed. Although not part of e-ViL, the dataset received the attention of researchers again more recently~\cite{sammani2022nlx}.

\section{Model Design Considerations for Faithful NLE Generation}
\label{sec:faithful-architectures}

As the overall architecture of neural models and their training process can likely both have a significant impact on the faithfulness of generated natural language explanations, we will explore, from related literature, one example for architectural considerations and one for a training process that aims to increase the faithfulness of generated explanations in this section.

\subsection{Architecture: Order of Prediction and Explanation}

Camburu et al.~\cite{camburu2018snli} presented a study of different architectural setups on the NLP dataset e-SNLI. We will highlight the differences from a faithfulness perspective for two high-level architectures that the authors presented: (i) the \textsc{PredictAndExplain} and (ii) the \textsc{ExplainThenPredict} architecture.

\begin{figure}
    \centering
    \includegraphics[width=0.5\textwidth]{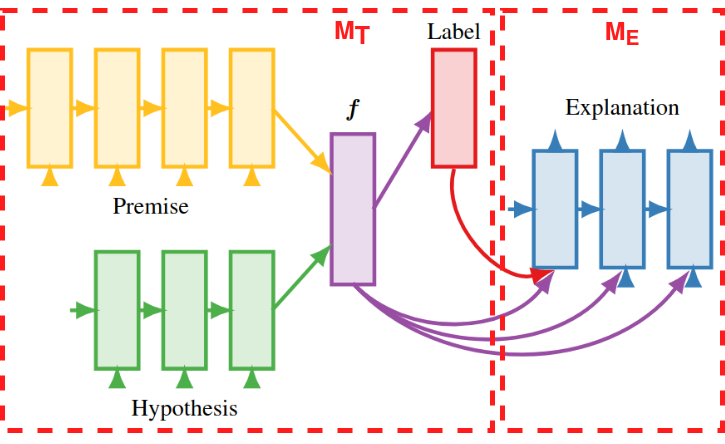}
    \caption{Structure of the \textsc{PredictAndExplain} architecture on a language-entailment and natural language explanation dataset. Figure adapted from Camburu et al.~\cite{camburu2018snli}}
    \label{fig:predict-and-explain}
\end{figure}

\paragraph{PredictAndExplain.}
The \textsc{PredictAndExplain} architecture is depicted in \autoref{fig:predict-and-explain}. This configuration can be considered the standard for rationalized models, as the explanation generation is a mere extension of an existing task model. Most models evaluated on the e-ViL benchmark correspond to this structure consisting of a task-module $M_T$, predicting the solution to a task, and subsequent explanation generation module $M_E$, generating the natural language explanation~\cite{kayser2021vil}. The explanation generation module has access to features from $M_T$, as well as to the predicted task label. However, as the explanation is generated post factum after the label was already predicted, the faithfulness of the explanation is not structurally guaranteed.

\paragraph{ExplainThenPredict.}
The order of prediction and explanation is reversed in the \textsc{ExplainThenPredict} architecture\footnote{Models designed according to this architecture are also known as pipeline models~\cite{wiegreffe2021measuring}}. Here, a sequence-to-sequence model is trained to directly predict an explanation from a given premise and hypothesis (i.e. no prediction of ``Label" in \autoref{fig:predict-and-explain}). Afterward, the label is predicted by a separate model solely from the explanation. On one hand, this requires the explanation to contain sufficient information to be able to predict the label correctly. On the other hand, the label prediction needs to be based only on the information contained in the explanation, which by design increases the faithfulness of the explanation. In experiments presented by Camburu et al.~\cite{camburu2018snli}, the \textsc{ExplainThenPredict} architecture resulted in only a minor decrease of prediction accuracy in the two investigated datasets e-SNLI (0.25\%) and InferSent~\cite{conneau2017supervised} (0.26), in comparison to the  \textsc{PredictAndExplain} variant. However, in a study by Wiegreffe et al.~\cite{wiegreffe2021measuring}, using a \textsc{ExplainThenPredict} architectural setup resulted in a significant decrease in label accuracy and explanation quality on the newer CoS-E datatset~\cite{rajani2019explain}, which contains more label classes than e-SNLI, compared to jointly predicting labels and explanations in a \textsc{PredictAndExplain} manner.

\subsection{Training Process: FME (Wu and Mooney 2018)}
Based on their faithfulness score discussed in Subsection~\ref{subsec:cos-sim-rel-wu-mooney}, which is calculated as the cosine-similarity of gradCAM attributions for explanation generation and answer prediction -- denoted $s_{GCAM}$ in the following -- Wu and Mooney introduced two interventions to the training process of their FME model to increase its faithfulness.

First, they filtered out explanations from the VQA-X training dataset that they considered being ``unfaithful" with regard to the model's behavior. The VQA-X dataset contains multiple ground-truth explanations for the correct answers to the VQA tasks. Using $s_{GCAM}$, the authors applied a filter during the training of the model, which removes all ground-truth explanations where the cosine is below $\epsilon~max(0.02~it, 1)$, such that $\epsilon$ is a set threshold, $it$ the number of iterations, and $max(0.02~it, 1)$ is, according to the authors, ``used to jump-start the randomly initialized explanation module''. We interpret this last term as a means to remove a sufficiently large subset of explanations already early during the first training iterations but are unsure about the concrete advantages of this approach.

As a second measure, they again used the same faithfulness metric $s_{GCAM}$ as a regularization term in the model's loss function, by aligning the important feature areas of the explanation generation and label prediction. More exactly, they introduce a faithfulness loss  $\mathcal{L}_F~=~{1-s_{GCAM}}$. The overall model is then trained on an equally weighted combination of explanation loss as cross-entropy on the ground-truth explanation, task label prediction cross-entropy loss, and $\mathcal{L}_F$ faithfulness loss.

The result of these two measures on the test split evaluation is a 7\% increase in $s_{GCAM}$ for the dataset filtering and an additional 11\% increase when combining the filtering with the $\mathcal{L}_F$ loss. Moreover, Wu and Mooney also reported an increase in task and explanation performance; however, Kayser et al. were unable to reproduce this result when re-implementing FME for evaluation on the e-ViL benchmark~\cite[appendix]{kayser2021vil}.

\cleardoublepage

\chapter{Preliminaries}\label{chapter:fundamentals}

The following chapter provides an overview of important preliminaries and describes the  specifics required for the approach, results, and conclusions presented in this thesis. This includes the datasets of the e-ViL benchmark, on which the proposed faithfulness metrics are evaluated, central aspects of the transformer architecture, as well as the transformer-based e-UG model that is employed and modified to assess the resulting changes on the faithfulness metrics, and, lastly, integrated gradients, a method used for feature attribution, which forms the basis for the faithfulness metrics.

\section{e-ViL Benchmark}
\label{sec:evil}

The e-ViL benchmark, proposed by Kayser et al.~\cite{kayser2021vil}, is an evaluation framework for models on vision-language and natural language explanation (VL-NLE) tasks.
It consists of the three datasets e-SNLI-VE, VQA-X, and VCR, and provides two evaluation dimensions: (1) the VL-task label accuracy and (2) an explanation rating assessed by collecting the subjective ratings of human participants for the natural-language explanations.

\subsection{Datasets}
\label{subsec:evil-datasets}

\begin{figure}
    \centering
    \includegraphics[width=\textwidth]{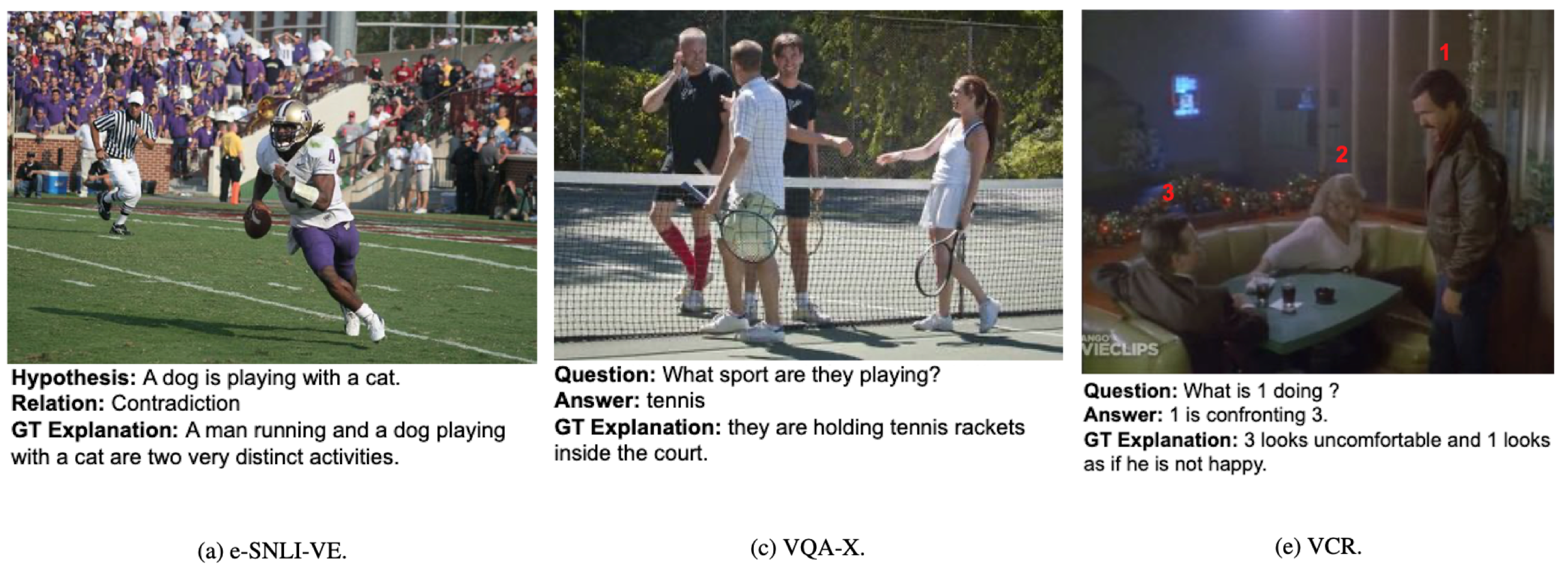}
    \caption{Example tasks from the three e-ViL datasets with the annotated question, answers and ground-truth explanations. Figure adapted from Kayser et al.~\cite{kayser2021vil}.}
    \label{fig:examples-evil}
\end{figure}

The three datasets of the benchmark vary considerably in the dataset size and task-specifics. \autoref{fig:examples-evil} provides an example for each dataset.

\paragraph{e-SNLI-VE.} The e-SNLI-VE dataset~\cite{kayser2021vil} was constructed by merging the e-SNLI~\cite{camburu2018snli} and the SNLI-VE~\cite{xie2019visual} datasets, which results in a visual entailment (VE) task with accompanying natural-language explanations. In a VE task, a premise in the form of an image and a textual hypothesis are supplied. The task is to predict the relationship between the premise and the hypothesis as \textit{entailment}, \textit{contradiction} or \textit{neutral}. e-SNLI-VE consists of a train, dev, and test split distribution of 401,717, 14,339 and 14,740 instances, respectively, which makes the dataset considerably larger than the VQA-X or VCR dataset.

\paragraph{VQA-X.} The VQA-X dataset~\cite{park2018multimodal} is a subset of the VQA-V2 dataset~\cite{balanced_vqa_v2} that is annotated with human explanations and contains tasks split into train, dev and test partitions of 29,500, 1,500 and 2,000 items. The images of the dataset are taken from the COCO dataset~\cite{lin2014microsoft}. Each task instance comprises of a question and an accompanying image to answer it by predicting the correct answer class from a set of 3,129 possibilities, where multiple answers can be correct.

\paragraph{VCR.} Visual Commonsense Reasoning (VCR)~\cite{zellers2019recognition} is a multiple choice (single answer) dataset of questions about images captured from movie scenes. Each instance has four answer possibilities that are appended to the input question. In the original VCR task formulation, a rationalization of the answer had to be selected from four possibilities as well, but the task was re-formulated as a free NLE-generation task for the e-ViL benchmark, similarly to the other two datasets.

\subsection{Benchmark Scores}

\begin{table*}[htbp]
    \begin{center}
    \begin{tabulary}{\linewidth}{LCCCCCCCCCC}
    \toprule
        & Overall  & \multicolumn{3}{c}{VQA-X} & \multicolumn{3}{c}{e-SNLI-VE} & \multicolumn{3}{c}{VCR} \\
          \cmidrule(r){2-2} \cmidrule(r){3-5} \cmidrule(r){6-8} \cmidrule(r){9-11}
          & $S_E$         & $S_O$         & $S_T$         & $S_E$         & $S_O$         & $S_T$         & $S_E$         & $S_O$         & $S_T$ & $S_E$ \\
    \midrule
    PJ-X  & 59.2          & 49.9          & 76.4          & 65.4          & 41.2          & 69.2          & 59.6          & 20.6          & 39.0  & 52.7          \\
    FME & 60.1          & 47.7          & 75.5          & 63.2          & 43.1          & 73.7          & 58.5          & 28.6          & 48.9  & 58.5          \\
    RVT   & 62.8          & 46.0          & 68.6          & 67.1          & 42.8          & 72.0          & 59.4          & 36.4          & 59.0  & 61.8          \\
    e-UG  & \textbf{68.5} & \textbf{57.6} & \textbf{80.5} & \textbf{71.5} & \textbf{54.8} & \textbf{79.5} & \textbf{68.9} & \textbf{45.5} & \textbf{69.8}  & \textbf{65.1} \\
    GT & 79.3          & \multicolumn{1}{l}{--} & \multicolumn{1}{l}{--} & 84.5          &   --            &     --          & 76.2          &           --    &    --   & 77.3  \\
    \bottomrule
    \end{tabulary}\vspace{-1ex}
  \caption{The e-ViL benchmark score results according to the experiments by Kayser et al.\cite{kayser2021vil}. GT denotes the ground-truth explanations. The best results are printed in bold.}%
  \label{tab:evil-scores}
  \end{center}
  \vspace{-4ex}
\end{table*}

The benchmark metrics consist of a task score $S_T$ and an explanation score $S_E$, as well as an overall score $S_O = S_T*S_E$, all of which are calculated for each dataset independently. $S_T$ is simply the test-split label accuracy of answering the VL task in percent. Kayser et al. assessed the explanation score $S_E$ using human participants that had been recruited via MTurk\footnote{\url{https://www.mturk.com/}}~\cite{kayser2021vil}. Participants were asked the question ``Given the image and the question/hypothesis, does the explanation justify the answer?" with the possible answer choices \textit{yes}, \textit{weak yes}, \textit{weak no} or \textit{no}, mapped to 1, $\frac{2}{3}$, $\frac{1}{3}$, and 0. For each dataset and model, a random sample of 300 correctly answered questions with unique images has been selected to evaluate the explanation rating. Each combination of model and dataset is evaluated by three workers to minimize the subjective annotator bias and the ratings of each example are averaged. $S_E$ is then computed as the average of these scores across the dataset.

Table \ref{tab:evil-scores} lists the benchmark scores that were assessed for different models by Kayser et al. When the benchmark was published, the highest score across all datasets and benchmark dimensions was achieved by the e-UG model, which is used in this work and described in more detail in Section~\ref{sec:eug_model}.

\section{Transformer Models}\label{sec:transformer}

The introduction of the transformer architecture was initially motivated by machine translation but quickly revolutionized the field in a variety of natural language processing tasks by leveraging transformer-based language models trained via self-supervised learning~\cite{vaswani2017attention, kenton2019bert, brown2020language}. Scaling up the transformer models to higher numbers of parameters, enabled by the high parallelization capabilities, further increased the performance of the models and opened up new opportunities for using the models, such as zero- or few-shot learning~\cite{brown2020language}. Models based on the architecture were also successful in reaching state-of-the-art performance in vision~\cite{dosovitskiy2020image}, vision-language tasks~\cite{chen2020uniter}, or even text-to-image generation~\cite{ramesh2021zero}.

\subsection{Background}

Before the introduction of the transformer architecture, successful sequence modeling relied primarily on recurrent neural networks (RNNs), such as the long short-term memory (LSTM)~\cite{hochreiter1997long} or the gated recurrent unit (GRU)~\cite{cho2014properties}. Recurrent neural networks process one element of a sequence at each timestep, and aggregate this input $i_t$ within a hidden variable $h_t$, where the input is combined with the hidden variable from the previous timestep $h_{t-1}$. This design has two inherent limitations: Firstly, the aggregated hidden representation $h$ is required to represent the entire sequence at the final timestep $t$, and, therefore, acts as an information bottleneck that inhibits the representation of long-range dependencies. Secondly, the input sequences are fed element-by-element, in a sequential manner, to the RNN. Therefore, the computation of long input sequences cannot be executed in parallel, which hinders the overall parallelization capability of RNNs during training, as the input sequences need to be held in memory for each running instance~\cite{vaswani2017attention}.

Sequence-to-sequence approaches based on convolutional neural networks (CNNs) have been proposed to mitigate the RNN's parallelization issues by computing a representation for each input simultaneously (e.g.~\cite{kalchbrenner2016neural, gehring2017convolutional}). However, CNNs face a similar issue as RNNs regarding long-term dependencies, since a convolution only operates on a local context, determined by the kernel-size $k$~\cite{lecun1995convolutional}. If the input size $n$ is larger than $k$, which is usually the case, $O(k/n)$ convolutional layers in sequence are required to connect two input representations~\cite{vaswani2017attention}. This can be mitigated by using different convolution algorithms, although the general problem of requiring a deeper network to represent longer dependencies still persists~\cite{vaswani2017attention, chollet2017xception, kalchbrenner2016neural}.

Both CNN and RNN architectures have been extended with attention layers to solve the problem of representing long-range dependencies effectively. The attention mechanism is fundamentally based on assigning weights to representations depending on certain information. By computing an attention map on an entire sequence, depending on each individual representation, each input can be related to any other input, regardless of the position and distance in the sequence. In machine translation, attention explicitly models the word alignments within the model architecture, which significantly improved the quality of translations, especially for long input sequences~\cite{DBLP:journals/corr/BahdanauCB14}.

Building on the success of attention mechanisms, the central idea behind the transformer architecture is to rely on attention alone to process sequences, without employing recurrency or convolutions, thereby solving both parallelization and long-range dependency issues. To this end, the authors introduced a self-attention mechanism that is described, alongside other central details of the architecture, in the following section.

\subsection{Architecture}

\begin{figure}
    \centering
    \includegraphics[width=0.45\textwidth]{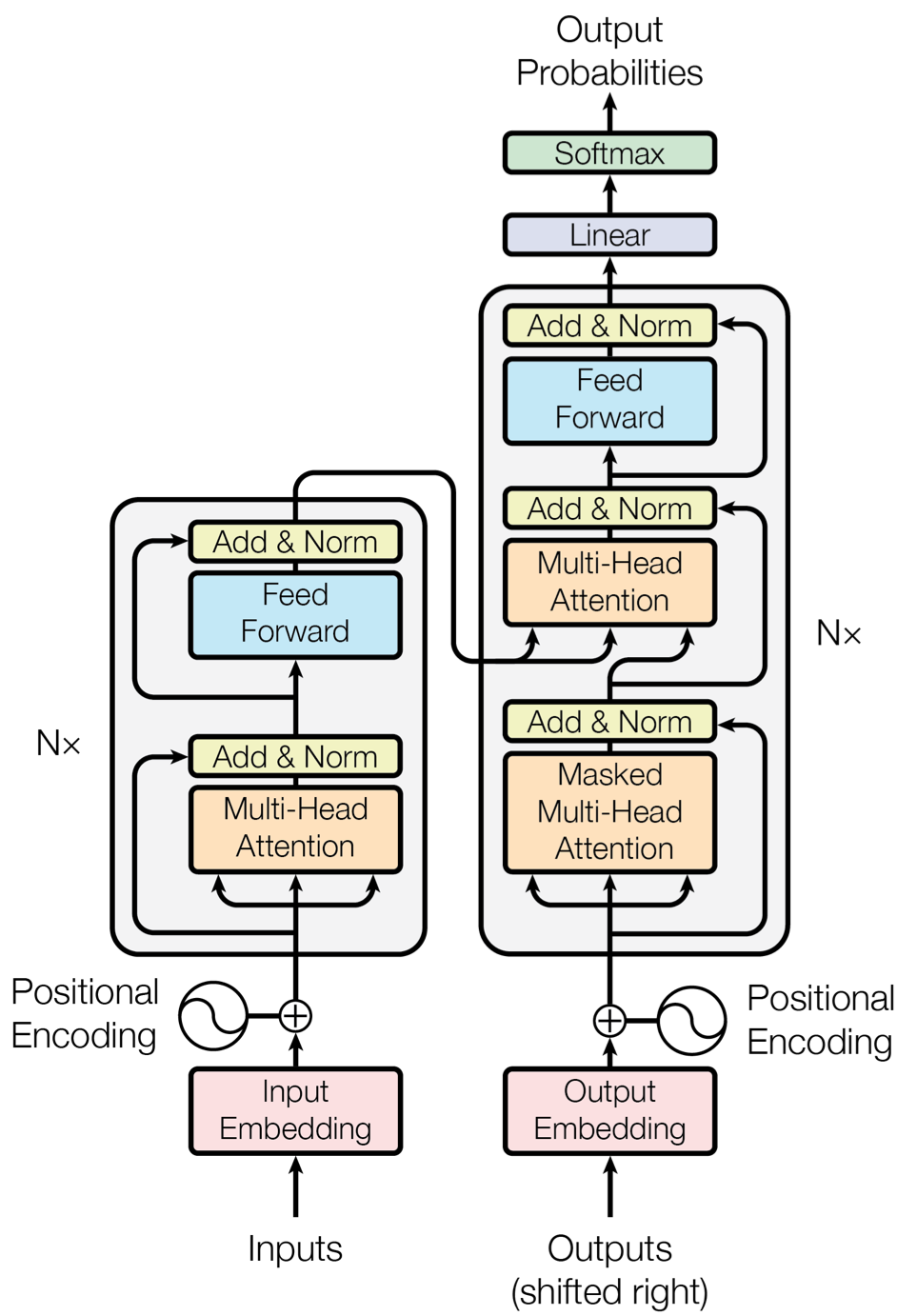}
    \caption{Overview of the transformer model architecture, with the encoder on the left and the decoder module on the right. Figure by Vaswani et al.~\cite{vaswani2017attention}.}
    \label{fig:transformer_architecture}
\end{figure}

An overview of the transformer architecture is depicted in \autoref{fig:transformer_architecture}. The architecture follows the encoder-decoder design, which was previously prevalent in successful sequence-to-sequence models (e.g. \cite{sutskever2014sequence}). Overall, the encoder converts symbolic inputs into a continuous representation that is then again converted to symbolic outputs by the decoder. The transformer model is auto-regressive, which means that it generates one output symbol with each forward pass, processing the previously generated symbols by appending them to the input.

The encoder of an NLP transformer receives symbolic inputs, in the form of token IDs generated from a linguistic input sequence by a tokenizer. These symbols are converted into embedding vectors of size $d$. An unparameterized positional encoding -- composed of sine and cosine functions creating an interference pattern that is unique for each position -- is subsequently added to the embeddings and injects information about the position of the token in the sequence.

Next, the encoder processes input embeddings with added positional encodings through a stack of $N$ encoder blocks. Each encoder block consists of a multi-head attention layer (described below), and a fully-connected feed-forward layer. To the output of both, a residual skip-connection is added (simply adding the input of the layer to its output) and the resulting vector is normalized.

The decoder is again composed of $N$ blocks, which are similar to the encoder blocks. Unlike the encoder blocks, another multi-head attention layer is inserted between the first attention layer and the feed-forward layer. This intermediate layer is attending to the encoder embeddings. To learn the prediction of the next token, which is required for the model to be auto-regressive, the outputs\footnote{As an example, for a machine-translation task, the "output" sequence that is fed into the decoder is the ground-truth translation appended to the input sentence during training. During inference, the input to the decoder is just the input sentence with a specific token indicating the beginning of the translation.} are shifted right, such that each input to the decoder layer is predicting the next token in the sequence. Additionally, the attention layers in the decoder are masked, ensuring that a token representation can attend only to those tokens that precede the current token in the sequence.
The decoder blocks are subsequently fed into a linear layer that converts the token representations of size $d$ into a vector that is the size of the model's vocabulary and shares the weights with the two embedding layers. This way, after applying a softmax function to the output, the probabilities for the next token are predicted. 

\subsubsection{Attention and Self-Attention}

\begin{figure}
    \centering
    \includegraphics[width=0.8\textwidth]{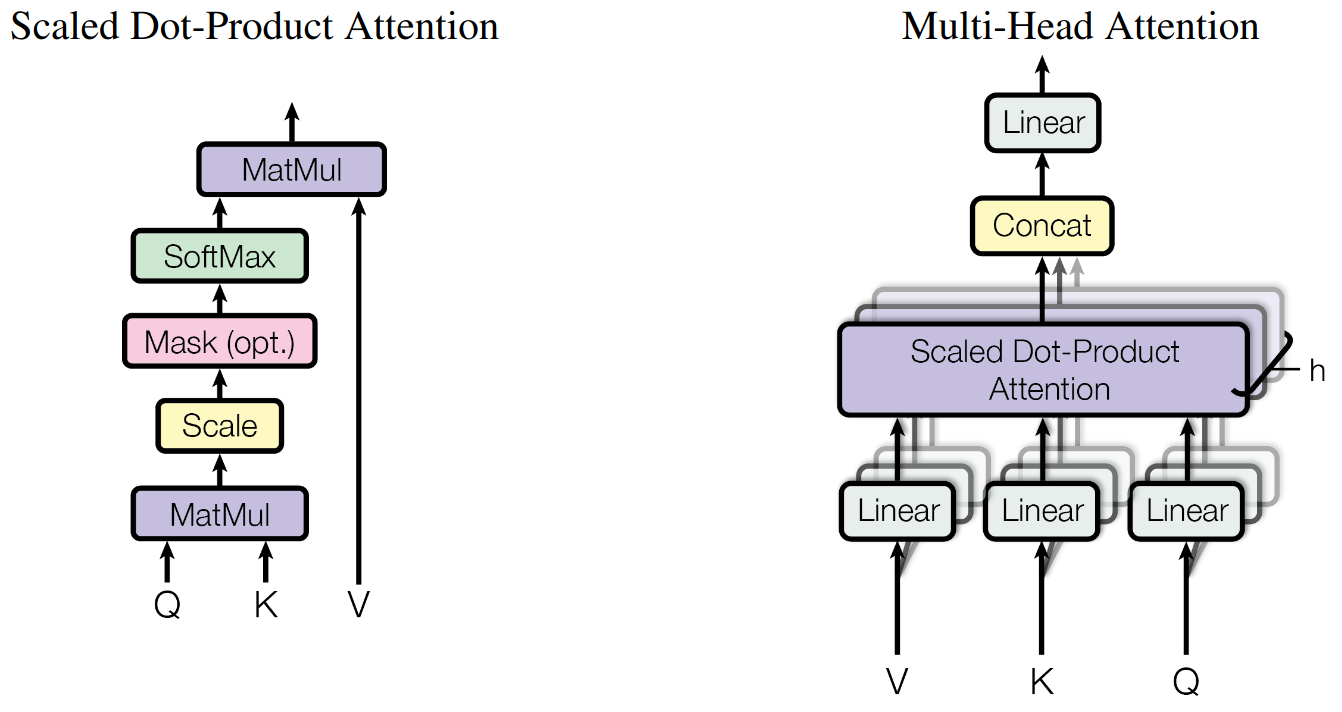}
    \caption{The scaled dot-product and multi-head attention used in transformer models. Figure by Vaswani et al.~\cite{vaswani2017attention}.}
    \label{fig:attention-transformer}
\end{figure}

Attention layers are the core component of transformer models. The attention mechanism that is employed in these models is the scaled dot-product attention, visualized in \autoref{fig:attention-transformer}. Given a query $Q$ and key-value pairs consisting of $K$ and $V$, the central idea behind attention is to first check the compatibility of $Q$ and $K$, which can be conceptualized as finding the keys that correspond closely to a query, yielding a set of ``attention weights". Next, the values $V$ are summed up according to these weights, which results in a scalar for each embedding in the sequence. For example, $K$ could be a sequence of word embeddings representing a sentence, with $Q$ being one of the words in the sentence and $V$ a sequence of scalars that represent the sentiment of each word. The output of the attention operation would then be a scalar representing the attention-weighted mean sentiment of a sentence, in relation to the token $Q$.

In the case of self-attention, each token embedding in a sequence $K$ is used as a query $Q$, implemented as a batch of queries, and the token sequence itself is used as the value matrix $V$, such that $K = Q = V$ for a simple, unparameterized attention mechanism. Effectively, this attention mechanism uses the tokens in a sequence directly to compute the ``compatibility'' between each of the tokens in the sequence, for example by way of matrix multiplication.

\subsubsection{Scaled Dot-Product Attention and (Masked) Multi-Head Attention}

The scaled dot-product attention that is used in the transformer architecture is a specific attention mechanism. It functions by multiplying matrices $Q$ and $K$, effectively computing the dot product for each query packed in the matrix $Q$ with each key in $K$. Afterward, the result is scaled by the square root of the embedding dimension of $K$ as $\sqrt{d_k}$ to prevent saturation of the later following softmax function for large values of $d_k$. At this stage, a mask can be applied to the attention mechanism by setting all values corresponding to a connection of a token that appears later in the sequence to $-\infty$, which results in 0 values when applying the softmax function in the next step that normalizes the vector in the range of 0~to~1. Subsequently, another matrix multiplication is applied to finally relate the attention weights to the values.

For the multi-head attention layer, multiple scaled dot-product attention mechanisms run in parallel. The inputs $V$, $K$, and $Q$ are projected onto a lower dimension using a parameterized linear layer, and the scaled dot-product attention is executed on these lower dimensional representations. Next, the results of the multiple attention operations are concatenated and projected back into the original dimension $d$. By choosing dimensions for $V, K, Q$ according to $d \div h$ with $h$ as the number of heads, different attention heads can be trained that learn to attend to different information at a lower resolution, instead of only one attention mechanism, without increasing the computational complexity significantly.

While the sequence embeddings from the current block are used for $V$, $K$, and $Q$ as self-attention for most of the operations in the architecture, the second layer of each decoder block uses the decoder embeddings as $Q$, and the encoder embeddings as $V$ and $K$, thereby attending to the encoder embeddings queried by decoder representations.

\section{e-UG model}
\label{sec:eug_model}

The e-UG model proposed by Kayser et al~\cite{kayser2021vil} is a modification of the Rationale\textsuperscript{VT} Transformer model proposed by Marasovic et al.~\cite{marasovic2020natural}, who combined different vision-language models with a fine-tuned GPT-2 language model for rationale generation. Instead of the vision-language models proposed by Marasovic et al., e-UG uses the UNITER vision-language transformer as a task-module ($M_T$) to predict a task answer label. The representations used to predict the label are then supplied to the GPT-2 explanation-module ($M_E$) to generate the rationale. In the following, we provide an overview of GPT-2 and UNITER, the two significant components of the e-UG model. For a more detailed description of the information flow to the explanation module in the e-UG architecture, we direct the reader to Section~\ref{sec:ablation-models}.

\paragraph{GPT-2.} GPT-2 is a transformer-based language model trained on the WebText dataset that consists of 88 million documents (40GB) of curated websites by collecting outgoing links from positively rated posts on the social media platform Reddit\footnote{\url{https://www.reddit.com/}} and is particularly suited for text-generation~\cite{radford2019language}. Architecturally, the model is a slight variation of the previous GPT model~\cite{radford2018improving}, which utilizes only the decoder stack of the transformer model for generative pre-training on unsupervised language data. For GPT-2, the number of model parameters was significantly increased, in addition to the modifications of the architecture and the WebText dataset. For e-UG, the GPT-2 model is receiving the UNITER representations directly without an additional embedding layer, as both models share the same embedding dimensionality, while additional language inputs are embedded using GPT-2's own input embeddings.

\paragraph{UNITER.} The UNITER model is a vision-language transformer that learns joint image-text embeddings via a set of masking-based self-supervised tasks across different datasets \cite{chen2020uniter}. After task-specific fine-tuning, it reached state-of-the-art results in many vision-language tasks when it was published. To generate image embeddings, a set of features are extracted from regions-of-interest using a Faster-RCNN model~\cite{ren2015faster}, unlike other approaches which simply split the input image into equal-sized patches (e.g. a previous approach by Dosovitskiy et al~\cite{dosovitskiy2020image}). The region features are then, together with positional information, projected onto the embedding space using a linear layer, while the text input is embedded using a BERT tokenizer~\cite{devlin2019bert}.

\section{Integrated Gradients for Feature Relevance Attribution}
\label{sec:ig-feat-attr}

Over the recent years, a multitude of approaches has been proposed for solving the problem of assigning a relevance score to each feature of a machine-learning model (particularly a neural network) according to their importance to either generating a particular prediction or according to their global importance for all predictions across a dataset. This is also known as the \textit{attribution problem} (cf. \cite{lundberg2017unified, sundararajan2017axiomatic, sundararajan2020many}).

Two early examples of methods developed to tackle the attribution problem are CAM~\cite{zhou2016learning} and GradCAM~\cite{selvaraju2017grad}, where the latter is a generalization of the former that does not require additional modifications to a model. CAM and GradCAM have been proposed primarily as object localization methods for convolutional neural networks in computer vision, although GradCAM has later been used in other settings as a more general attribution method. In contrast to GradCAM, which requires a differentiable model, the method LIME~\cite{ribeiro2016should} learns a local linear approximation (explanation model) of any given model type and explains the importance of features based on their contribution to the linear model. Other methods, such as Layer-wise Relevance Propagation (LRP) are designed to back-propagate relevance through a neural network and require additional implementations of propagation rules depending on the model architecture.

Lundberg and Lee~\cite{lundberg2017unified} argued that the aforementioned approaches and other similar methods can be unified into a class of ``additive feature attribution methods''. According to a set of desiderata, Lundberg and Lee claimed that a unique solution to the attribution problem exists in the form of the Shapley values~\cite{shapley1997value}, which can be approximated by running the model with different combinations of feature-subsets in a method termed SHAP. The uniqueness of this approach was later challenged, as different assumptions about the model, training, and data distribution lead to different solutions~\cite{sundararajan2020many}.

Similarly to SHAP, Integrated Gradients (IG)~\cite{sundararajan2017axiomatic} is also created from a set of desiderata of relevance attribution properties. Unlike SHAP, however, Integrated Gradients uses model gradients to compute the relevance attribution. Therefore, IG requires differentiable models instead of the entirely model-agnostic SHAP approach but is also significantly faster in a practical implementation.

To calculate the relevance attribution for the $i$-th feature of a model $f$ according to Integrated Gradients, the difference between the dataset example $x$ we are explaining and a baseline $x'$ is multiplied by the integral of the gradient between the baseline and the example:

\begin{equation*}
 IG_i(f, x, x') = (x_i - x'_i) \times \int_{\alpha=0}^{1} \frac{\delta f(x' + \alpha (x - x'))}{\delta x_i}d \alpha
\end{equation*}

The approach can be viewed as an extension of the Input $\times$ Gradients method, which simply multiplies the input activation with (positive) gradients, as, intuitively, the areas of the input that have a high gradient are particularly important for the model's prediction. However, in practice, the activation functions of neurons inside the neural network are quickly saturated, leading to flat gradients in areas that are actually highly salient for the prediction. Therefore, the salient regions are taken into account close to the baseline $x'$, even if saturation effects occur closer to the example $x$\footnote{Cf. \url{https://distill.pub/2020/attribution-baselines/} for an intuitive and interactive visualization.}~\cite{sturmfels2020visualizing}. The selection of the correct baseline to compute the IG attributions is an ongoing debate~\cite{sturmfels2020visualizing}. A common choice is using 0-values, such as a black image\ for visual features or a [PAD] token for language input\footnote{Cf. \url{https://captum.ai/tutorials/Bert_SQUAD_Interpret}}.

As the calculation of the integral in the above equation is computationally expensive, it is numerically approximated in practice as the Riemann integral:

\begin{equation*}
 IG_i{approx}(f, x, x') = (x_i - x'_i) \times \sum_{k=0}^{m} \frac{\delta f(x' + \frac{k}{m} (x - x'))}{\delta x_i} \frac{1}{m}
\end{equation*}

With the number of steps $m$ to calculate the Riemann integral approximation.
\cleardoublepage

\chapter{Methods}\label{chapter:methods}

In this chapter, we introduce three metrics for assessing the faithfulness of a natural language explanation (NLE) based on relevance attribution. Further, we outline different modifications that are introduced to the e-UG model to evaluate the resulting changes in the faithfulness metrics.

\section{Assessing NLE Faithfulness using Relevance Attribution}
The proposed faithfulness metrics are based on the identification of relevant input features of a model using feature attribution methods. In the following section, we first provide an intuitive understanding, followed by a formal description of the fundamental idea of using feature attribution methods to asses the faithfulness of an NLE. Afterward, the concrete implementations of faithfulness metrics on the e-ViL benchmark using \textit{Attribution-Similarity}, \textit{NLE-Sufficiency} and \textit{NLE-Comprehensiveness} are described.

\subsection{Intuition}

The calculations of the proposed faithfulness measures are based on identifying relevant features in the input for the model's answer and explanation separately, and then analyzing their differences from each other, as well as the effects of removing certain features.
To identify which parts of input features are decisive for a model's prediction, a multitude of additive feature importance measures, such as SHAP~\cite{lundberg2017unified}, Integrated Gradients~\cite{sundararajan2017axiomatic}, or GradCAM~\cite{selvaraju2017grad}, have been proposed. As models on the benchmark receive an image and a question as inputs, on the basis of which they generate both an answer as well as an explanation for each task, the relevance attribution can be computed either with respect to the answer, or the explanation of a model, yielding the relevant parts of the input image and question for each. 

\begin{figure}
    \centering
    \includegraphics[width=\textwidth]{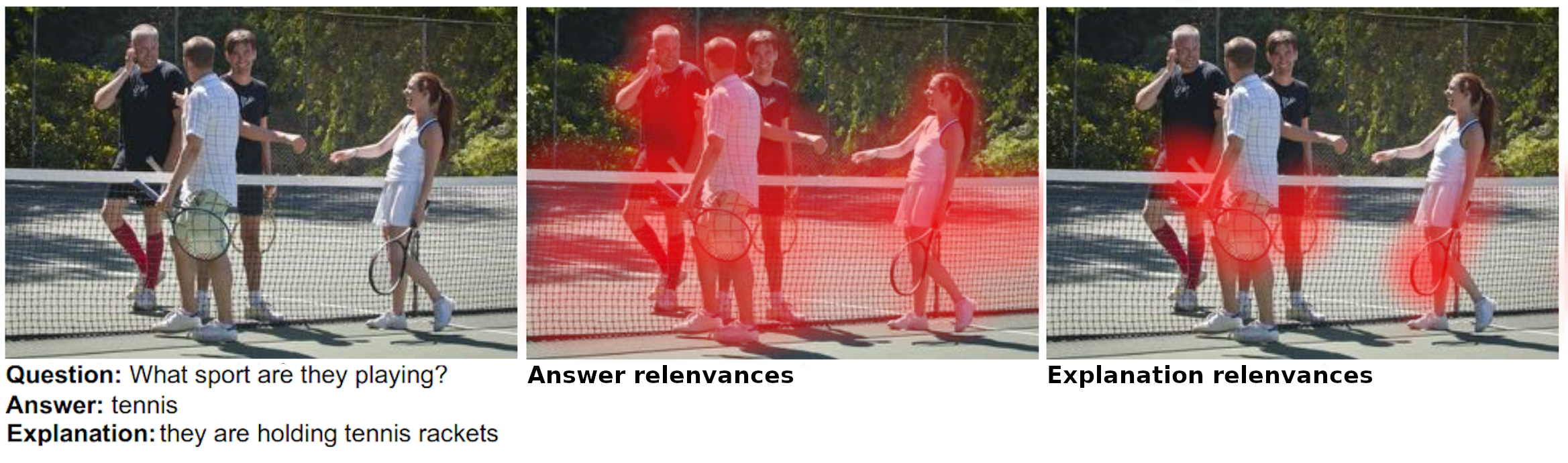}
    \caption{Illustration of feature attribution relevance computed for the explanation and the answer. Since the explanation exhibits different areas of relevance in the image when compared to the answer, the generated explanation cannot be faithful.}
    \label{fig:tennis_example}
\end{figure}

As an example, consider the task in Figure~\ref{fig:tennis_example}: The model receives an image of people standing on a tennis court, along with the question ``What sport are they playing?". The model now correctly predicts the answer ``tennis" and generates the explanation ``They are holding tennis rackets", which was rated as convincing by participants. However, when examining the distribution of feature relevance on the image, it becomes clear that the explanation generation was based only on a subset of the image aspects that were considered for the answer prediction. An interpretation of the relevance attribution could be that the answer prediction was based on detecting the people, as well as the environment (specifically the net), while the explanation generation is rooted solely in the detection of tennis rackets. Due to its lack of \textit{comprehensiveness}\footnote{The property of explanation \textit{comprehensiveness} is discussed in detail in Subsection~\ref{subsec:sufficiency-comprehensiveness}}, the explanation is not faithful, since it leads a user to believe that the model's decision was based primarily on the tennis rackets, while the heatmap in Figure~\ref{fig:tennis_example} suggests that a removal of the rackets from the image could potentially still lead to the same results, as the people and the environment both played an important role in generating the prediction. The three measurements proposed in this thesis are based on a similar automatic assessment of the input relevance distribution of answer and explanation.

\subsection{Formal Rationale}
\label{subsec:formal-rationale}

As the proposed faithfulness measurements share the underlying principle of using additive feature importance measures, they also share the limitations that are introduced with this approach. In this Section, a formal definition and rationale for the proposed faithfulness measures are provided to clearly state the underlying definitions and assumptions. To this end, we first define faithfulness, based on definitions from in the related literature reviewed in Subsection~\ref{sec:faithfulness}.

\begin{definition}[Faithfulness]
\label{def:faithfulness}
The faithfulness of an explanation is the degree to which it represents the reasoning process of a model that it is explaining.
\end{definition}

As this definition relies on the concept of a ``reasoning process", a formal definition of this is required as well.

\begin{definition}[Reasoning Process]
\label{def:reasoning_process}
A reasoning process is a computation represented as a 2-Tuple $(in, op)$, consisting of:

\begin{itemize}
    \item $in$: Input of the process
    \item $op$: A function operating on the inputs, yielding output $out$, with $op(in)~=~out$. In neural networks, $op$ is a compositional function consisting of multiple sub-operations, e.g. corresponding to network layers.
\end{itemize}

\end{definition}

As this thesis is concerned with the assessment of faithfulness of Natural Language Explanations (NLEs), we further define the concept of a faithful NLE: 

\begin{definition}[Faithful NLE]
\label{def:faithful_NLE}
A faithful NLE communicates, in natural language, the most important aspects of a model's reasoning process. Important aspects are those inputs or operations that determine the output most strongly. Further, the faithful NLE does not include irrelevant or contradictory information with respect to the model's reasoning process.
\end{definition}

Firstly, the communication of an explanation in natural language implies the need for it to be human-understandable, which is partly assessed by quality ratings provided by participants (e.g. the human-evaluated explanation score $S_E$ in the e-ViL benchmark). A further discussion of which features are "most important" is a question of relevance threshold that is excluded from the scope of this work.

According to the above Definition \ref{def:reasoning_process}, one class of important aspects that need to be communicated by a faithful NLE are the relevant inputs as they are part of the reasoning process. The proposed measures are relying only on determining the relevant inputs for predicting the answer and generating an explanation and are aimed at identifying a necessary, although not sufficient condition for a highly faithful explanation, similar to related work by Wiegreffe et al.~\cite{wiegreffe2021measuring} focusing on different properties of faithfulness.

There is a wide range of approaches that have been proposed in recent years for determining relevant features of models. Notably, the class of additive feature attribution methods \cite{lundberg2017unified} assign a continuous relevance value to each feature, corresponding to their importance in generating the model's prediction. As the correct identification of relevant input features is still an actively discussed research topic~\cite{ijcai2022-90}, we need to, for now, assume their sufficient reliability and precision for practical purposes:

\begin{assumption}[Relevance Attribution Measures]
\label{assumpt:relevance_measures}
Additive feature attribution methods (e.g. SHAP, LRP, Integrated Gradients) identify, reliably and precisely, which parts of input features are important for generating a model's prediction
\end{assumption}

As a faithful NLE $e$ is, per definition \ref{def:faithful_NLE}, required to represent the same reasoning process that led to the prediction of an answer $a$, it must contain the relevant inputs and reasoning operations within its linguistic contents. More formally, we assume a reasoning process $RP_a~=~(in_a, op_a)$, that is predicting the answer via $op_a(in_a)~=~a$. The faithful explanation $e$ contains the most relevant aspects of $RP_a$ as information in natural language. The explanation $e$ is generated by a separate reasoning process $RP_e~=~(in_e, op_e)$ via $op_e(in_e)~=~e$.
Evidently, $RP_a$ and $RP_e$ can have the same inputs ($in_a~=~in_e$), but at most similar operations leading to different outputs as one yields the answer while the other generates an explanation ($a~\neq~e$).

The reasoning processes $RP_a$ and $RP_e$ are realized by a respective explanation -module $m_e$ and a task-answering model $m_a$. Let us assume both models receive the same inputs ($in_a~=~in_e$). If we presume a model $m_e$ that is using the operations $op_e$ to produce a faithful explanation $e_i$ for each input $in$ from a large dataset $D$, then $m_e$ is required to model the behavior of $m_a$ to be able to express a faithful explanation of $m_a$'s reasoning process $RP_{a}$ for each input $in$. In simple terms, the explanation-module is required to \textit{model} the task-answering model in order to be able to consistently produce faithful explanations. 

Such a modeling process implies a similarity of the relevance distributions $r_{e}$ and $r_{a}$, as the following thought-experiment shows: On one hand, if the relevance distribution of $RP_{e}$ contains relevant inputs $r_{e}$ that are not a subset of $r_{a}$, than the explanation $e_i$ is produced from information that was not relevant for $RP_{a}$. Therefore, $m_e$ does not faithfully model $m_a$ in this instance and cannot reliably express the behavior of $m_a$ across all inputs $in \in D$. If, on the other hand, $r_{a}$ assigns a high relevance to inputs where $r_{e}$ does not, $m_e$ does not model an important aspect of the reasoning process $RP_{a}$ for this instance and cannot reliably include this information in producing $e$ across all inputs.
From the previous arguments, we can conclude that the relevance distributions $r_{e}$ and $r_{a}$ are similar across a large set of inputs if $m_e$ faithfully models $m_a$ to produce explanations. This is the central idea behind the proposed faithfulness measures and is formalized in the following statement:

\begin{implication}
[Similarity of Input Relevance Attribution Distributions]
\label{impl:similarity_inp_rel}
Two reasoning processes $RP_a$ and $RP_e$, realized by models $m_a$ and $m_e$, with equal inputs ($in_a~=~in_e$), exhibit similar feature relevance distributions on the inputs ($r_{a}~\sim~r_{e}$) across all possible inputs $i$, if $m_e$ is faithfully modelling $M_t$ to produce an explanation $e_i$ of $RP_{ai}$.
\end{implication}

It should be noted that the above implication is a property of a model $m_e$ faithfully representing $m_a$: If this is the case, $m_a$ and $m_e$ exhibit similar input relevance distributions. However, the property of similar relevance distributions is not a sufficient condition for a faithful model, as the input features could be assigned a similar relevance, even though $m_e$ does not model $m_a$. We can think of $m_e$ modelling $m_a$ as a similarity of their operations $op_e$ and $op_a$. If a reasoning processes $RP_e~=~(in_e, op_e)$, realized by $m_e$, and a reasoning process $RP_a~=~(in_a, op_a)$, realized by $m_a$, have the same inputs ($in_a~=~in_e$) and a similar relevance distribution $r_a~\sim~r_a$, we cannot conclude that their operations $op_a$ and $op_e$ are similar, as we have no further information about the processing of the inputs apart from their relevance. 

On the contrary, there are reasoning processes, which exhibit similar input relevance distributions but employ completely different operations.\footnote{Examples for this can be easily constructed. For instance, presume two models $m_x$ and $m_y$ which receive a 2-dimensional vector $i$ as inputs. $m_x$ outputs whichever value is contained in the second dimension of the vector, e.g. $m_x((12,3))~=~3$, whereas $m_x$ always outputs 1 if the value in the second dimension of $i$ is different from 0 and, in the other case, -1. Both models will have a feature relevance attribution over $i$ of $r_x~=~r_y~=(0.0,1.0)$, as the input is completely determined by the second dimension of $i$, but the two models carry out different computations.}

More generally, for reasoning processes x and y:
\begin{equation}
\forall x, y: (in_x = in_y) \land (op_x \sim op_y) \implies r_{x} \sim r_{y}
\end{equation}

But:

\begin{equation}
\exists x, y: (in_x = in_y) \land (r_{x} \sim r_{y}) \land (op_x \nsim op_y)
\end{equation}

In this sense, the measurement aims at the evaluation of a necessary, but not sufficient condition of faithfulness. However, we argue that across a large number of complex inputs and operations, a consistent similarity of $r_a~\sim~r_e$ is a plausible basis for assuming an explanation-module $m_e$ to faithfully represent $m_a$. This is especially the case when considering complex tasks such as visual question answering for $m_a$ and NLE generation for $m_e$. In these cases it is not obvious how a counterexample can be constructed in which $m_e$ creates explanations for every answer $a$, features a highly similar relevance attribution as $m_a$ ($r_a~\sim~r_e$), but still does not faithfully model $m_a$. We, therefore, formulate the following conjecture:

\begin{conjecture}
[Relevance Similarity Implies Faithful Modeling]
\label{conj:faithful-modeling}

Given a task model $m_a$, predicting an answer $a$, and an explanation model $m_e$, generating an explanation $e$ on how $a$ was predicted, we can assume that if and only if the relevance distributions are similar across a large dataset of non-trivial tasks ($r_a~\sim~r_e$), $m_e$ is faithfully modelling $m_a$.

\end{conjecture}

Grounded on implication \ref{impl:similarity_inp_rel} and conjecture \ref{conj:faithful-modeling}, and acknowledging the limitation mentioned above, the faithfulness measures proposed in this work are based on comparing the attributed relevance distributions of $m_a$ and $m_e$ directly (Section~\ref{subsec:cosine-sim-rel-attr}) and on the assessment of the influence of important features of the explanation $e$ on the prediction of the task-answer $a$ (Section~\ref{subsec:sufficiency-comprehensiveness}).

\section{Attribution-Similarity Faithfulness Metric}
\label{subsec:cosine-sim-rel-attr}

As a straightforward derivation of Implication \ref{impl:similarity_inp_rel} discussed in the previous section, the Attribution-Similarity faithfulness measure is directly comparing the relevance vectors $r_{a}$ and $r_{e}$ using the cosine function:

\begin{definition}[Attribution-Similarity]
\label{def:faithfulness-measure}
The degree of faithfulness $F$ of a Natural Language Explanation $e$, generated by an explanation -module $m_e$, explaining a model's task answer $a$ predicted by $m_a$, is approximated by

\begin{equation*}
F_{attr-sim} = cos(r_a, r_e),
\end{equation*}

with $r_a$ and $r_e$ representing the feature relevance distribution on the input $in$ that is shared by $m_a$ and $m_e$ to predict $a$ and $e$, respectively.
\end{definition}
A similar approach was previously proposed by Wu and Mooney, who proposed using GradCAM to compare the attribution on the visual modality of the explanation generation and task-label prediction of their model~\cite{wu2018faithful}. In contrast to their approach, we are comparing the answer attribution of both the visual and the linguistic modality and use Integrated Gradients as the attribution method due to GradCAMs limitation when computing the gradient of saturated activation functions (See Section~\ref{sec:ig-feat-attr}). 
Using the cosine function, as opposed to the dot-product, ignores the magnitude of the relevance scores. This is a desired effect, as it makes the score invariant to the total amount of relevance that is attributed to the explanation versus the answer prediction, which can vary depending on the additive feature importance measure used or the implementation specifics of the model. Further, it allows us to sum up the attributions per generated token for autoregressive language generation models, as described in the following paragraphs.

\subsubsection{Implementation and Integration into the e-ViL Benchmark}\label{subsubsec:feat-rel-similarity}

Based on the \textit{Attribution-Similarity} metric, we propose a faithfulness score for the vision-language tasks of the e-ViL benchmark as an extension to the already existing scores. To calculate the additive feature relevance attribution for answers and explanations on the benchmark, we use the Integrated Gradients~\cite{sundararajan2017axiomatic} implementation of the library captum~\cite{kokhlikyan2020captum}. We first calculate the cosine similarity of relevance vectors for both modalities of the vision-language models separately.

To compare the text relevance values for explanation and answer models, the relevance values need to be mapped to common tokens. Some models, such as e-UG, employ different tokenizers for their task and explanation -module. Therefore, the relevances are mapped to words to be able to compare them using cosine similarity. As an example, we consider the tokenization of the following sentence: ``I sink under the weight of the splendour of these visions!". A WordPiece encoder, used in most BERT models, would yield the following tokenization: 

\begin{verbatim}
    ['i', 'sink', 'under', 'the', 'weight', 'of', 'the', 's', 
    '##ple', '##ndo', '##ur', 'of', 'these', 'visions', '!']
\end{verbatim}

On the other hand, a Byte-Pair-Encoding tokenizer, used in GPT2 models, will split the string differently:

\begin{verbatim}
    ['I', 'sink', 'under', 'the', 'weight', 'of', 'the', 'splend', 
    '##our', 'of', 'these', 'visions', '!']
\end{verbatim}

Since, in the above example, the relevance attributed to the rare word ``splendour" is distributed onto three tokens for the WordPiece tokenizer, while the Byte-Pair-Encoding tokenizer split the word into only two tokens, the relevance attribution is aggregated for each word to enable a comparison between the attributions of each word.

Similarly, a common unit for feature attribution needs to be found for images. For models operating directly on raw input images, pixels or pixel clusters (e.g. superpixels) can be used. In the case of e-UG, the explanation generation model is supplied with features from the task-module. Here, the relevance can be attributed to the input of the task-module in relation to the explanation output.

Generative language models producing a natural language explanation, such as the GPT-2 model used in e-UG, are generating one token per forward pass, appending the new token to the previously generated sequence when used for text generation (autoregression). Each time, a probability distribution for the next token is predicted, from which one token is selected according to a specific decoding strategy, such as greedy selection or beam search. The selected token is then appended to the input of the model in the next forward pass, iteratively generating one token at a time, until the desired text length or a specific end-token is reached. The selection of a specific token for appending it in the subsequent iteration is a non-differentiable operation. As feature attribution using IG requires a fully-differentiable model (see Section \ref{sec:ig-feat-attr}), the relevance values for the generated text cannot be calculated in a single pass. Instead, relevance values are computed for each generated token. The relevance of tokens that have been generated in previous passes is then disregarded, and the relevance values of the original input tokens are summed up for each generated token, following previous work~\cite{he-etal-2019-towards, ding-etal-2019-saliency}.

Once the cosine similarity is calculated for both modalities, the resulting values are averaged and normalized to a $0 - 1$ range to integrate it with the explanation score $S_E$ and the task score $S_T$:

\begin{definition}[Faithfulness Score $S_F$ (Attribution-Similarity)]
The Attribution-Similarity faithfulness score $S_F$ for feature relevances $r_{te}$ and $r_{ie}$ for text and image relevances of an explanation, and $r_{ta}$ and $r_{ia}$ for a corresponding task-answer is defined as:
\begin{equation*}
    S_F~=~\frac{0.5 * (1 + cos(r_{ta}, r_{te})) + 0.5 * (1 + cos(r_{ia}, r_{ie}))}{2}
\end{equation*}
\end{definition}

\section{NLE Feature Relevance Sufficiency and Comprehensiveness}\label{subsec:sufficiency-comprehensiveness}

In addition to the Attribution-Similarity score $S_F$, we propose sufficiency and comprehensiveness metrics adapted to natural-language explanations intended for a more detailed analysis of the explanation faithfulness.

The concepts of sufficiency and comprehensiveness have been previously used to evaluate the faithfulness of rationalized models in natural language processing (NLP). In this context, rationales are not natural language explanations of a model's decision but rather the selection of important input tokens according to their importance with respect to the model's decision. Recently, and following prior work proposing the use of sufficiency~\cite{zaidan2007using} and comprehensiveness~\cite{yu2019rethinking}, the ERASER benchmark~\cite{deyoung2020eraser} proposed the usage of both concepts across a multitude of NLP datasets that had been extended with target rationales supplied by human annotators.

The comprehensiveness of a rationale is the degree to which the prediction is influenced when removing the input tokens $r_i$ that have been marked as relevant according to the rationale. For a model prediction $m(x_i)_j$ of with respect to an input $x_i$ and the output class $j$, the comprehensiveness is calculated as:

\begin{equation}
   \text{comprehensiveness} = m(x_i)_j - m(x_i \setminus r_i)_j
\end{equation}

As noted by Zaidan et al.~\cite{zaidan2007using}, this calculation can also be viewed as creating a \textit{contrastive example} $\Tilde{x}_i$ for the original datapoint $x_i$ and evaluating the model's prediction on this contrastive input. Constituting one desideratum of a faithful rationale, the set of relevant tokens $r_i$ should encompass all input features that are significantly leading to the prediction of class $j$. When removing these features, the model's prediction confidence should be significantly lower for class $j$. Therefore, the comprehensiveness score is higher for a more faithful rationale.

Analogously, the sufficiency of a rationale is defined as the change in the model's prediction with respect to class $j$ when using \textit{only} the relevant tokens $r_i$:

\begin{equation}
   \text{sufficiency} = m(x_i)_j - m(r_i)_j
\end{equation}

\begin{figure}
    \centering
    \includegraphics[width=\textwidth]{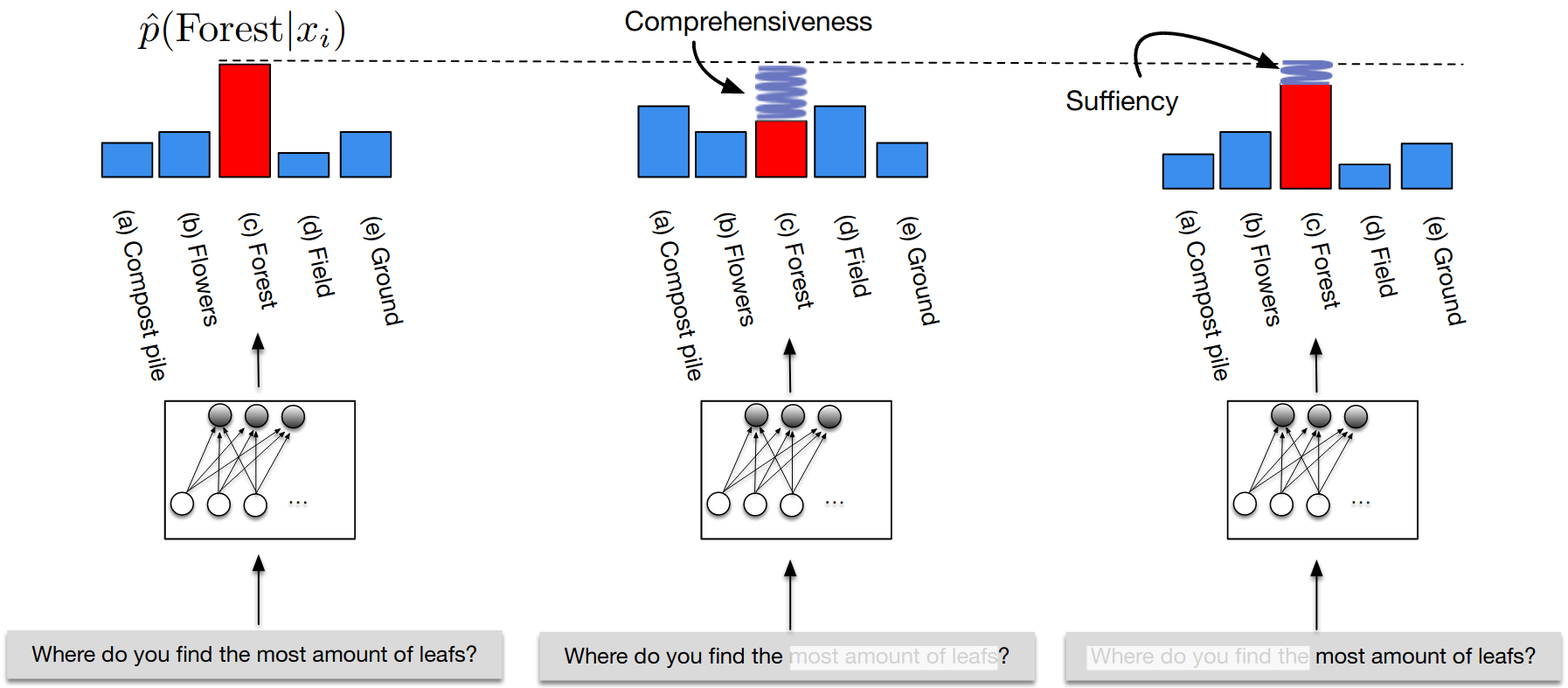}
    \caption{Illustration of comprehensiveness and sufficiency for an NLP question-answering model. For comprehensiveness, removing the relevant tokens yields a comparatively high decrease in the prediction confidence for the class "Forest", while the sufficiency of a faithful rationale is characterized by a low decrease. Figure by DeYoung et al.~\cite{deyoung2020eraser}.}
    \label{fig:comp_suff_nlp}
\end{figure}

The idea behind this desideratum is that the set of relevant tokens $r_i$ should be \textit{sufficient} for the prediction of class $j$ for a faithful rationale. Consequently, a set of relevant inputs $r_i$ lead to a change in prediction confidence that is lower for more faithful rationales. An illustration of the two measures is provided in ~\autoref{fig:comp_suff_nlp}.

The above definitions assume a discrete selection of relevant tokens $r_i$. In contrast, many methods employed to generate rationales, such as additive feature attribution methods or soft attention weights, assign a continuous relevance value to each input. To adapt the measures to continuous relevance weights, a threshold $k$ can be determined, where tokens with a relevance weight above $k$ are considered as included in the rationale $r_i$. DeYoung et al.~\cite{deyoung2020eraser} suggested calculating the sufficiency and comprehensive as a function of $k$, calculating the mean using different thresholds to calculate an aggregate measure, termed \textit{Area Over the Perturbation Curve} (AOPC). Using $K$ bins of tokens according to certain thresholds, such as grouping the tokens in the top $k = (1\%, 5\%, 10\%, 20\%, 50\%)$, depending on their relevance, we can calculate the aggregate comprehensiveness in the following way:

\begin{equation}
    \label{eq:aopc_comp}
   \text{AOPC comprehensiveness} = \frac{1}{K} \sum _{k=0} ^{K} m(x_i)_j - m(x_i \setminus r_{ik})_j
\end{equation}

The aggregate sufficiency (AOPC) is calculated analogously:

\begin{equation}
    \label{eq:aopc_suff}
   \text{AOPC sufficiency} = \frac{1}{K} \sum _{k=0} ^{K} m(x_i)_j - m(r_{ik})_j
\end{equation}

\subsubsection{Adapting Comprehensiveness and Sufficiency to Natural Language Explanations}

The underlying idea of sufficiency and comprehensiveness of explanations can also be extended to quantify the faithfulness of a natural language explanation for a prediction. We propose to use the set of relevant features $r_{ei}$ for the generation of the natural-language explanation $e_i$ of a corresponding prediction of the answer model $m_a(x_i)_j$. Using the set of relevant features $r_{ei}$ as the rationale set of features $r_i$, we can calculate the AOPC for comprehensiveness and sufficiency according to \autoref{eq:aopc_comp} and \autoref{eq:aopc_suff}:

\begin{equation}
    \label{eq:nle_comp}
   \text{NLE-comprehensiveness} = \frac{1}{K} \sum _{k=0} ^{K} m_a(x_i)_j - m_a(x_i \setminus r_{eik})_j
\end{equation}

\begin{equation}
    \label{eq:nle_suff}
   \text{NLE-sufficiency} = \frac{1}{K} \sum _{k=0} ^{K} m_a(x_i)_j - m_a(r_{eik})_j
\end{equation}

Intuitively, the above NLE-comprehensiveness measure indicates in how far a removal of the features that did not significantly influence the generation of the explanation leads to lower confidence of the predicted answer class $j$. Similarly to the conventional use of comprehensiveness, a high value is desired, since this implies that the explanation -module primarily uses all the features that are also contributing to the predicted answer class.
On the other hand, NLE-sufficiency gives an indication of how far using only those features that were relevant for generating the NLE is sufficient to predict the answer class $j$ in the answer model. Analogously, a low NLE-sufficiency is desired, as this means that the explanation -module selected the most important features for predicting the answer class $j$.

While a high comprehensiveness and low sufficiency value indicate a faithful NLE in relation to the input features, a low comprehensiveness and high sufficiency value conversely point to a more unfaithful model. Nevertheless, both measures are not necessarily correlated: An NLE can score high on comprehensiveness and sufficiency, which is an indication for an explanation -module that selected a subset of features that, while being critical for the predicted class $j$, were not sufficient to predict $j$ when taken individually. This can also be interpreted as an NLE model that is selecting important features to generate the rationale but does not take into account other features that give critical context for the selected features to make sense and lead to the prediction of $j$.
Contrarily, low comprehensiveness and low sufficiency indicate that the relevant features that led to the prediction of the NLE are sufficient to predict $j$ with high confidence. However, the low comprehensiveness indicates that there are other features in the input that were not taken into account by the explanation -module, but that would also lead to the prediction of class $j$. In this case, the explanation -module would base its prediction on a sufficient subset of relevant features, but not on the comprehensive set of all relevant features.

\subsubsection{Implementation}

To assess the NLE-comprehensiveness and -sufficiency for the vision-language tasks investigated in this thesis, we assess both measures separately for the vision and language modality. The input features are grouped by selecting the top $k~=~(1\%,\\5\%,\,10\%,\,20\%,\,50\%)$ features according to the relevance weights of the explanation generation model. The NLE-relevance-comprehensiveness and -sufficiency measures are then computed according to \autoref{eq:nle_comp} and \autoref{eq:nle_suff}.

As described in Subsection~\ref{subsec:cosine-sim-rel-attr}, the attributed relevance weights for the language modality need to be mapped to words as a common feature unit to apply the measure to models that use different tokenizers for their answer and explanation -modules. For the visual modality, both measures are applied to the individual input features of the visual input directly, without grouping features. The removal of visual features is implemented by replacing the respective features with a zero value while replacing the words that are to be removed with a "[PAD]" token for the language input. The resulting prediction from the task-answer classifier is processed by a softmax function to normalize the classification output.

The relevance weights attributed to both input modalities were computed using Integrated Gradients~\cite{sundararajan2017axiomatic}, according to the process outlined in Subsection~\ref{subsec:cosine-sim-rel-attr}.

\section{eUG Ablation Models}
\label{sec:ablation-models}

\begin{figure}[hb]
    \centering
    \includegraphics[width=0.5\textwidth]{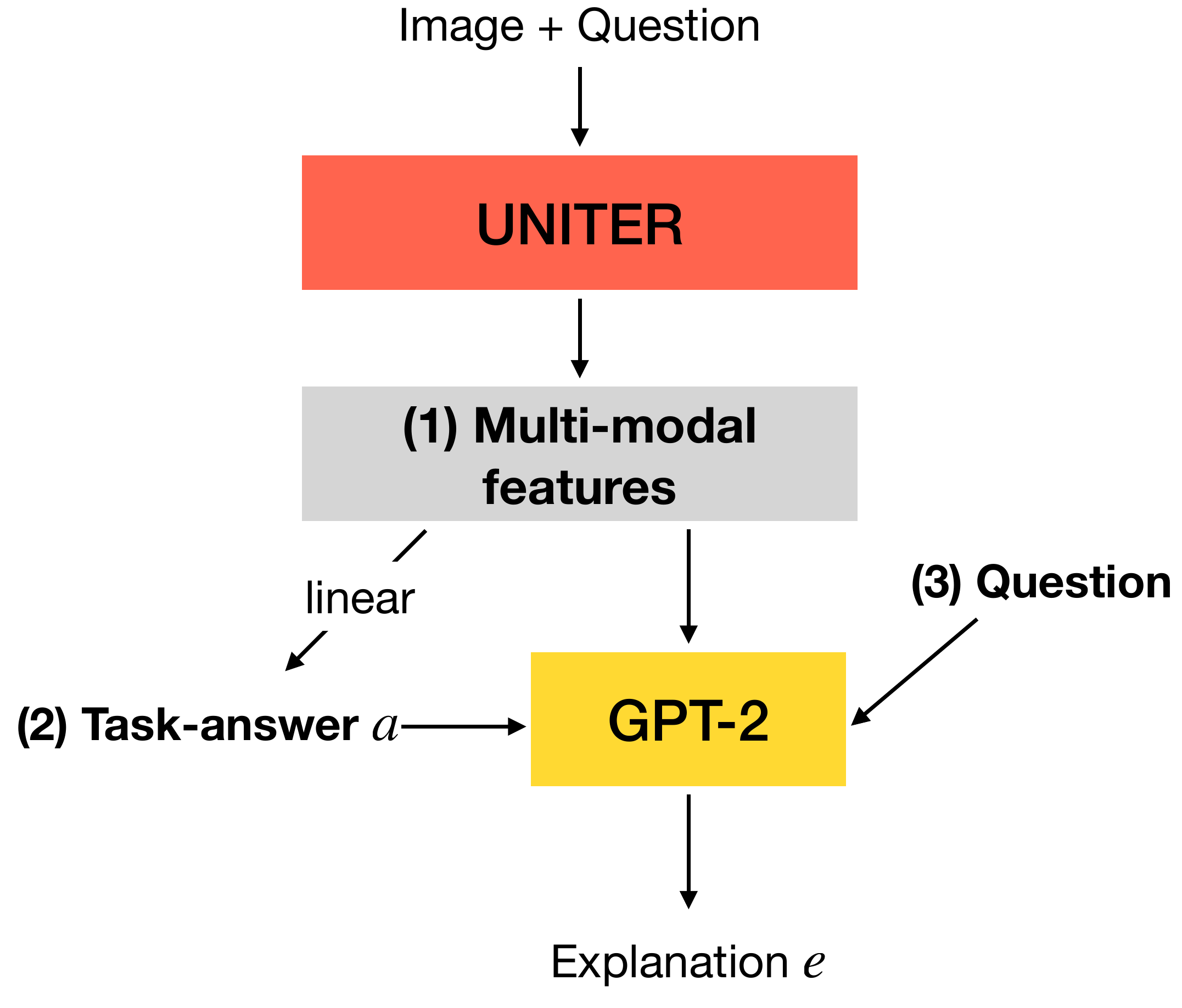}
    \caption{Inputs to the GPT-2 explanation generator of the e-UG model. The GPT-2 explanation module generates the explanation $e$ by receiving (1) Features extracted from the task-module, (2) the predicted answer $a$ as text input (e.g. "yes", "elephant", etc.), and (3) The initial question as text input. We are ablating the e-UG model by removing subsets of inputs to the GPT-2 explanation generator.}
    \label{fig:gpt-2_inputs_ablation}
\end{figure}

The e-UG model consists of task-module, in the form of an UNITER vision-language transformer processing the vision and language inputs and predicting the answer, as well as an explanation generator model, a GPT-2 generative language model, to generate a natural language explanation (NLE) based on the prediction of the task-module and the task inputs. The e-UG model was proposed originally with the following inputs to the explanation generation model: (1) Features extracted from the task-module, (2) the predicted answer as text input (e.g. "yes", "elephant", etc.), (3) The initial question as text input (see \autoref{fig:gpt-2_inputs_ablation}). By ablating individual inputs of the explanation generation model, we observe the effects on task performance, explanation metrics, and faithfulness. 

All models are trained until no improvement on the validation dataset for the combined task and explanation score was registered for three epochs. We then selected the model with the best global performance when considering both task and explanation scores and used this model for the evaluation of task, explanation, and faithfulness metrics. In the following, the different ablation models are described.

\subsubsection{No-additional-question (NQ)}
As the language input, e.g. the question of a VQA task, was already processed by the task-module and influences the features that are processed by the explanation generation module, supplying the question a second time to the GPT-2 explanation module is redundant. We are therefore removing the additional question under the assumption that the explanation -module generates more faithful explanations if it has only access to the question features processed and pre-selected by the UNITER task-module.

\subsubsection{No-answer (NA)}
Similarly to the NQ ablation model, the answer input in clear text is redundant, as information about the answer is encoded in the feature vector. The multi-modal features are only processed by a single linear output layer which translates the output to the number of classes, in order to yield logits for selecting the task answer $a$. Therefore, it is reasonable to assume that the representation of the task answer can be easily extracted from the multimodal feature vector by the explanation -module. Analogous to the NQ model, we hypothesize that removing the redundant input increases the faithfulness of the explanations.

\subsubsection{Only-uniter-features (OU)}
The Only-uniter-features variant is a combination of the NQ and NA models and processes only the multi-modal features from the UNITER model to produce the explanation. Following the rationale from the previous subsections, we hypothesize this model to generate explanations that are most faithful in comparison to the other variants.

\subsubsection{No-uniter-features (NU)}
Inversely to the OU model, using no UNITER features at all removes all access to the reasoning of the task-module from the explanation -module. Therefore, the explanation is generated purely as a post-hoc rationale, with no further information about the reasoning process. Consequently, we would expect low faithfulness scores for this model. As the NU variant has only access to linguistic input, the faithfulness measures with respect to the input image cannot be computed.

\subsubsection{Answer-only baseline (OA)}
As a baseline for the automatic explanation metrics, we are evaluating a model that only receives the task answer as an input and generates rationales without access to any task inputs. As no common inputs are processed by this variant, faithfulness measures cannot be calculated for this model.

\cleardoublepage

\chapter{Results}\label{results}

In the following sections, we state our research hypotheses and present the results of evaluating the previously described variations of the e-UG model on the e-ViL benchmark datasets, by assessing task accuracy, automatic explanation metrics, and faithfulness scores. The chapter is concluded with a general discussion of the results.

\section{Hypotheses}

Our evaluation is focused on the impact of ablating inputs to the explanation generation module on explanation quality and faithfulness results. As described in Section~\ref{sec:ablation-models}, we expect that the removal of inputs that are redundant to the task-module features leads to higher faithfulness scores. In addition, we argue that the models with deleted redundant inputs will be able to produce explanations that are of similar quality compared to the original model, as the explanation generation module should be able to learn the extraction of the relevant information from the task-module features. Further, we expect the three proposed faithfulness measures to agree on the relative change in terms of faithfulness with the changes to the model architecture.

Consequently, we pose the following main hypotheses:

\begin{itemize}
    \item[(H1)] The ablation models with removed redundant features achieve similar results on the automatic explanation quality metrics compared to the default e-UG configuration.
    \item[(H2)] Models with fewer redundant inputs are more faithful according to the employed metrics. The model without any redundant input (OU) is the most faithful, while the model not using any task-module features (NU) is the least faithful, according to the metrics.
    \item[(H3)] The three faithfulness metrics \textit{Attribution-Similarity}, \textit{NLE-Sufficiency} and \textit{NLE-Comprehensiveness} are in agreement on the relative faithfulness of the evaluated models.
\end{itemize}

\section{Reproduction of Original Results}

\begin{table}[th]
    \centering
    \begin{tabular}{lcc}
        \toprule
        Dataset & $S_T$ original & $S_T$ reproduced 
        \\
        \midrule
        e-SNLI-VE & 79.5 & 79.15 \\
        VQA-X & 80.5 & 77.5 \\
        VCR & 69.8 & 35.1 \\
        \bottomrule
    \end{tabular}
    \caption{Reported and reproduced task score results for e-UG.}
    \label{tab:eUG_task_score}
\end{table}

As the basis for investigating the effects of ablating the inputs for the explanation generator, we trained the e-UG model on the three datasets of the e-ViL benchmark according to the hyperparameters reported by the original authors Kayser et al.~\cite{kayser2021vil}. We then evaluated the task accuracy for the predicted answers, as well as the automatic evaluation metrics for the explanation score of the model (Table \ref{tab:eUG_task_score}). For the task score $S_T$, which corresponds to the accuracy of the predicted VQA labels in the test splits of the three datasets, we found the performance to be slightly lower in the dataset e-SNLI-VE, considerably lower in VQA-X (3 percent), and only half of the reported accuracy in VCR. The results were achieved using the default seed provided in the e-UG parameters and with the hyperparameters reported in the original paper. We used the VCR-pretrained UNITER model for VCR as stated in the readme of the e-ViL repository\footnote{\url{https://github.com/maximek3/e-ViL}}. Considering the inadequate performance of the reproduced model on VCR, we excluded the dataset from our further analyses.

\begin{table}[tbh]
    \centering
    \begin{tabular}{lp{2.4cm}p{2.6cm}p{2.4cm}p{2.6cm}}
    \toprule
        Dataset & METEOR original & METEOR \newline reproduced & BERTScore original & BERTScore reproduced 
        \\
        \midrule
        e-SNLI-VE & 19.6 & 19.24 (-1.83\%) & 81.7 & 81.56 (-0.17\%) \\
        VQA-X & 22.1 & 20.56 (-6.96\%)& 87.0 & 86.37 (-0.72\%)\\
        \bottomrule
    \end{tabular}
    \caption{Reported and reproduced automatic explanation metric results for e-UG. In parentheses: relative change in percent.}
    \label{tab:eUG_expl_score_repro}
\end{table}

For reproducing the results on the persuasiveness of the generated explanations, we are considering the automatic $S_E$ metrics METEOR and BERTSCORE (see Section \ref{subsec:results_expl}).
While results for e-SNLI-VE are similar to the originally reported and reproduced results, metrics on VQA-X are lower by around 1-2\% of the absolute score (see \autoref{tab:eUG_expl_score_repro}), with a relative METEOR score difference of 6.96\% and 0.72\% for BERTScore.

\paragraph{Discussion.}

Plausible reasons for the diminished performance in task and explanation score include the possibility that the original results were achieved using a different seed for the initialization of the weights, using incorrectly reported hyper-parameters, or with a different method of choosing the evaluation model from the training checkpoints. Upon contacting the original authors regarding the inadequate results on VCR, the first author suggested that the result could  be caused by the possibly incorrectly documented hyperparameter ``max\_sequence\_length" that might be set too low for the VCR dataset, which appends the possible answers to the input sequence\footnote{See also: \url{https://github.com/maximek3/e-ViL/issues/8}}. Raising the maximum sequence length could also have an effect on task performance in the e-SNLI-VE and VQA-X datasets. At the time of writing the thesis, this presumption could not yet be validated. Due to the high computational resources involved with training the models, we were unable to perform comprehensive hyperparameter optimization. 

Under the assumption that the results are caused by differences in weight initialization, results on the metrics for e-SNLI-VE and VQA-X that fall into the range between the reproduced and reported results could be interpreted as examples of model performance variation caused by initialization and training procedure, rather than significant outcomes that inform about the performance differences between model architectures. This assumption is supported by the observed variation in task-score results when evaluating the ablation models in \autoref{sec:task-score}, which is discussed in more detail in the following section. 

\section{Task Score}
\label{sec:task-score}

The task score $S_T$ corresponds to the accuracy of the predicted VQA labels on the test splits of the e-ViL datasets and is presented in \autoref{tab:eUG-task-score}. All results for e-SNLI-VE lie within 1\% of absolute difference, while the variation in the VQA-X dataset is larger with a maximum of 3\% total difference.

\begin{table}[h!]
    \centering
    \begin{tabular}{lcc}
    \toprule
        Model & e-SNLI-VE & VQA-X 
        \\
        \midrule
        e-UG (reported) & 79.5 & 80.5 \\
        \midrule
        e-UG (reproduced) & 79.15 & 77.50 \\
        No-additional-question (NQ) & 79.17 & 80.04 \\
        No-answer (NA) & 78.95 & 80.04 \\
        Only-uniter-features (OU) & 78.89 & 79.53 \\
        No-uniter-features (NU) & 79.10 & 79.48 \\
        Answer-only (OA) & 78.88 & 80.17 \\
        \bottomrule
    \end{tabular}
    \caption{Task score $S_T$ results for e-UG and ablation models.}
    \label{tab:eUG-task-score}
\end{table}

\paragraph{Discussion.}
The similarity between the scores is expected, as the ablation models do not change the architecture of the task-module that is producing the task-answer. However, if features from the task-module are used or even required for generating an explanation (default, NQ, NA, OU models), the model is required to learn common representations that contain information for generating the task-answer as well as an explanation. Kayser et al.~\cite{kayser2021vil} argued that this setup can be viewed as an instance of multi-task learning, where the explanations could serve as additional learning instructions during the training of the task-module. Kayser et al. acquired small improvements of 0.5\% accuracy on VQA-X and VCR, and a similar result on e-SNLI-VE with a 0.1\% accuracy improvement. At least for the VQA-X dataset, our results suggest that the differences noted by the authors are likely not caused by a facilitated training due to a multi-task learning setting, but rather by random variation stemming from differences in the training process. Likewise, the general variation of 1-3\% task accuracy on e-SNLI-VE and VQA-X indicates that the 0.5\% improvement for joint training on VCR is not a significant result.

\section{Explanation Score}\label{subsec:results_expl}
The e-ViL benchmark requires an empirical evaluation of the explanations using participants. Considering the focus of this work on explanation faithfulness, we opted for employing the automatic evaluation metrics proposed in the original benchmark as a proxy for the explanation score. Analogously to the previous section, we are reporting the automatic $S_E$ metrics METEOR and BERTScore, which have the highest correlation to human judgement according to previous results~\cite{kayser2021vil}. The METEOR metric was originally developed for machine translation as a combination of unigram-precision and -recall, as well as alignment measures and further features such as stemming and synonymy matching~\cite{banerjee2005meteor}. BERTScore is a text-generation metric based on the cosine-similarity of per-token contextual embeddings extracted from the candidate and target sentences using a trained BERT language model~\cite{zhang2019bertscore}.

\begin{figure}
    \centering
    \includegraphics[width=\textwidth]{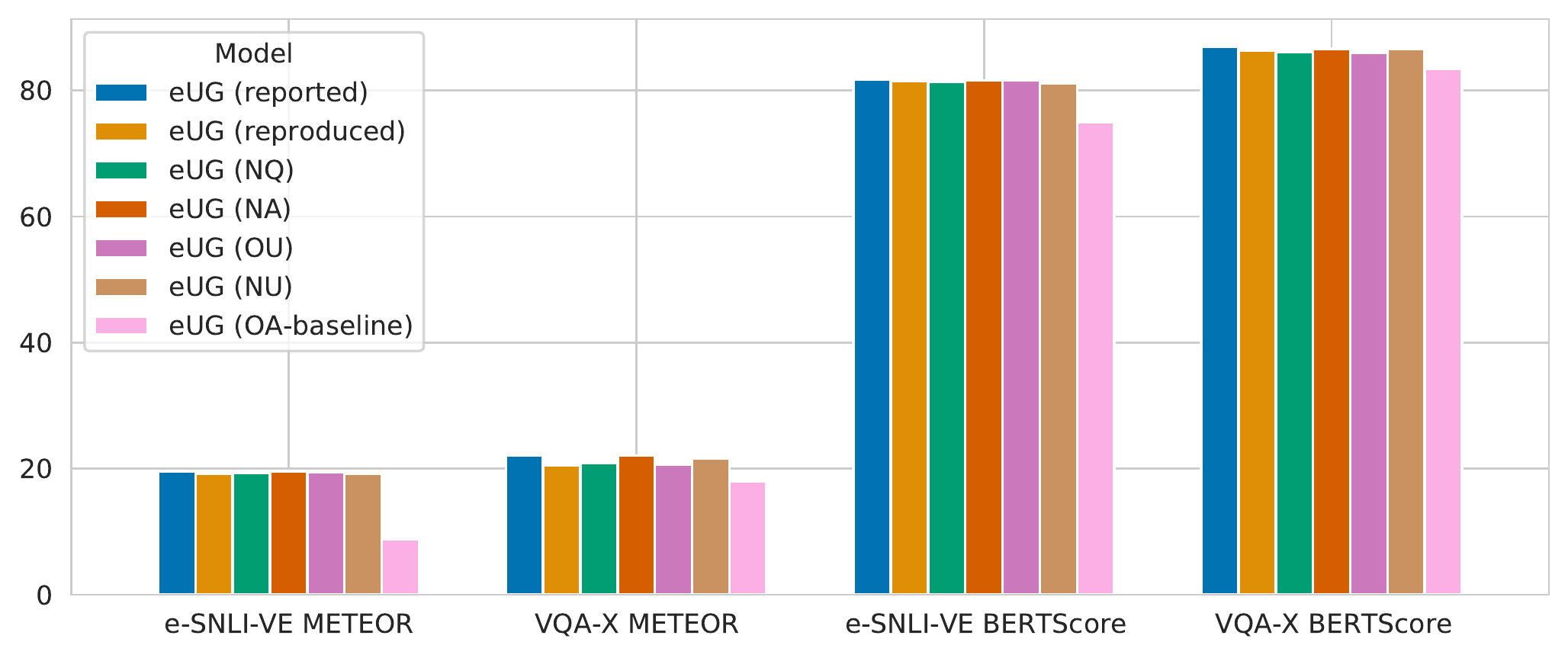}
    \caption{METEOR and BERTScore explanation generation metrics grouped by dataset and metric to visualize the differences between the datastets.}
    \label{fig:SE-barplot-perscore}
\end{figure}

The automatic explanation metric results are presented in \autoref{tab:eUG-expl-score} and visualized in \autoref{fig:SE-barplot-perscore}. All explanation scores assessed from our reproduced training of the e-UG model and its ablation versions are consistently lower than the reported scores from Kayser et al~\cite{kayser2021vil}, with the exception of the VQA-X METEOR score of the no-answer (NA) derivative reaching an equal score. All scores of the ablation models with the exception of the only-answer baseline (OA) lie within 1 to 2 percentage points on the metric when compared to the reproduced default model. The OA model achieves 81.32\% of the eUG (reported) performance in VQA-X METEOR and 95.13\% in VQA-X BERTScore, while the relative OA baseline performance on e-SNLI-VE is considerably lower with only 44\% METEOR score and 91.82\% BERTScore.

Further, two qualitative examples for the explanation generation capability of each model on the e-SNLI-VE dataset are presented in \autoref{tab:qulitative-explanation-generation}. From our observations on dataset examples, the default e-UG, NQ, NA and OU models produce explanations of comparable quality, while the NU model introduces information into the rationale that is not grounded in the image or hypothesis, which is not reflected in the automatic explanation metrics results. The OA baseline learns a simple rationale for each answer class, as no further information is supplied.

\begin{table}[]
    \centering
    \begin{adjustbox}{max width=\textwidth}
    \begin{tabular}{lp{2.3cm}p{2.3cm}p{2.2cm}p{2.3cm}}
        \toprule
        Model & e-SNLI-VE METEOR & e-SNLI-VE BERTScore & VQA-X METEOR & VQA-X BERTScore
        \\
        \midrule
        eUG (reported) & 19.6 & 81.7 & 22.1 & 87.0 \\
        \midrule
        eUG (reproduced) & 19.24 & 81.56 & 20.56 & 86.37 \\
        eUG (NQ) & 19.33 & 81.43 & 20.93 & 86.07 \\
        eUG (NA) & \textbf{19.58} & 81.58 & \textbf{22.10} & \textbf{86.64} \\
        eUG (OU) & 19.39 & \textbf{81.61} & 20.66 & 85.95 \\
        eUG (NU) & 19.15 & 81.19 & 21.63 & 86.61 \\
        eUG (baseline - OA) &  8.75 & 75.02 & 17.97 & 83.43 \\
        \bottomrule
    \end{tabular}
    \end{adjustbox}
    \caption{Automatic explanation metric results for e-UG. Best results per dataset and score are in bold text.}
    \label{tab:eUG-expl-score}
\end{table}

\begin{table}[ht]
\begin{minipage}[b!]{0.29\linewidth}
    \centering
    \includegraphics[width=4cm]{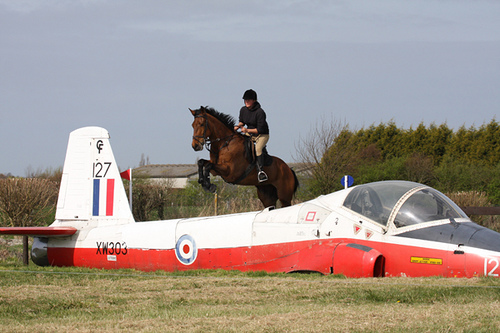}
    \scriptsize
    Hypothesis: A person is at a diner, ordering an omelette.
\end{minipage}
\hfill
\begin{varwidth}[b]{0.69\linewidth}
    \centering
    \scriptsize
    \begin{tabular}{lp{2cm}p{5.5cm}}
    \toprule
    Model & Answer & Explanation \\
    \midrule
    e-UG & contradiction & A person cannot be riding a horse and ordering an omelette at the same time. \\
    NQ & contradiction & A person cannot be riding a horse and ordering an omelet simultaneously. \\
    NA & contradiction & A person cannot be at a diner and riding a horse at the same time. \\
    OU & contradiction & A person cannot be at a diner and riding a horse at the same time. \\
    NU & contradiction & A person cannot be at a diner and at a restaurant at the same time. \\
    OA & contradiction & A man cannot be sitting and standing at the same time. \\   
    \bottomrule
    \end{tabular}
    \label{tab:my_label}
\end{varwidth}%

\begin{minipage}[b!]{0.29\linewidth}
    \centering
    \includegraphics[width=4cm]{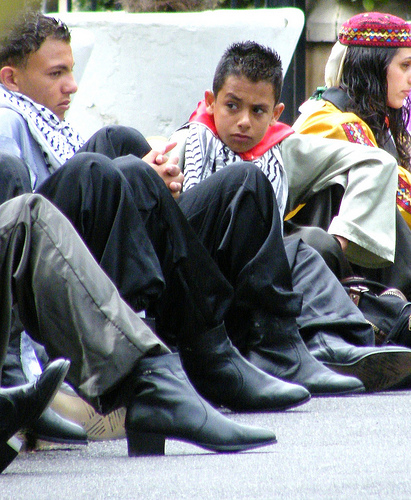}
    \scriptsize
    Hypothesis: there is a group of people waiting outside
\end{minipage}
\hfill
\begin{varwidth}[b]{0.69\linewidth}
    \centering
    \scriptsize
    \begin{tabular}{lp{2cm}p{5.5cm}}
    \toprule
    Model & Answer & Explanation \\
    \midrule
    e-UG & entailment & A group of people are waiting outside on the sidewalk. \\
    NQ & entailment & A group of people are sitting on the sidewalk waiting for the bus. \\
    NA & neutral & - \\
    OU & entailment & A group of people are waiting outside. \\
    NU & entailment & A group of people are waiting outside in a park. \\
    OA & entailment & A man is a person. \\   
    \bottomrule
    \end{tabular}
    \label{tab:my_label}
\end{varwidth}%

\caption{Qualitative examples of explanations generated from the different models on the e-SNLI-VE dataset. Explanations are only generated if the task-answer corresponds to ground-truth.}
\label{tab:qulitative-explanation-generation}
\end{table}

\paragraph{Discussion.}

Overall, the scores of all ablation models with the exception of the answer-only (OA) baseline are similar. As the automatic explanation metrics are only \textit{correlated} with human judgment, insights about the explanation generation performance based on the interpretation of small differences between individual models and datasets are limited. Therefore, from the perspective of the METEOR and BERTScore metrics, there is no significant difference in explanation performance between the models on the two datasets. A preliminary subjective sighting of explanation generation examples indicates that the quality of the ablation models that receive task-model features (e-UG, NQ, NA, OU) is comparable, while the NU model introduces new information that is not grounded in the images or questions, which is undesirable behavior. As expected, the NA baseline simply learns an approximation to rationales based on the answer classes.
Acknowledging that further validation by human explanation rating is required to support the findings, the presented results indicate that additional inputs of the question and answer in plain text to the explanation model are in fact redundant, as explanations can be generated from the UNITER task-module features alone -- supporting hypothesis~(H1). Furthermore, explanation generation is also learned if no features from the task model are supplied. In this case, the generated explanations are rationalizations of the predicted answer, without any access to the reasoning process leading to this prediction, and contain inconsistent information. From the point of view of the automatic explanation metrics, these rationalizations are indistinguishable from the explanation based on task-model features, stressing the importance of human explanation ratings and other methods to assess the rationale faithfulness.

Investigating the differences between the VQA-X and e-SNLI-VE datasets reveals a notable difference in OA baseline performance between the two, especially on the METEOR metric. This can be explained by the difference in answer classes: The VQA-X dataset contains 3,129 possible answer classes, while the entailment tasks of e-SNLI-VE only have three possible answers (\textit{entailment, contradiction, neutral}). Therefore, the answer alone contains significantly more information about the expected explanation on the VQA-X dataset.

\section{Faithfulness}
\subsection{Attribution Similarity ($S_F$)}
\label{subsec:fscore-means}

\begin{table}[]
    \centering
    \begin{adjustbox}{max width=\textwidth}
    \begin{tabular}{llccccc}
    \toprule
        Dataset &   Modality   & e-UG (default)   & NQ              & NA     & OU     & NU
        \\
        \midrule
                    & Vision   & 0.5206          & 0.4140          & \textbf{0.5621} & 0.5458 & -      \\
        VQA-X       & Language & \textbf{0.6367} & 0.5473          & 0.4813          & 0.4853 & 0.5994 \\
                    & Combined & \textbf{0.5787} & 0.4806          & 0.5216          & 0.5155 & -      \\
        \midrule
                    & Vision   & 0.4973          & \textbf{0.5164} & 0.5008          & 0.4970          & -      \\
        e-SNLI-VE   & Language & 0.5236          & 0.5714          & 0.5856          & \textbf{0.6463} & 0.4500 \\
                    & Combined & 0.5104          & 0.5439          & 0.5432          & \textbf{0.5716}  & -      \\
        \bottomrule
    \end{tabular}
    \end{adjustbox}
    \caption{Faithfulness ($S_F$) results for the default eUG model and the ablation derivatives. The highest scores per dataset are printed as bold text.}
    \label{tab:fscore-cosine-results}
\end{table}

The evaluation results of the Attribution-Similarity faithfulness metric $S_F$ (see Section~\ref{subsec:cosine-sim-rel-attr}) are presented in \autoref{tab:fscore-cosine-results}.

\begin{figure}
    \centering
    \includegraphics[width=0.75\textwidth]{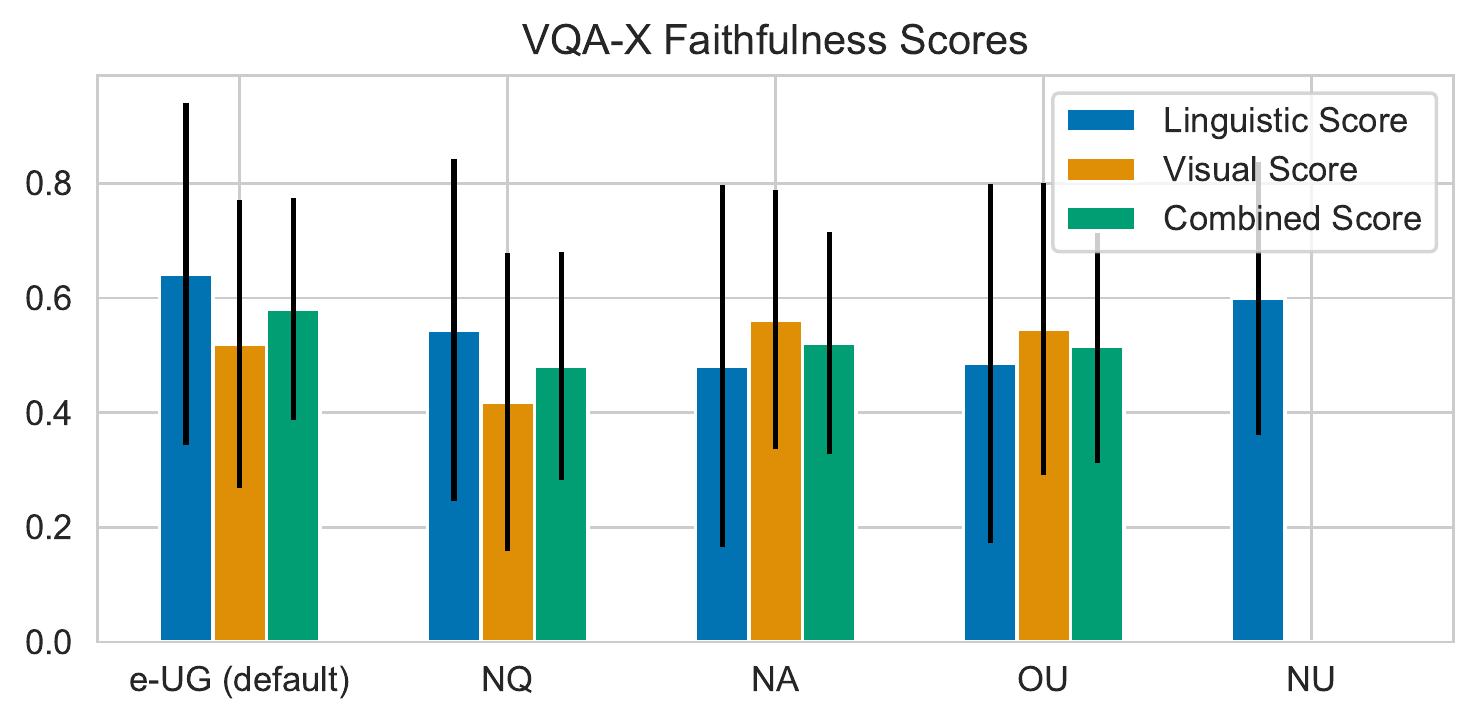}
    \includegraphics[width=0.75\textwidth]{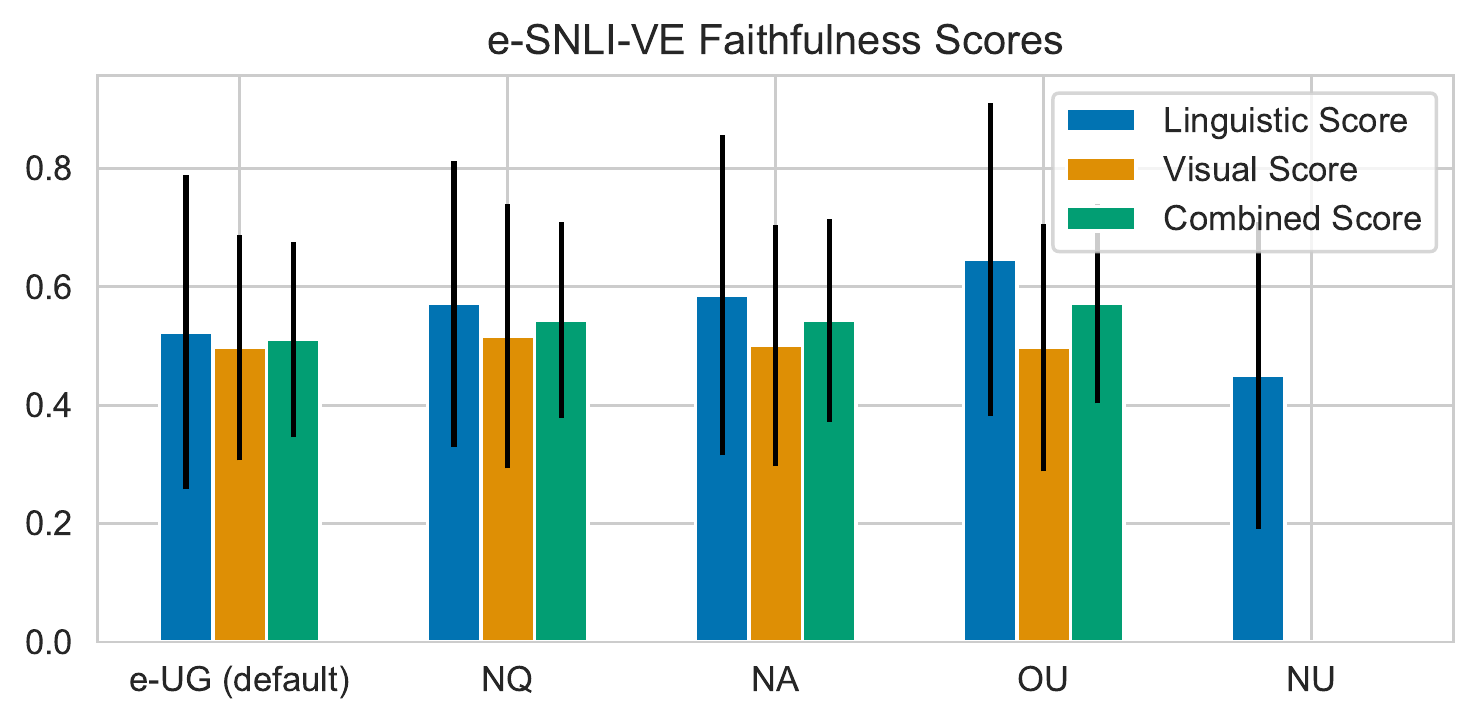}
    \caption{Linguistic, visual and combined faithfulness scores ($S_F$) on the VQA-X and e-SNLI-VE datasets. Error bars indicate the standard deviation across dataset examples.}
    \label{fig:fscores-barplot}
\end{figure}

On the VQA-X dataset, visualized in \autoref{fig:fscores-barplot}, the default configuration of the e-UG model achieves the highest performance on the linguistic and combined modality scores. On the linguistic score, the default model leads by a margin of 0.0373 points to the No-uniter-features (NU) model and with further distance to the next best model NQ by 0.0894 points. The lowest linguistic score is attained by the No-answer (NA) model with 0.4813, which is close to the result of the Only-uniter-features (OU) model with 0.4853. Surprisingly, the NU baseline reaches the highest linguistic score out of all ablated models. 

The visual modality behaves differently: While the visual score of NQ is decidedly lower (-0.1066) than the default model as well, higher scores are observed for NA (+0.0415) and OU (+0.0252) in comparison to the default e-UG model. Due to the larger difference in linguistic scores between the models, the combined faithfulness score of the default model remains the highest. Across both modalities and all models, the standard deviation is at about 50\% of the average score.

The effects of the model architecture on the faithfulness metric on the e-SNLI-VE dataset stand in contrast to the previously discussed results. While the score for the visual modality remains almost constant at around 0.5 across the models, with a maximum difference to the baseline model of (+0.0191) displayed by the NQ variant, the linguistic score is increased with fewer additional inputs to the GPT-2 explanation module: Removing the additional question (NQ: 0.5714) or the answer (NA: 0.5856) noticeably increases the score in comparison to the baseline model (0.5236). The removal of both the question and answer inputs compounds the effect to the highest score of 0.6463, which thereby also yields the highest combined score. The No-uniter-features baseline produces a significantly lower score than the baseline on the e-SNLI-VE dataset (NU: 0.4500).

\subsection{Distribution of $S_F$ Score Results}

To further analyze the effects of the ablations on the e-UG model, we investigate the distribution of score results across the models on the two datasets. We are grouping the results in 20 buckets according to their score to visualize the distributions.

\begin{figure}[h!]
    \centering
    \includegraphics[width=\textwidth]{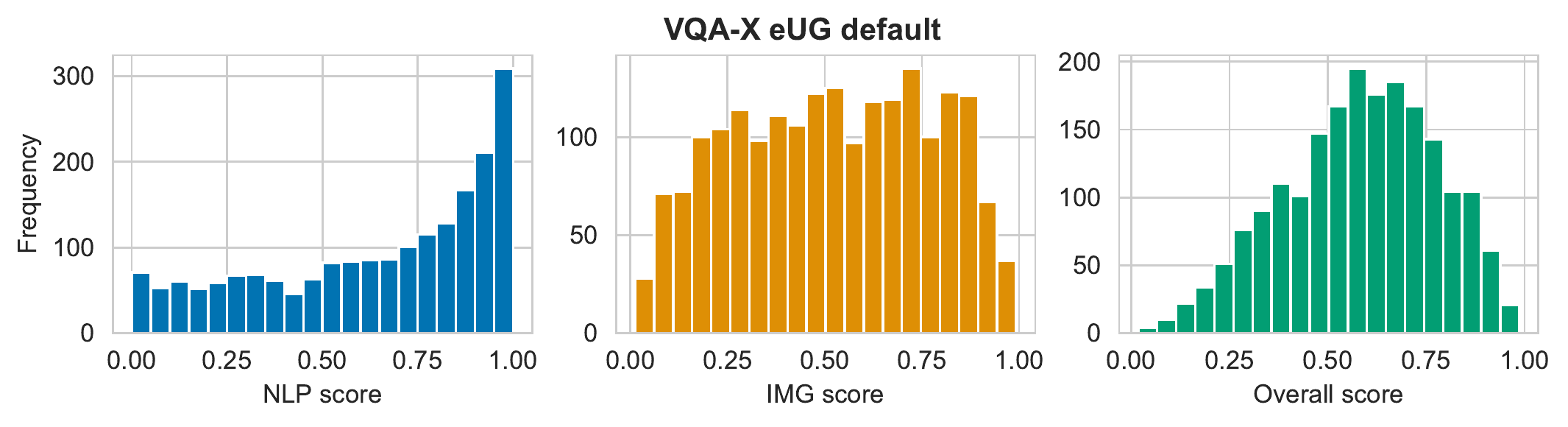}
    \includegraphics[width=\textwidth]{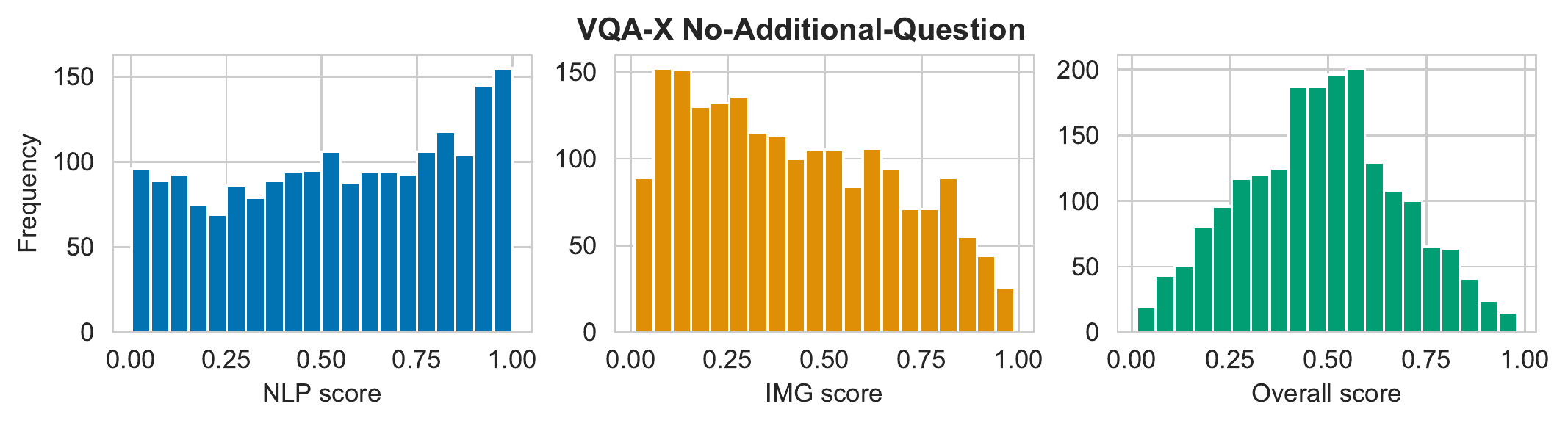}
    \includegraphics[width=\textwidth]{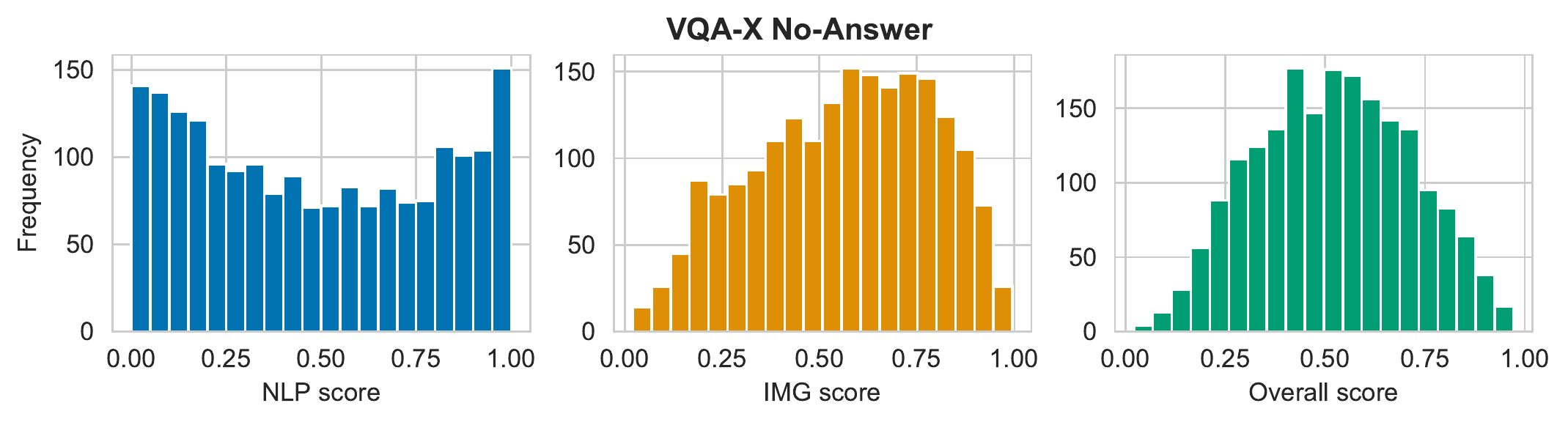}
    \includegraphics[width=\textwidth]{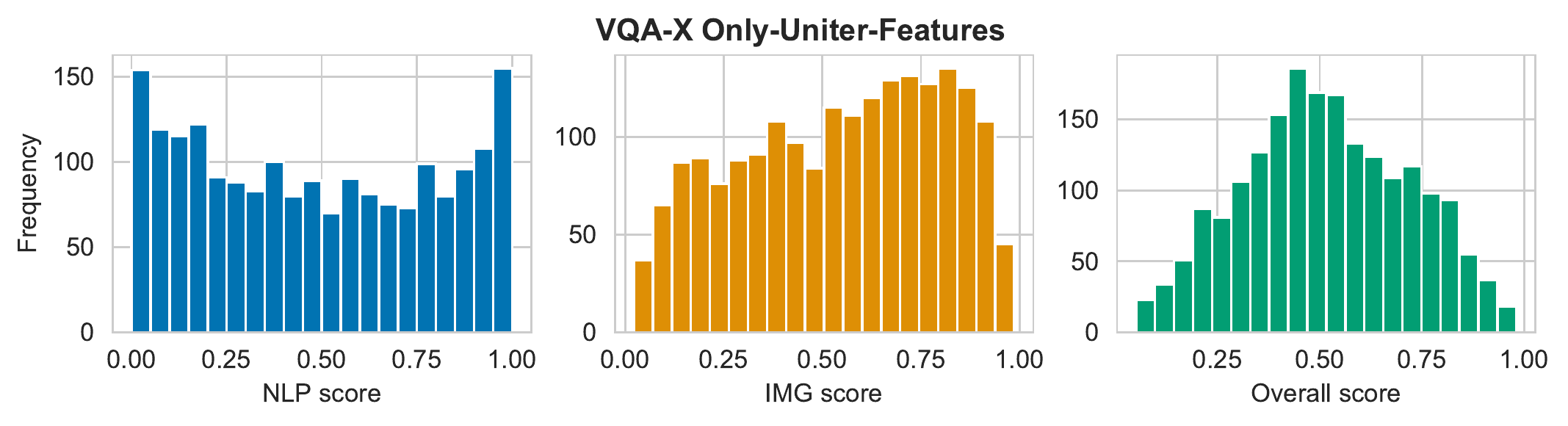}
    \includegraphics[width=\textwidth]{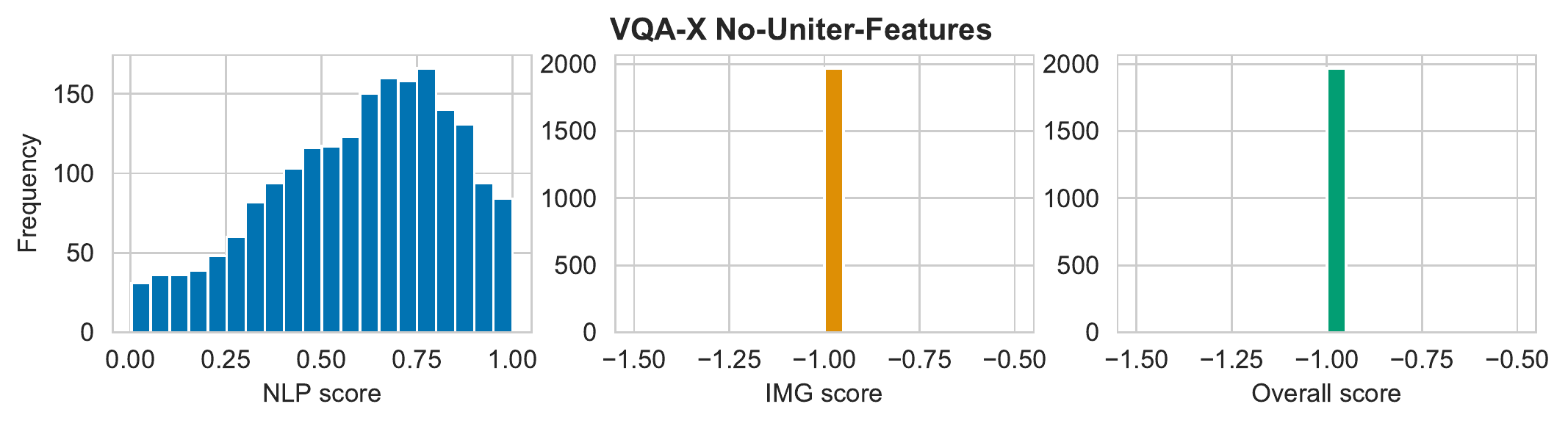}
    \caption{Histogramm of linguistic, visual, and combined faithfulness scores ($S_F$) on the VQA-X dataset for the model variations. Scores have been divided into 20 buckets for generating frequency values of the histogram.}
    \label{fig:vqax-hist}
\end{figure}

On the VQA-X dataset (Figure \ref{fig:vqax-hist}), the linguistic scores for the default eUG model are left-skewed towards high faithfulness with the top 5\% bucket representing the largest amount of values. For the NU baseline model, the linguistic score distribution is also skewed towards high scores, but with considerably fewer instances falling into the top 5-10 percentiles. The other ablated models distribute the linguistic scores more broadly, with a tendency towards a U-shaped bimodal distribution, especially for the NA and OU models. Overall, there are more examples with very high faithfulness scores ($> 0.9$) for these ablation models in comparison to the NU baseline, despite the NU model's higher mean score. On the visual modality, the score distribution of the default model is relatively uniform with few scores falling into the top and bottom 5 percentiles. Remarkably, the No-additional-question model's distribution is shifted towards the bottom scores, as already indicated by the mean score reported in the previous subsection \ref{subsec:fscore-means}, while NA and OU models are slightly left-skewed towards higher scores.

\begin{figure}[h!]
    \centering
    \includegraphics[width=\textwidth]{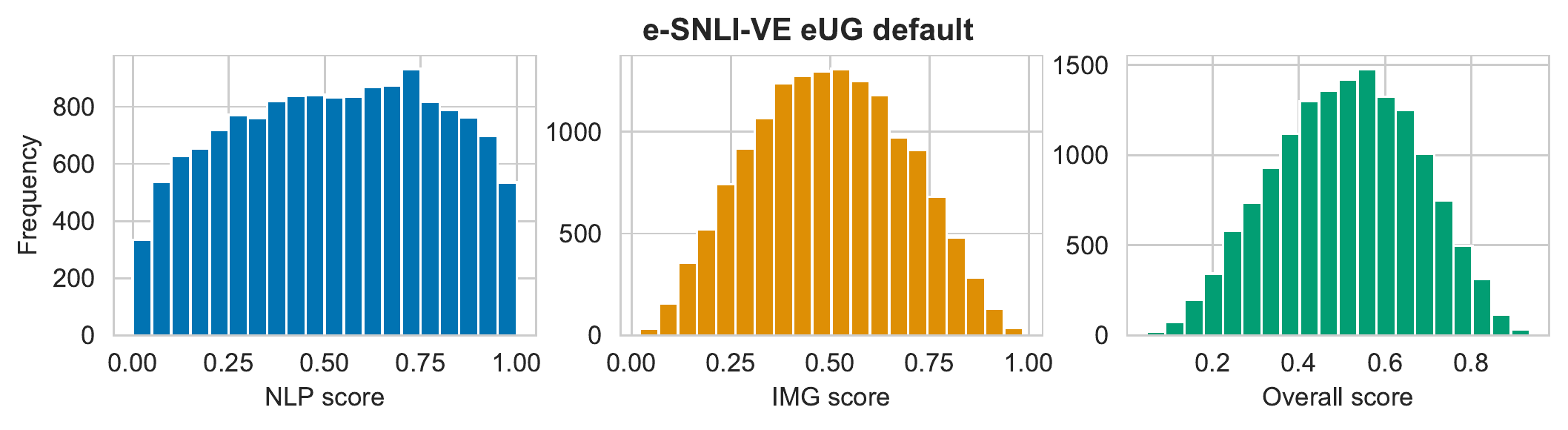}
    \includegraphics[width=\textwidth]{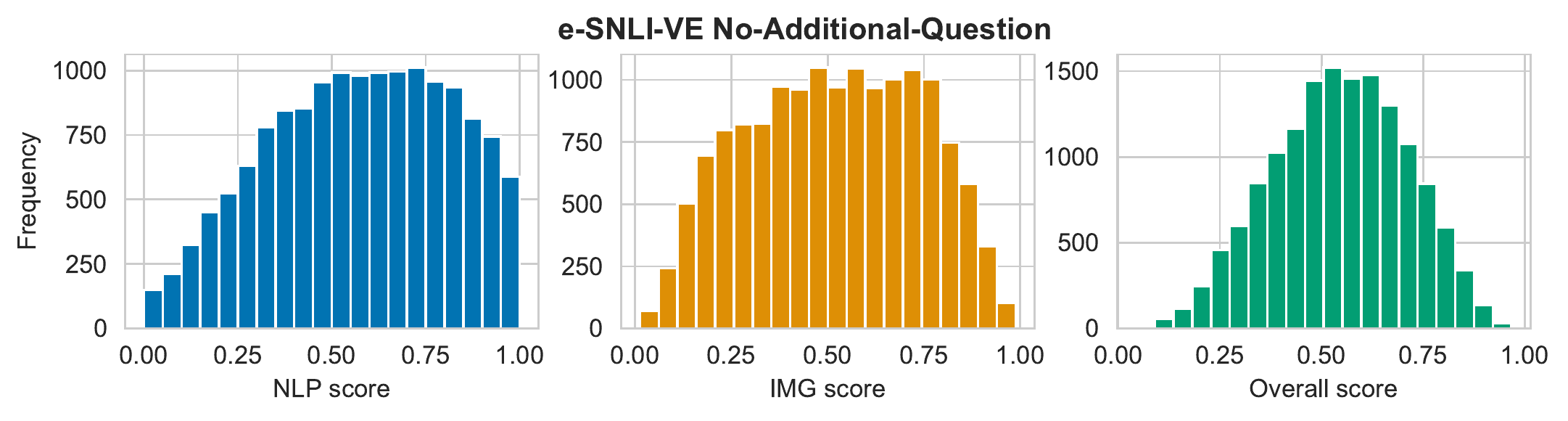}
    \includegraphics[width=\textwidth]{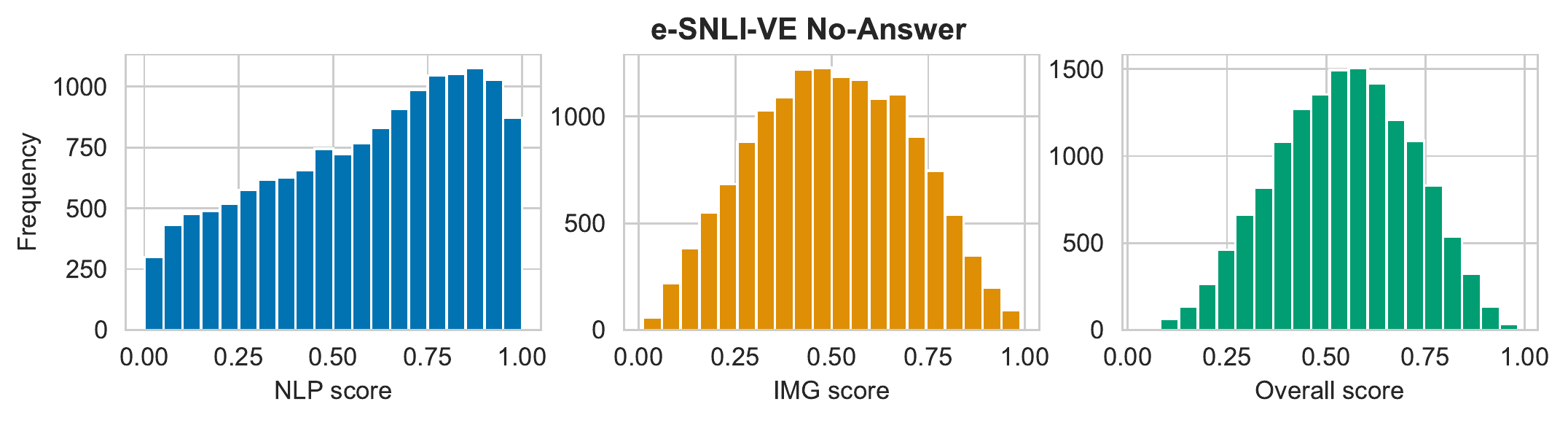}
    \includegraphics[width=\textwidth]{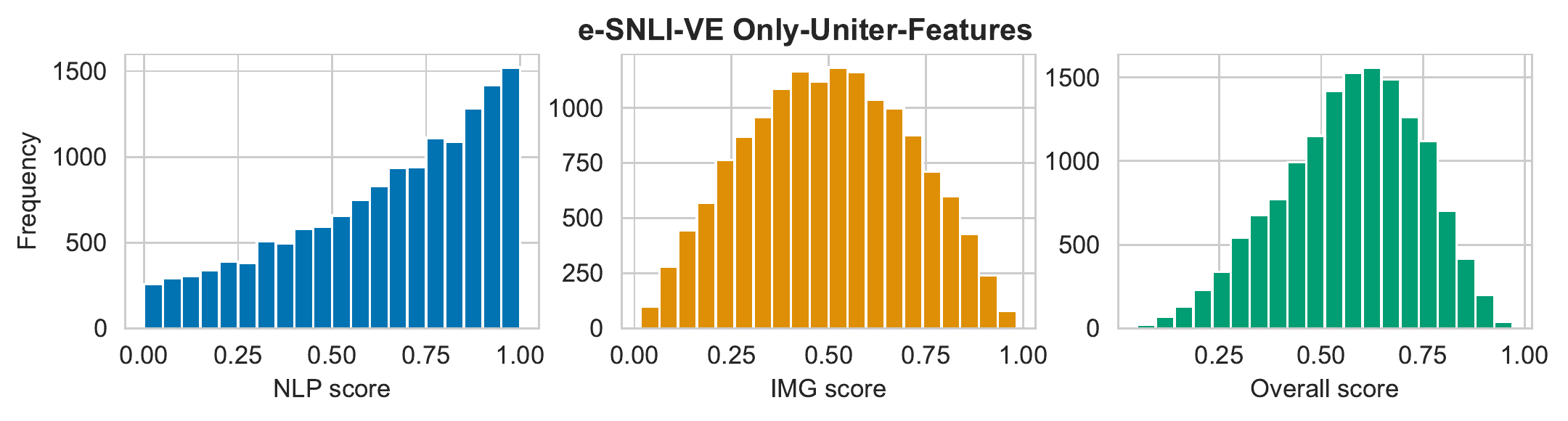}
    \includegraphics[width=\textwidth]{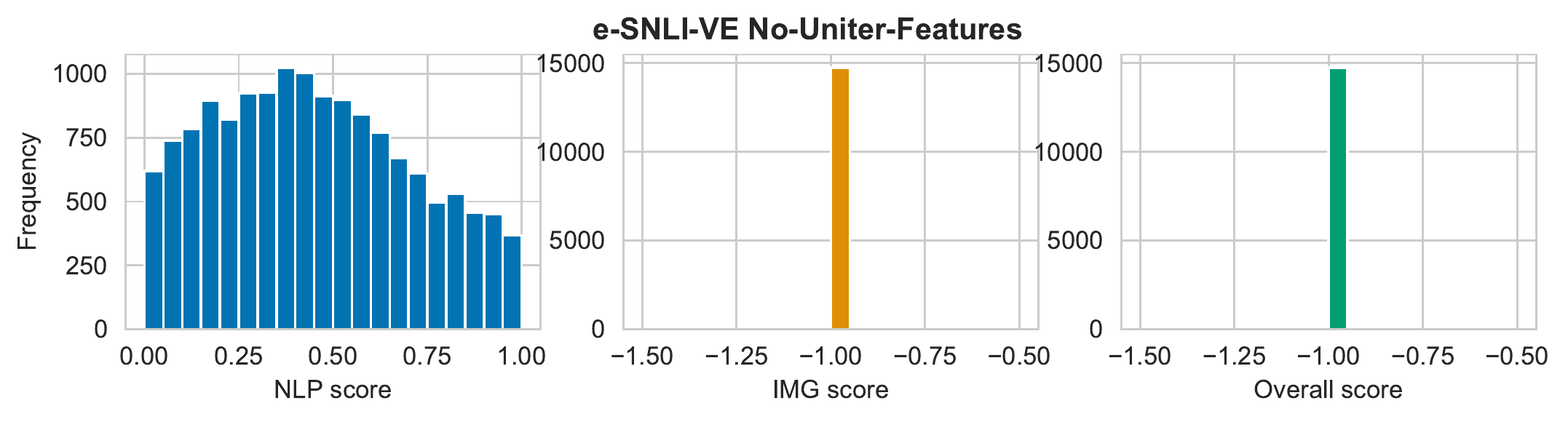}
    \caption{Histogramm of linguistic, visual, and combined faithfulness scores ($S_F$) on the e-SNLI-VE dataset for the model variations. Scores have been divided into 20 buckets for generating frequency values of the histogram.}
    \label{fig:esnlive-hist}
\end{figure}

Contrarily, the histogram results on the e-SNLI-VE dataset (Figure~\ref{fig:esnlive-hist}) prominently express the impact of model ablations on the linguistic modality: The introduction of NQ and NA ablations lead to a more left-skewed distribution shifting the median towards higher scores. The OU model, which consists of both the NQ and NA ablation further pronounces this change and leads to a clearly left-skewed distribution with the highest score count falling into the top 5 percentile. The visual modality distributions resemble normal-distribution bell curves centered around the 0.5 score. Removing only the question from the explanation (NQ) leads to a flatter curve with a slight left-skew, while no substantial change is observed on the other model variants.

\subsection{Qualitative Examples: Low, Medium and high $S_F$ scores}
\label{subsec:qualitative-examples}

In this subsection, we present different qualitative examples for low, medium, and high $S_F$ results by visualizing the feature relevance attribution with respect to the predicted answer and generated explanation.\footnote{Further examples, including model variants and the e-SNLI-VE dataset are contained in the Appendix B (PDF only)} The default e-UG model without modifications is employed on the test-split of the VQA-X dataset in these examples to generate representative examples selected from the top, middle, and bottom 6 instances ranked by the $S_F$ result on the visual modality (denoted as $S_F$-IMG in the following).
\begin{figure}
    \centering
    \includegraphics[width=0.9\textwidth]{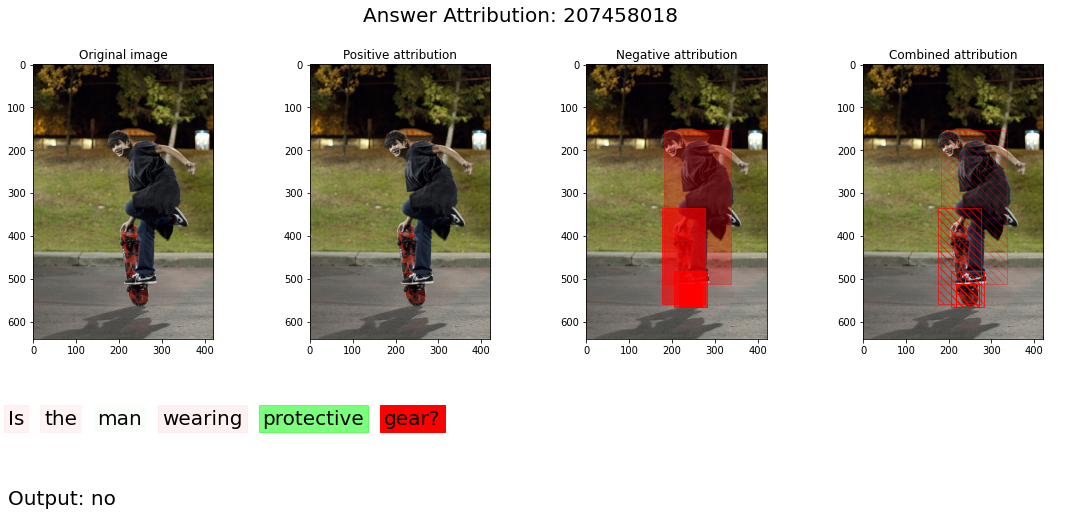}
    \includegraphics[width=0.9\textwidth]{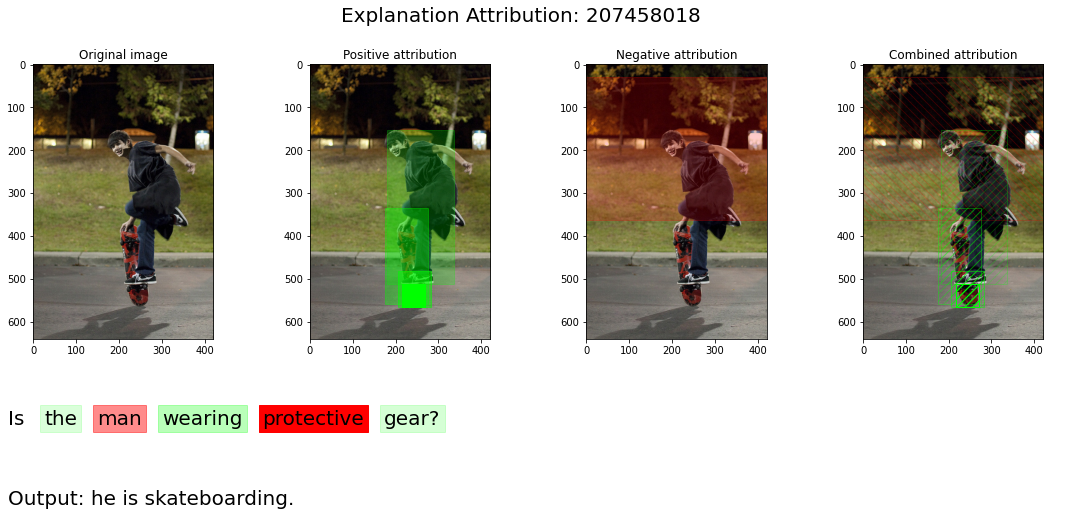}
    \caption{Example for an explanation with a low $S_F$ score for the image modality of 0.03. Positive attributions are colored green, and negative red, with the magnitude of the attribution visualized as the opacity of the color.}
    \label{fig:example-low-img-fscore-vqax}
\end{figure}

\autoref{fig:example-low-img-fscore-vqax} represents an instance of low faithfulness with a low $S_F$-IMG score of 0.030 and a linguistic $S_F$ score of 0.228. As apparent from the answer attribution, the predicted solution ``no" was primarily generated due to the word ``protective" in the input question. The visual features provided negative attribution to the prediction, possibly due to the correlation between skateboarding and protections such as helmets. A plausible interpretation of the observed attribution is that the model did not identify any significant features in the image that were associated with the word  ``protective" and therefore predicted the answer ``no'', besides the skateboarder and skateboard, which slightly inhibited the activation of the prediction. The generated explanation ``he is skateboarding'' is not explaining the answer properly. On the visual modality, the same features that negatively influenced the answer prediction are positively contributing to the explanation generation, as the generated sentence is a linguistic description of these features. Interestingly, the word ``protective'' inhibited the output, while ``gear'' has a positive influence, contrary to the answer prediction.

\begin{figure}
    \centering
    \includegraphics[width=0.9\textwidth]{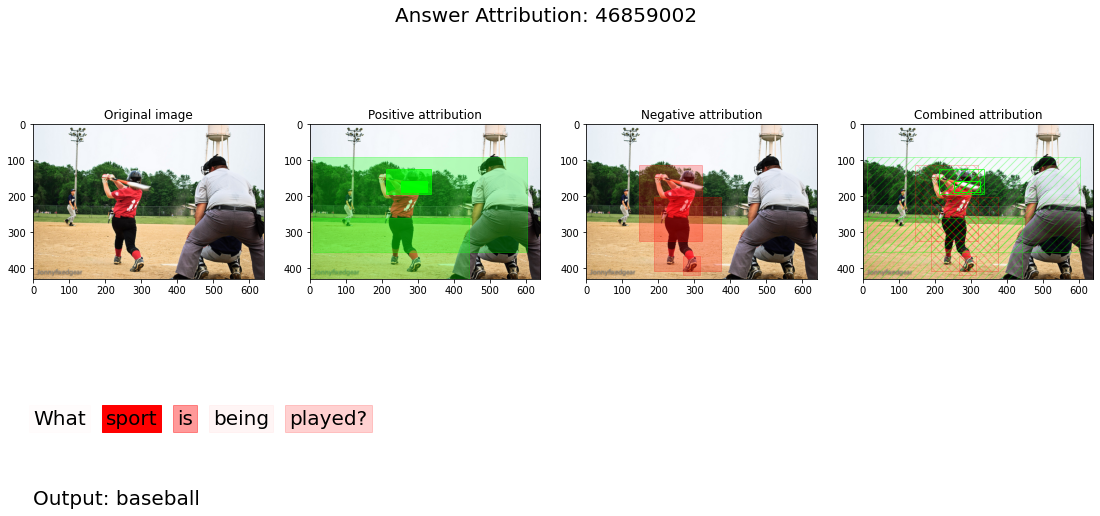}
    \includegraphics[width=0.9\textwidth]{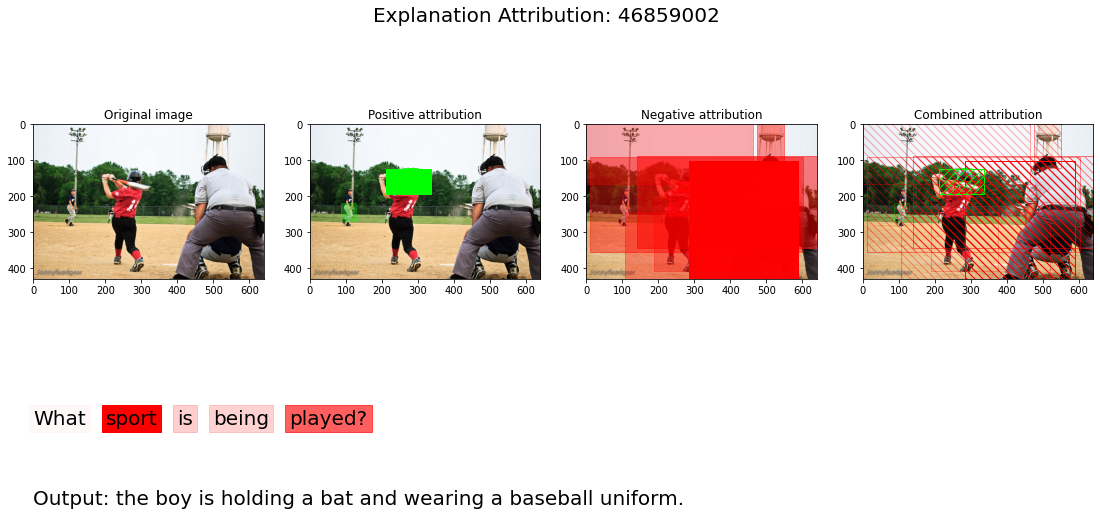}
    \caption{Example of an explanation with a medium $S_F$ score for the image modality of 0.457.}
    \label{fig:example-mid-img-fscore-vqax}
\end{figure}

As an example for a medium $S_F$-IMG score, consider \autoref{fig:example-mid-img-fscore-vqax}: On the visual modality, the area of the baseball bat is most strongly contributing to the prediction of the correct answer ``baseball", together with larger areas presumably representing the baseball court. Meanwhile, the features centered around the other areas of the man holding the baseball bat receive a slight negative attribution. Surprisingly, on the linguistic modality, all words receive a negative attribution of varying magnitude, which is reproduced in a similar manner when considering the explanation attribution. Similarly, on the visual modality, the baseball bat is the strongest positive influence on the explanation generation. A feature originating from a person in the background is another positively attributing factor. However, the larger feature areas that were assigned a positive attribution are negative for the explanation generation.

\begin{figure}
    \centering
    \includegraphics[width=0.9\textwidth]{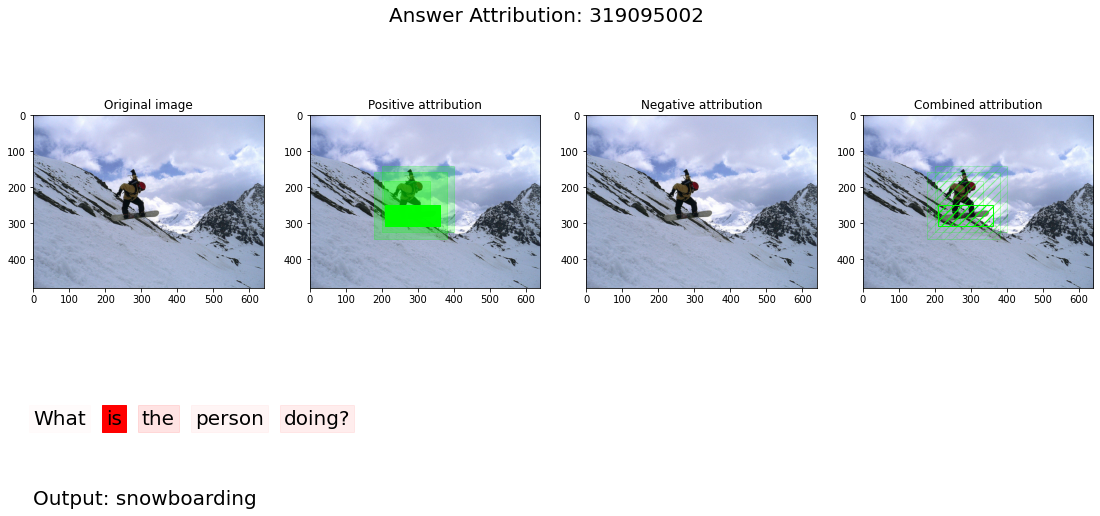}
    \includegraphics[width=0.9\textwidth]{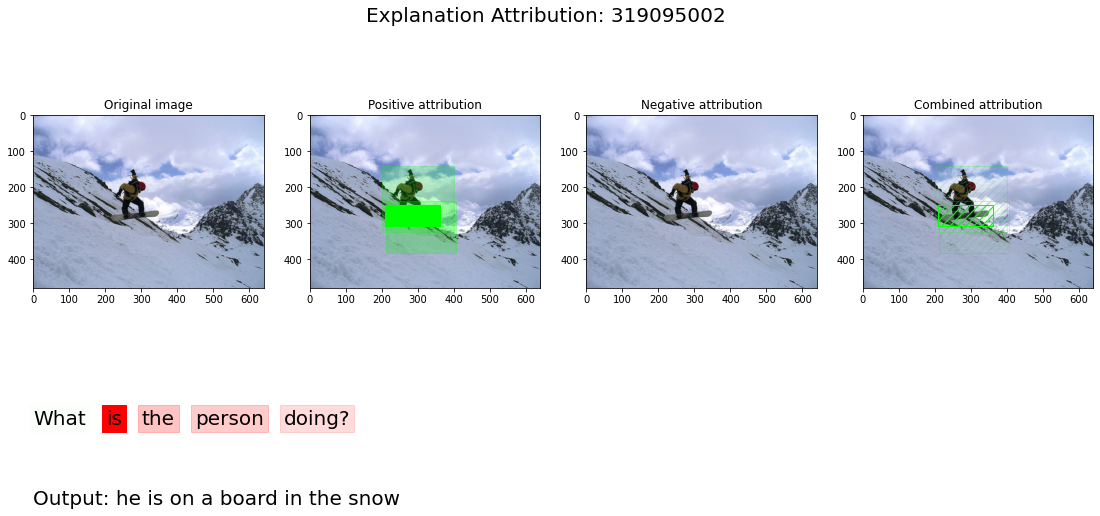}
    \caption{Example of an explanation with a high $S_F$ score for the image modality of 0.9687, as well as the language modality (0.9891).}
    \label{fig:example-high-img-fscore-vqax}
\end{figure}

Finally, \autoref{fig:example-high-img-fscore-vqax} is an example for a high faithfulness score on the visual (0.9687) and linguistic modality (0.9891). On the visual modality, both the answer and the explanation display similar relevance attributions, with positively influencing features centered around the snowboarder, with the actual snowboard receiving the strongest attribution. There is no significant relevance attributed to the rest of the scene in the image, which could indicate that the explanation referring to ``snow'' could be produced on the basis of a bias that snowboards are usually in the snow, without regarding the evidence in the Image. Another interesting observation is the fact that the words of the input question are assigned negative attributions, with the exception of the word ``What''. A possible explanation for this attribution is that the detection of a person on a snowboard is already sufficient to generate the prediction ``snowboarding" if no strong signal is extracted from the input question such as ``How many" that would lead the model to a different prediction. This explanation is plausible considering the relatively small VQA-X dataset with 32,886 question-answer pairs and 3,129 answer classes, in comparison to the e-SNLI-VE dataset that contains 401,717 question-answer pairs and only three possible answer classes.

\subsection{NLE-Sufficiency and -Comprehensiveness}

\begin{table}[]
    \centering
    \begin{adjustbox}{max width=\textwidth}
    \begin{tabular}{llccccc}
        \toprule
        Dataset &   Metric                         & e-UG     & NQ     & NA   & OU    & NU
        \\
        \midrule

        VQA-X     & NLP-Suff. $\downarrow$ & 0.2213 & 0.2045 & 0.1706 & 0.1999 & \textbf{0.1293} \\
                  & NLP-Compr. $\uparrow$  & \textbf{0.6077} & 0.5691 & 0.5642 & 0.5473 & 0.5999 \\
                  & IMG-Suff. $\downarrow$ & \textbf{0.2370} & 0.2564 & 0.2375 & 0.2657 & - \\
                  & IMG-Compr. $\uparrow$  & \textbf{0.2183} & 0.1905 & 0.1917 & 0.2097 & - \\
        \midrule
        e-SNLI-VE & NLP-Suff. $\downarrow$ & 0.1844 & 0.1602 & 0.1855 & 0.1660 & \textbf{0.0973} \\
                  & NLP-Compr. $\uparrow$  & 0.4608 & 0.3833 & \textbf{0.4690} & 0.3858 & 0.3946 \\
                  & IMG-Suff. $\downarrow$ & 0.1131 & 0.1073 & \textbf{0.0916} & 0.1490 & - \\
                  & IMG-Compr. $\uparrow$  & \textbf{0.1417} & 0.0897 & 0.1160 & 0.1113 & - \\
      \bottomrule
    \end{tabular}
    \end{adjustbox}
    \caption{Faithfulness Sufficiency and Comprehensiveness results for the default eUG model and the ablation derivatives. The best scores per dataset and metric are printed as bold text.}
    \label{tab:fscore-suff-comp}
\end{table}

\begin{figure}
    \centering
    \includegraphics[width=0.9\textwidth]{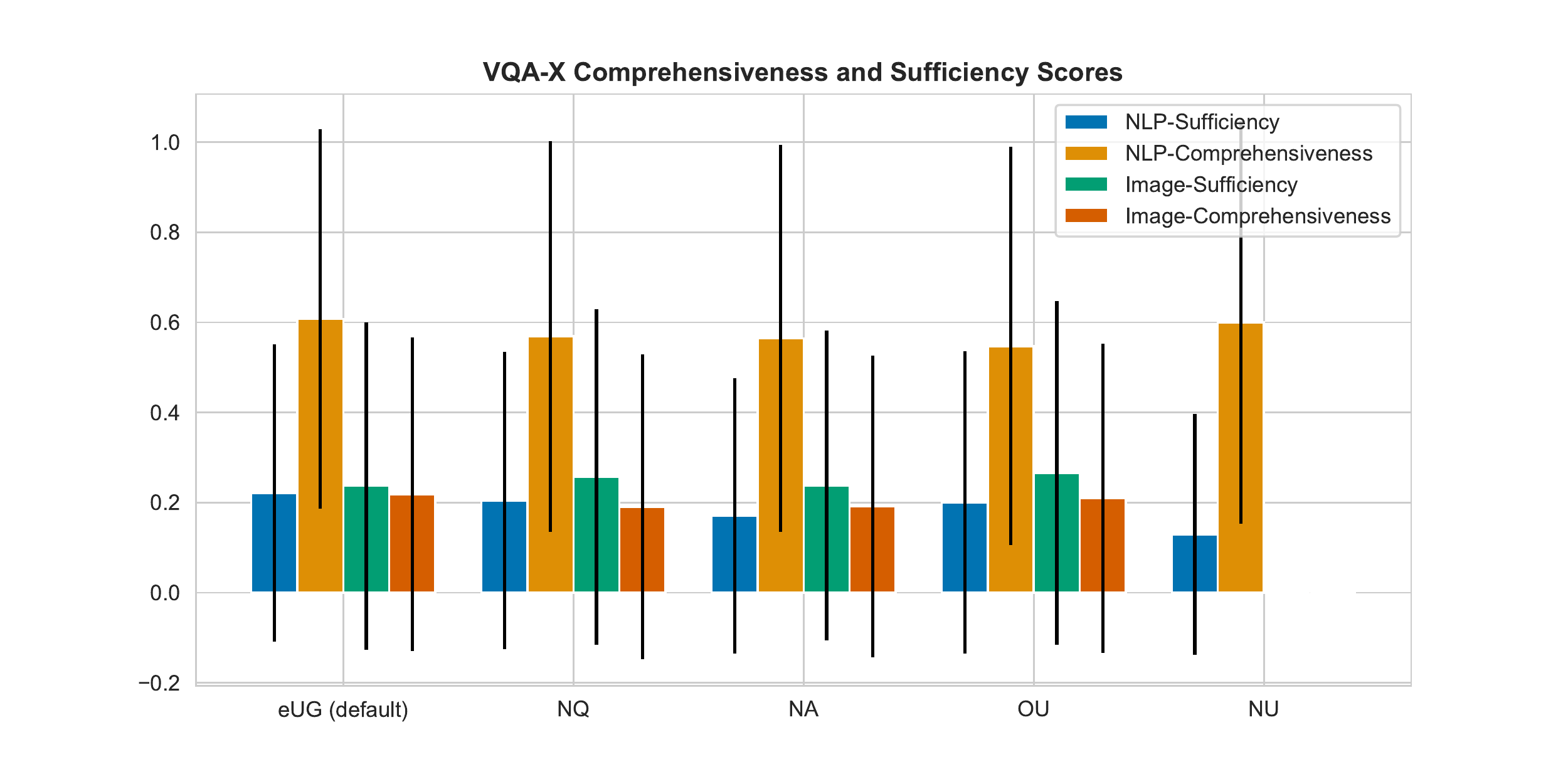}
    \caption{Faithfulness sufficiency and comprehensiveness results on the VQA-X dataset across the models.}
    \label{fig:comp_suff_vqax}
\end{figure}

\begin{figure}
    \centering
    \includegraphics[width=0.9\textwidth]{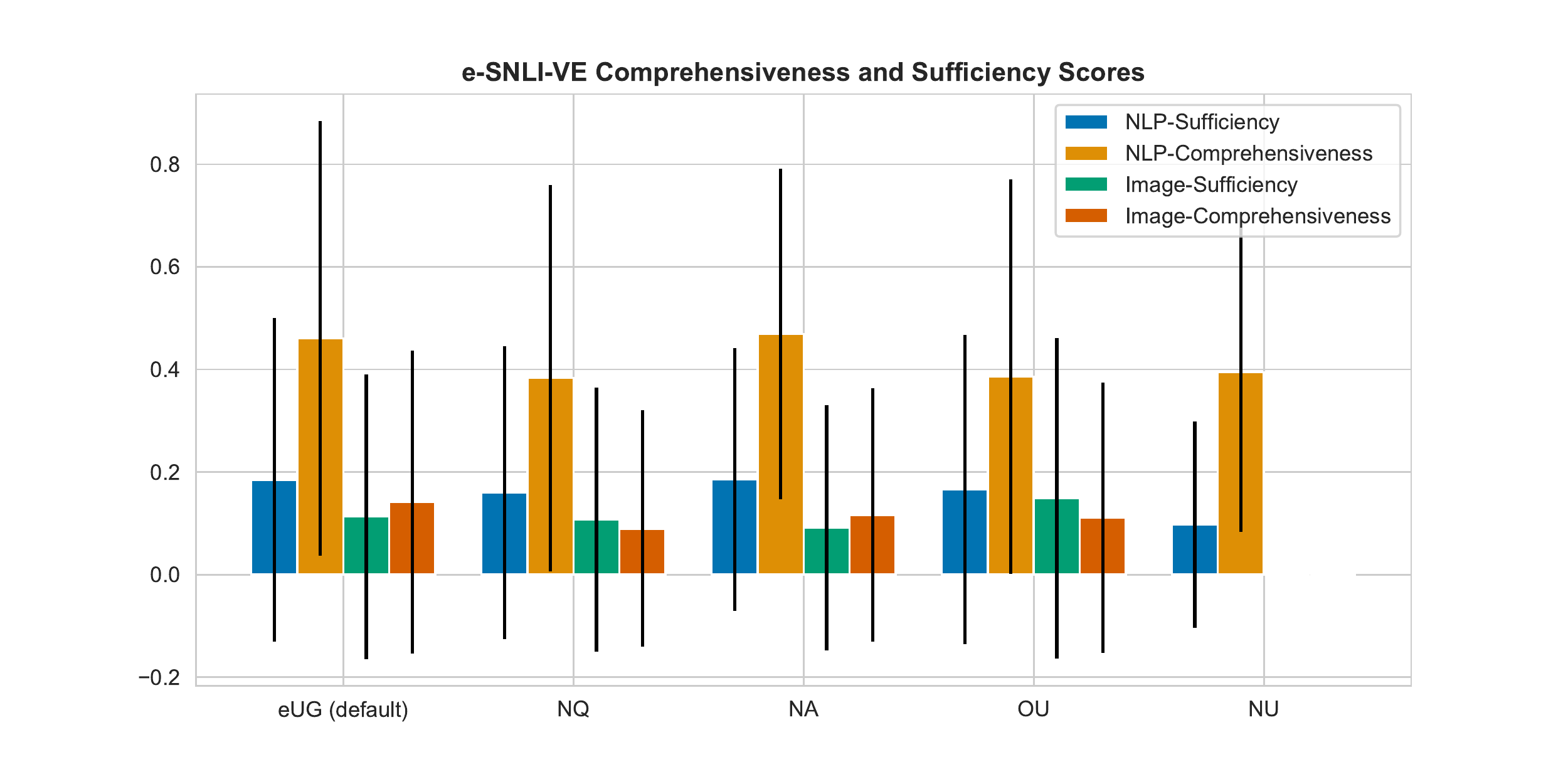}
    \caption{Faithfulness sufficiency and comprehensiveness results on the e-SNLI-VE dataset across the models.}
    \label{fig:comp_suff_enslive}
\end{figure}

The NLE-Sufficiency and -Comprehensiveness scores, as described in Section \ref{subsec:sufficiency-comprehensiveness}, are presented in Table~\ref{tab:fscore-suff-comp}.

Concerning the VQA-X dataset, the best scores are recorded for the default e-UG model without any ablations, with the only exception of NLP-Sufficency (linguistic modality), where the No-uniter-features baseline reaches a significantly lower (better) score with a 0.09 difference. Further, NLP-Sufficiency score is the only metric where the other models reached better scores than the original model.

Similarly, on the e-SNLI-VE dataset, the best NLP baseline result is produced by the NU baseline as well, with a 0.09 difference to the next best result of the NQ variant. The NA model yields the best scores in NLP-Comprehensiveness and IMG-Sufficiency, while the best IMG-Comprehensiveness result is generated by the default configuration of the e-UG model.

As an additional analysis, to investigate the agreement between the Sufficiency and Comprehensiveness scores and the cosine-similarity based $S_F$ score, we assess the correlation between the measurements. To this end, the Pearson correlation coefficient is calculated between all scores across the dataset examples. As an illustration, the results for the default e-UG model on the e-SNLI-VE dataset are provided in Table~\ref{tab:correlation-eUG-esnlive}. Across all models on each of the datasets, no linear relationship ($r < 0.2$) is observed between the Sufficiency, Comprehensiveness, and $S_F$ scores, with the exception of the $S_F$-Overall score, which is by definition a combination of the individual modality scores and therefore correlated to them.\footnote{Additional results are included in Section~\ref{appx:fmetrics-correlation} of the Appendix}

\begin{table}
\centering
\begin{adjustbox}{max width=\textwidth}

\begin{tabular}{lrrrrrrr}
\toprule
\textbf{e-UG (default)} &  NLP-Suff. &  NLP-Comp. &  IMG-Suff. &  IMG-Comp &  $S_F$-NLP &  $S_F$-IMG &  $S_F$-Overall \\
\midrule
NLP-Suff.     &      1.000 &      0.007 &      0.049 &     0.072 &      0.133 &     -0.012 &          0.100 \\
NLP-Comp.     &      0.007 &      1.000 &     -0.055 &    -0.020 &     -0.057 &     -0.135 &         -0.124 \\
IMG-Suff.     &      0.049 &     -0.055 &      1.000 &     0.075 &     -0.081 &      0.027 &         -0.049 \\
IMG-Comp      &      0.072 &     -0.020 &      0.075 &     1.000 &     -0.020 &      0.035 &          0.004 \\
$S_F$-NLP     &      0.133 &     -0.057 &     -0.081 &    -0.020 &      1.000 &      0.013 &          0.815 \\
$S_F$-IMG     &     -0.012 &     -0.135 &      0.027 &     0.035 &      0.013 &      1.000 &          0.590 \\
$S_F$-Overall &      0.100 &     -0.124 &     -0.049 &     0.004 &      0.815 &      0.590 &          1.000 \\
\bottomrule
\end{tabular}
\end{adjustbox}
\caption{Pearson correlation coefficient of the faithfulness metrics correlation on the e-SNLI-VE dataset for the default e-UG model. No correlation ($r < 0.2$) between comprehensiveness/sufficiency results and the similarity-based metrics is observed.}
\label{tab:correlation-eUG-esnlive}
\end{table}

\subsection{Analysis of Modality Influence}

To further analyze the importance of different inputs to the explanation module, we assessed the total attribution per modality, as well as the attribution to the UNITER question input versus the additional question input.

\paragraph{Attribution Per-Modality.}
The attribution per modality is calculated as the sum of all feature attributions to the image (visual) and NLP (linguistic) modality. 

\begin{figure}[h]
    \centering
    \includegraphics[width=\textwidth]{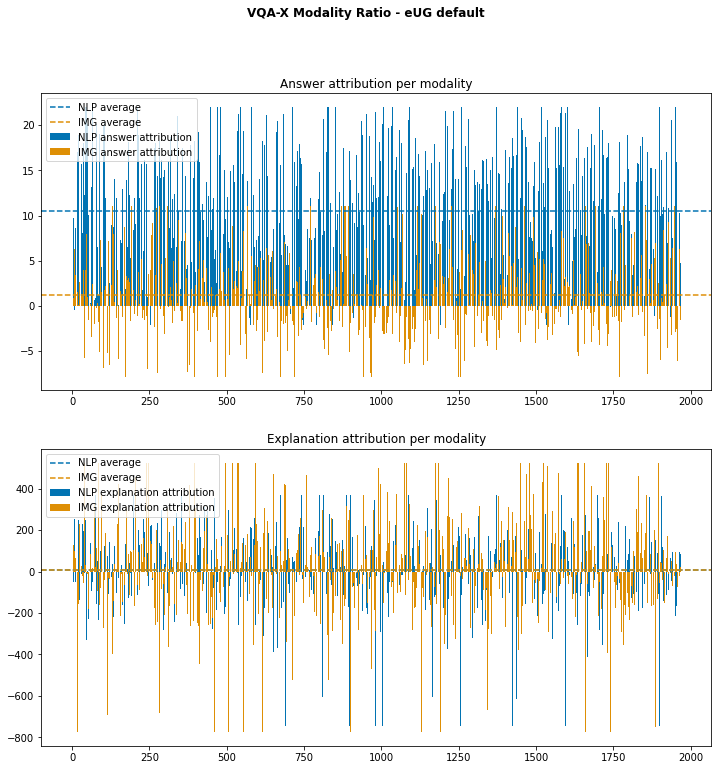}
    \caption{Attribution to visual and linguistic modality on the VQA-X dataset across dataset examples for the default eUG model.}
    \label{fig:vqax-modality-vanilla}
\end{figure}

\begin{figure}
    \centering
    \includegraphics[width=\textwidth]{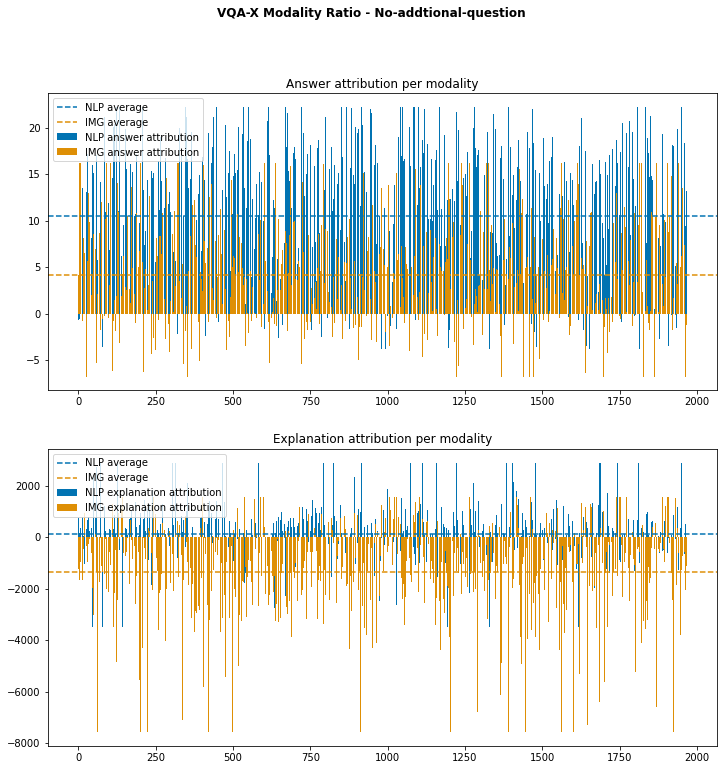}
    \caption{Attribution to visual and linguistic modality on the VQA-X dataset across dataset examples for the No-additional-question model.}
    \label{fig:vqax-modality-noquestion}
\end{figure}

On VQA-X with the default e-UG model (\autoref{fig:vqax-modality-vanilla}), we observe a higher contribution of the linguistic modality in both magnitude and average total contribution for the answer prediction. While negative attribution (inhibiting the prediction) only rarely occur, negative attributions are frequently assessed for the visual modality. For the explanation generation, both image and language modality are averaging close to zero, but display a large magnitude\footnote{The attribution results for the explanation are presumably higher in overall magnitude than the answer attributions due to the process of summing the attribution vectors for each token.}. This finding implies stark differences in the explanation generation modality importance across the dataset examples. In comparison, on the answer attribution, the No-additional-question model (NQ, \autoref{fig:vqax-modality-noquestion}) displays a considerably higher average relevance attributed to the image modality than the default model, with negative image attributions occurring less frequently. While the attribution to the linguistic modality of the explanation generation was similar to the default model, the image modality is receiving an overall strongly negative attribution with rare occurrences of positive attributions.

\begin{figure}
    \centering
    \includegraphics[width=\textwidth]{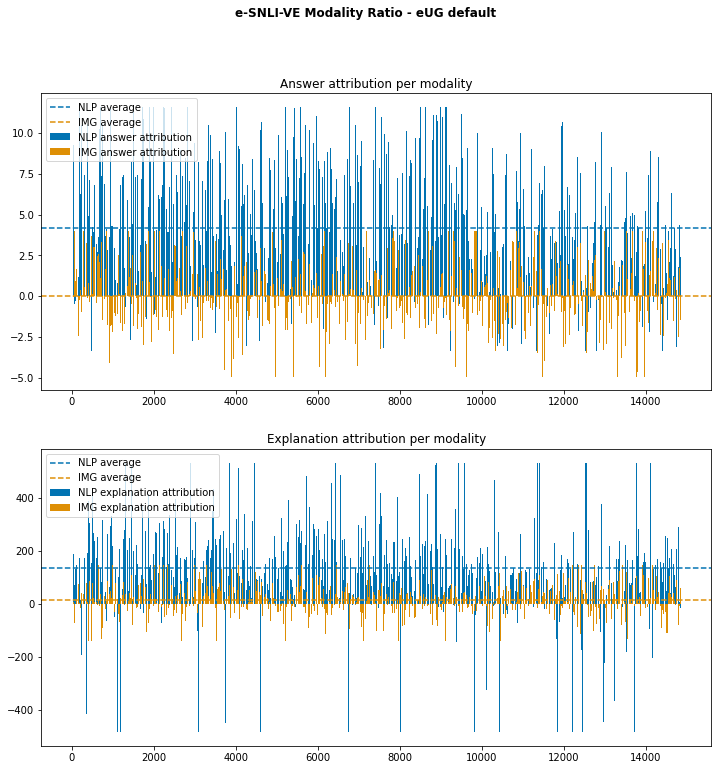}
    \caption{Attribution to visual and linguistic modality on the e-SNLI-VE dataset across dataset examples for the default eUG model.}
    \label{fig:esnlive-modality-vanilla}
\end{figure}

\begin{figure}
    \centering
    \includegraphics[width=\textwidth]{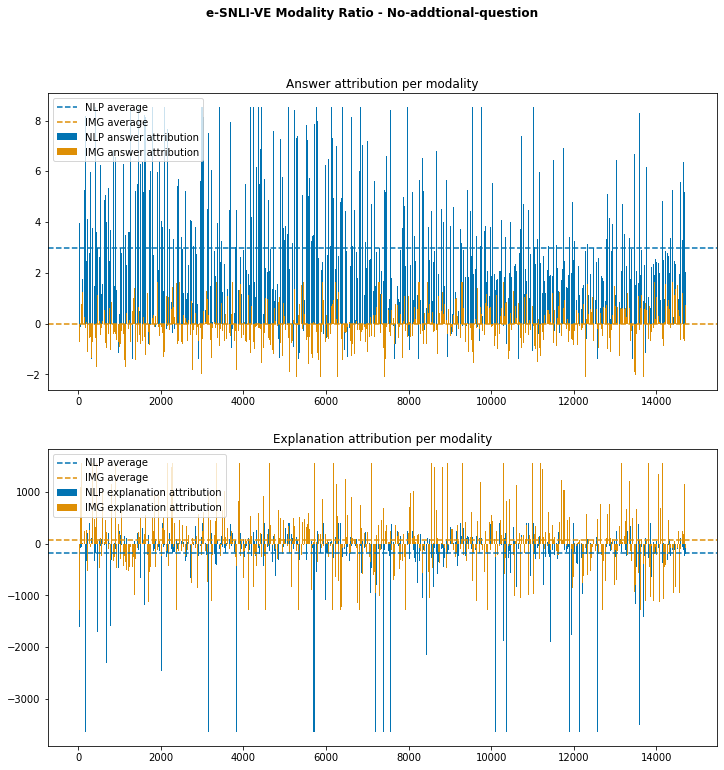}
    \caption{Attribution to visual and linguistic modality on the e-SNLI-VE dataset across dataset examples for the No-additional-question model.}
    \label{fig:esnlive-modality-noquestion}
\end{figure}

The modality influence results on the e-SNLI-VE dataset (\autoref{fig:esnlive-modality-vanilla}) are similar for the answer prediction on the default e-UG model  compared to VQA-X, although the image modality contribution is lower in magnitude in relation to the linguistic modality and also on average as it is closer to zero. For the explanation generation, the NLP attribution is strongly positive on a large majority of examples in contrast to the VQA-X results, while the image modality is equally distributed along a zero mean with relatively low magnitude. The NQ variant (\autoref{fig:esnlive-modality-noquestion}) exhibits similar modality attributions for the predicted answers in comparison to the default e-UG model. However, for the explanation generation, the magnitude of the attributions is substantially higher in total and the language modality is attributed mostly negative relevance scores. 

\paragraph{Attribution to UNITER and GPT-2 Language Input.} The finding of the previously presented changes in modality importance for the explanation generation from the default to the NQ variant can be extended by an analysis of the linguistic modality of the default model, by splitting the language modality importance into the UNITER attribution (question/hypothesis input to the task model) and the additional and direct question input to the GPT-2 explanation generation module, that is ablated for the NQ variant.

\begin{figure}
    \centering
    \includegraphics[width=\textwidth]{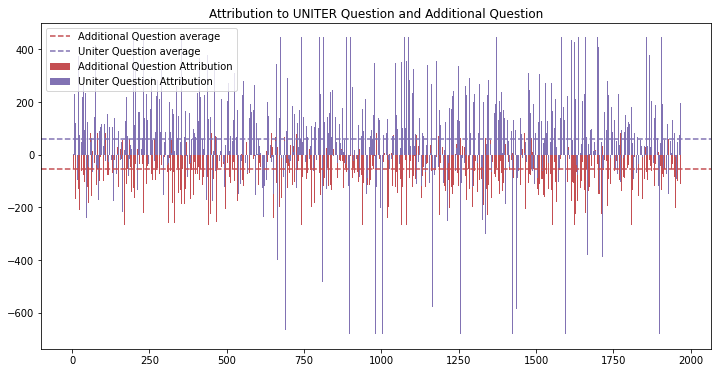}
    \caption{Attribution to UNITER language input versus GPT-2 language input ("additional question") on the VQA-X dataset across dataset examples for the default eUG model.}
    \label{fig:vqax-uniter-vs-gpt}
\end{figure}

\begin{figure}
    \centering

    \includegraphics[width=\textwidth]{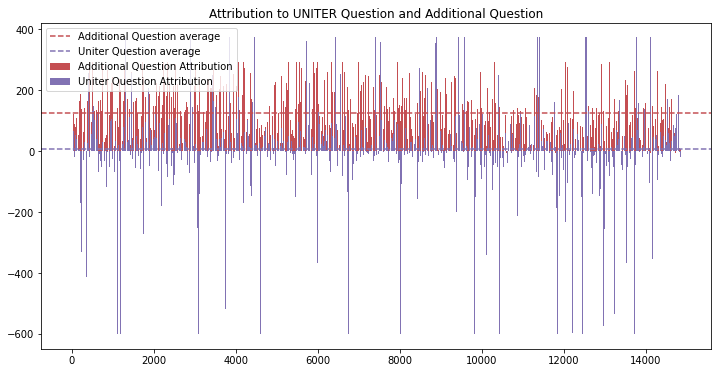}
    \caption{Attribution to UNITER language input versus GPT-2 language input ("additional question") on the e-SNLI-VE dataset across dataset examples for the default eUG model.}
    \label{fig:esnlive-uniter-vs-gpt}
\end{figure}

The language input attribution results for the VQA-X dataset are presented in \autoref{fig:vqax-uniter-vs-gpt}; e-SNLI-VE results are shown in \autoref{fig:esnlive-uniter-vs-gpt}. While on VQA-X the UNITER question input receives positive attribution on average, the negative relevance scores are attributed to the additional question input. By contrast, the attributions on the e-SNLI-VE dataset are positive for the additional question and close to zero for the UNITER question, indicating that the default eUG model relies more on the additional question input on the e-SNLI-VE dataset than on the VQA-X dataset, where the UNITER question input is more important.

\section{Discussion}

\paragraph{Attribution-Similarity.} The results from the presented faithfulness evaluation using the Attribution-Similarity $S_F$ score, as well as the NLE-Comprehensiveness and -Sufficiency metrics reveal different outcomes on the two investigated datasets across the several model variations. A possible determining factor could be the comparatively small number of examples in the VQA-X dataset, as discussed previously. Especially regarding the high number of classes, the dataset might still be too small and contain too many biases to show the expected effects of the model variations on the faithfulness metrics. For example, the No-Uniter-Features baseline achieves a higher linguistic $S_F$ score than the ablated models that have to rely to a varying degree on common representations and should therefore be more likely to display a similar relevance distribution for the explanation and the task answer. This result indicates that the model is unable to learn common representations without the additional question.

However, on the e-SNLI-VE dataset, we observe that the removal of additional inputs from the explanation generator increases the $S_F$-NLP score. This result demonstrates the utility of the proposed $S_F$ metric and indicates that the explanation module is successfully relying on representations shared with the task-module that are pre-selected according to their relevance for the task-answer. 
An analysis of the distribution of influence on the linguistic modality to the UNITER and additional question inputs reveals that the default e-UG model relies more strongly on the additional question on e-SNLI-VE. After the removal of this input on the ablation models, more faithful explanations are learned by requiring the model to rely on the UNITER inputs.
On the visual modality, the $S_F$-IMG distribution does not change decidedly and remains as a bell curve centered around the 0.5 score. We therefore only observe the expected effects according to hypothesis~(H2) on the linguistic and combined $S_F$ score on the e-SNLI-VE dataset, and not  on the visual modality or on the VQA-X dataset.

There are at least two potential explanations for these results: 
First, the modality influence results indicate that, in terms of average score and magnitude, the visual modality is not as influential on the e-SNLI-VE models for the answer prediction as on the VQA-X models. For the explanation generation, the total magnitude of the image modality is lower as well for VQA-X. This suggests that, for the majority of the generated tokens on the e-SNLI-VE explanations, the visual modality is not influential, resulting in the attribution of low-magnitude noise to the input, resulting in a bell-shaped normal distribution across the models. 
Another possible explanation is 
that the shared representations learned on e-SNLI-VE contain the majority of the image information without the emergence of task-specific representations, leading the task and explanation model to select different features from the image. A potential remedy to this finding could be a restriction of the capacity of these shared features, forcing the representations to pre-process and pre-select the visual features in a way that is suitable for both the answer and explanation output simultaneously. Alternatively, the introduction of a faithfulness loss that regularizes the model to exhibit similar relevance distributions for explanation and answer could also lead to shared representations, an option that was explored by Wo and Mooney~\cite{wu2018faithful}. Further, changes to the model architecture that promote the explanation module to attend to the relevant features can be introduced, such as biasing the explanation generator inputs based on the relevance of features when predicting the answer, i.e. via using the attention maps from the task module as a bias on the explanation generator inputs. 
These proposed interventions could also be combined.

\paragraph{NLE-Sufficiency and -Comprehensiveness.} The NLE-Sufficiency and NLE-Comprehensiveness results were mostly not in agreement with the previously discussed $S_F$-Score, leading us to reject the corresponding hypothesis~(H3) and also, but limited to these two metrics, hypothesis (H2) on the relative impact of the model ablations, based on the present results. Although all proposed faithfulness metrics a closely related from a theoretical standpoint (see Section ~\ref{subsec:sufficiency-comprehensiveness}), analyzing the correlation between the NLE-Sufficiency and Comprehensiveness and $S_F$ measures reveals that they do not have a significant relationship at a per-example level, at least for the results discussed. The low correlation implies that the two measures are indeed assessing a measurement that is fundamentally different from the $S_F$ results, but with the presented data, it is unclear in how far the NLE-Sufficiency and Comprehensiveness metrics are useful. As discussed in Section~\ref{subsec:sufficiency-comprehensiveness}, the measures could provide a more detailed insight into how the relevance of the explanation model is assigned: For example,  whether the explanation is based on a subset of important features or whether the explanation is generated using features that are not important for the answer prediction. These measurements might be more suitable for analyzing a model that is already scoring high on the $S_F$ metric.

\paragraph{Qualitative Examples and General Discussion.} From the qualitative analysis of individual examples in Section~\ref{subsec:qualitative-examples}, we notice that especially for the explanation generation, not every word is necessarily supported by evidence from the input question or image. It is reasonable to assume that the model is frequently relying on biases that are represented in the weights of the model, thereby acting as stored ``knowledge" extracted from the dataset. As an example, consider \autoref{fig:example-high-img-fscore-vqax}, where the explanation referred to ``snow'', even though it was not evident from the relevance attribution that the environment around the snowboarder was considered. Rather, the model might have learned that snowboards are always in the snow and that referring to the word ``snow'' in the explanation reduces the explanation loss. Another example is the reference to ``baseball uniforms'' in \autoref{fig:example-mid-img-fscore-vqax}, despite the fact that the uniforms did not receive significant relevance. Currently, we do not differentiate between the generated tokens that directly refer to inputs and those that primarily rely on the model's internal knowledge. This is also the case for syntactic words, such as articles and prepositions. The relevance attributed to these words is included in the overall explanation relevance and could introduce undesired noise into the calculation of the faithfulness measures, especially if the total relevance for a particular modality is low (see discussion of $S_F$-IMG e-SNLI-VE results above). Wu and Mooney~\cite{wu2018faithful} trained a ``source selector'' in their model, which should learn to differentiate if tokens should be predicted based on evidence or internal linguistic or world-knowledge. An alternative approach that does not require modification of the model and the training procedure filtering out token attributions for the calculation of $S_F$, if their total attribution falls beneath a certain threshold, which can be explored in future work.

Due to this limitation of not considering the impact of this internal model knowledge, especially with biases in smaller datasets, the three proposed metrics might not be suited to reliably compare the faithfulness of relatively unfaithful models. Rather, they could be useful to verify a faithful model, as such a model should reach a high $S_F$ score according to the rationale laid out in Section~\ref{subsec:formal-rationale}. The score could then potentially be more suitable to assess the impact of design decisions on models that are already producing faithful explanations, as observed on the $S_F$-NLP score of the e-SNLI-VE models.

\subsection{Limitations}

The interpretation of the presented results is subject to several limitations. Some of these limitations have been previously discussed. For example, the current formulation of the $S_F$-score and the Sufficiency and Comprehensiveness metrics do not distinguish between explanation tokens that are generated based on evidence in the inputs, or according to internal knowledge. Due to this limitation that is likely introducing noise, the current faithfulness measures could be suited more for verifying highly faithful models than comparing the relative faithfulness of low-faithful models.

Regarding the explanation quality in terms of human ratings, we have only evaluated the explanations on automatic metrics that are solely correlated with human judgments. As the models are very close to each other according to these results, an evaluation by participants could give a better insight into the actual quality of explanations given by the ablated models.

Further, there is currently an ongoing debate about the quality and correctness of feature attribution methods (see e.g.~\cite{ijcai2022-90}). For calculating our metrics, we have to assume that the attribution quality of Integrated Gradients is high enough to produce sufficiently accurate relevance vectors for comparing the answer to the explanation attributions.

Lastly, the external validity of our results is limited by the evaluation of e-UG variations as only a single type of model and the two datasets investigated. As the differences between the e-SNLI-VE and VQA-X datasets on the faithfulness results are substantial, the utility of the presented metrics should be further validated by employing different kinds of models on more datasets. 
\cleardoublepage

\chapter{Conclusion}\label{conclusion}

In this thesis, we developed and evaluated three metrics for assessing and comparing the faithfulness of natural language explanations in vision-language tasks: 

Firstly, we extended a faithfulness score initially presented by Wu and Mooney \cite{wu2018faithful} to the proposed Attribution-Similarity metric by augmenting the existing method to consider multiple modalities and implementing it using the model-agnostic feature attribution method Integrated Gradients with respect to the model inputs, thereby opening up the possibility of using the proposed score as an additional dimension in VL-NLE benchmarks such as e-ViL~\cite{kayser2021vil}. Additionally, we provide a formal rationale for the proposed score, which was previously missing from the literature. In contrast to the necessary faithfulness conditions proposed by Wiegreffe et al~\cite{wiegreffe2021measuring}, our Attribution-Similarity approach is an arguably more direct measure of NLE-faithfulness as it does not depend on retraining the model (Feature Importance Agreement) or on auxiliary models assessing simulatability as an indirect measure using quality metrics (Robustness Equivalence).

As two further metrics, we adapted the Sufficiency and Comprehensiveness measurements, as proposed by De Young et al.~\cite{deyoung2020eraser}, to Natural Langauge explanations. We show that our proposed NLE-Sufficiency and -Comprehensiveness metrics are not correlated to Attribution-Similarity for the present results. Although the conclusions drawn from the results of the presented models were limited, we laid out how the metrics can be used to better understand and analyze the features utilized for explanation generation in contrast to the answer prediction.

Furthermore, we evaluated various modifications of the e-UG model by ablating the additional inputs to the explanation generation module. We show that, according to the automatic explanation metrics and analysis of qualitative examples, explanation generation is learned without additional question and/or answer input in the form of textual embeddings. Rather, features from the task module seem to be sufficient to generate high-quality rationales, a finding that has to be verified by human evaluation in a participant study. Additionally, our results of the models on the task scores of e-ViL indicate that previous findings by Kayser et al.~\cite{kayser2021vil} about slight benefits of training task and explanation of e-UG jointly are likely invalid and instead caused by variances in training and initialization.

On the e-SNLI-VE dataset, we show that removing the additional inputs cumulatively increases the faithfulness on the linguistic modality, additionally verifying the utility of the Attribution-Similarity metric. However, on other modalities and on the VQA-X dataset, the removal of the additional inputs does not show the intended effect of increasing the faithfulness score. We conclude that the development of further methods that are regularizing the training process and changes to the architecture are required to build trustworthy vision-language models that faithfully explain their predictions via intuitively understandable natural language explanations.

\section{Future Work}

The conclusions of the presented results call attention to various promising avenues of future research. First of all, a comprehensive comparison of further model and dataset combinations on the proposed faithfulness metrics would provide valuable insights into the faithfulness of existing models and into the differences across datasets. Additionally, future work should focus on designing more faithful VL-NLE models, as well as improving the NLE-faithfulness metrics that have been proposed so far.

\paragraph{Improving NLE-faithfulness metrics.} Due to the pre-processed features used in the e-UG model, we decided to attribute the visual modality relevance to these features instead of actual image pixels as this allowed us to use the same model as the original authors for comparability. However, this limits a fair comparison when using the proposed faithfulness metrics on other models, which use a different set of image features. Therefore, when comparing different models on the e-ViL benchmark, the faithfulness scores and image feature extractors should be re-implemented to enable an end-to-end attribution to image pixels. As a conceptual improvement to the Cosine-Faithfulness score, we discussed filtering out the relevance attributed to explanation tokens that are generated primarily from ``internal knowledge'' of the explanation model and could introduce noise to the overall score. One possibility that could be explored is removing the attribution of tokens where the overall sum of the relevance of a modality falls beneath a certain threshold.

\paragraph{Designing more faithful VL-NLE models.} The overall architecture of previous models that have been evaluated on VL-NLE tasks is, on a high level, characterized by distinct modules for task prediction and explanation generation. By design, it is more difficult to learn the generation of faithful explanations in this setup, especially when the modules are pre-trained separately and if the explanation generator receives additional input to attend to. Instead, we believe that monolithic, generative models that generate prediction and explanation in the same way by merging task- and explanation-module, are better suited for faithful NLE generation. Concurrent work by Sammani et al.~\cite{sammani2022nlx} followed this line of reasoning and claimed that their model is more faithful due to this architecture. However, these claims should be, on one hand, backed by assessing the faithfulness with available metrics, such as the ones presented in this work. Further, our results indicate that bringing task prediction and explanation generation closer together might not be sufficient, hence other interventions to the model architecture and training process might be required to incentivize models to learn faithful explanations. Some ideas discussed in this work that we find promising are (i) biasing the explanation generation attention on the attention utilized during task prediction and (ii), as demonstrated by Wu and Mooney~\cite{wu2018faithful}, introducing a faithfulness loss that enforces the similarity of feature relevance on inputs or intermediate feature representations. 

\cleardoublepage

\fancyhead[LE,RO]{\it Bibliography}       
    \bibliographystyle{plain}             
    \addcontentsline{toc}{chapter}{Bibliography}
    \bibliography{thesis}

\begin{thebibliography}{10}

\bibitem{ijcai2022-90}
Leila Amgoud and Jonathan Ben-Naim.
\newblock Axiomatic foundations of explainability.
\newblock In Lud~De Raedt, editor, {\em Proceedings of the Thirty-First
  International Joint Conference on Artificial Intelligence, {IJCAI-22}}, pages
  636--642. International Joint Conferences on Artificial Intelligence
  Organization, 7 2022.
\newblock Main Track.

\bibitem{arrieta2020explainable}
Alejandro~Barredo Arrieta, Natalia D{\'\i}az-Rodr{\'\i}guez, Javier Del~Ser,
  Adrien Bennetot, Siham Tabik, Alberto Barbado, Salvador Garc{\'\i}a, Sergio
  Gil-L{\'o}pez, Daniel Molina, Richard Benjamins, et~al.
\newblock Explainable artificial intelligence (xai): Concepts, taxonomies,
  opportunities and challenges toward responsible ai.
\newblock {\em Information fusion}, 58:82--115, 2020.

\bibitem{bach2015pixel}
Sebastian Bach, Alexander Binder, Gr{\'e}goire Montavon, Frederick Klauschen,
  Klaus-Robert M{\"u}ller, and Wojciech Samek.
\newblock On pixel-wise explanations for non-linear classifier decisions by
  layer-wise relevance propagation.
\newblock {\em PloS one}, 10(7):e0130140, 2015.

\bibitem{baehrens2010explain}
David Baehrens, Timon Schroeter, Stefan Harmeling, Motoaki Kawanabe, Katja
  Hansen, and Klaus-Robert M{\"u}ller.
\newblock How to explain individual classification decisions.
\newblock {\em The Journal of Machine Learning Research}, 11:1803--1831, 2010.

\bibitem{DBLP:journals/corr/BahdanauCB14}
Dzmitry Bahdanau, Kyunghyun Cho, and Yoshua Bengio.
\newblock Neural machine translation by jointly learning to align and
  translate.
\newblock In Yoshua Bengio and Yann LeCun, editors, {\em 3rd International
  Conference on Learning Representations, {ICLR} 2015, San Diego, CA, USA, May
  7-9, 2015, Conference Track Proceedings}, 2015.

\bibitem{banerjee2005meteor}
Satanjeev Banerjee and Alon Lavie.
\newblock Meteor: An automatic metric for mt evaluation with improved
  correlation with human judgments.
\newblock In {\em Proceedings of the acl workshop on intrinsic and extrinsic
  evaluation measures for machine translation and/or summarization}, pages
  65--72, 2005.

\bibitem{blunsom2019can}
Phil Blunsom, Oana-Maria Camburu, Jakob Foerster, Eleonora Giunchiglia, and
  Thomas Lukasiewicz.
\newblock Can i trust the explainer? verifying post- hoc explanatory methods.
\newblock {\em CoRR}, 2019.

\bibitem{brown2020language}
Tom Brown, Benjamin Mann, Nick Ryder, Melanie Subbiah, Jared~D Kaplan, Prafulla
  Dhariwal, Arvind Neelakantan, Pranav Shyam, Girish Sastry, Amanda Askell,
  et~al.
\newblock Language models are few-shot learners.
\newblock {\em Advances in neural information processing systems},
  33:1877--1901, 2020.

\bibitem{camburu2020explaining}
Oana-Maria Camburu.
\newblock Explaining deep neural networks.
\newblock {\em arXiv preprint arXiv:2010.01496}, 2020.

\bibitem{camburu2018snli}
Oana-Maria Camburu, Tim Rockt{\"a}schel, Thomas Lukasiewicz, and Phil Blunsom.
\newblock e-snli: Natural language inference with natural language
  explanations.
\newblock {\em Advances in Neural Information Processing Systems}, 31, 2018.

\bibitem{chen2020uniter}
Yen-Chun Chen, Linjie Li, Licheng Yu, Ahmed El~Kholy, Faisal Ahmed, Zhe Gan,
  Yu~Cheng, and Jingjing Liu.
\newblock Uniter: Universal image-text representation learning.
\newblock In {\em European conference on computer vision}, pages 104--120.
  Springer, 2020.

\bibitem{cho2014properties}
Kyunghyun Cho, Bart van Merri{\"e}nboer, Dzmitry Bahdanau, and Yoshua Bengio.
\newblock On the properties of neural machine translation: Encoder--decoder
  approaches.
\newblock In {\em Proceedings of SSST-8, Eighth Workshop on Syntax, Semantics
  and Structure in Statistical Translation}, pages 103--111, 2014.

\bibitem{chollet2017xception}
Fran{\c{c}}ois Chollet.
\newblock Xception: Deep learning with depthwise separable convolutions.
\newblock In {\em Proceedings of the IEEE conference on computer vision and
  pattern recognition}, pages 1251--1258, 2017.

\bibitem{conneau2017supervised}
A~Conneau, D~Kiela, H~Schwenk, L~Barrault, and A~Bordes.
\newblock Supervised learning of universal sentence representations from
  natural language inference data.
\newblock In {\em Proceedings of the 2017 Conference on Empirical Methods in
  Natural Language Processing}, pages 670--680. Association for Computational
  Linguistics, 2017.

\bibitem{devlin2019bert}
Jacob Devlin, Ming-Wei Chang, Kenton Lee, and Kristina Toutanova.
\newblock Bert: Pre-training of deep bidirectional transformers for language
  understanding.
\newblock In {\em Proceedings of the 2019 Conference of the North American
  Chapter of the Association for Computational Linguistics: Human Language
  Technologies, Volume 1 (Long and Short Papers)}, pages 4171--4186, 2019.

\bibitem{deyoung2020eraser}
Jay DeYoung, Sarthak Jain, Nazneen~Fatema Rajani, Eric Lehman, Caiming Xiong,
  Richard Socher, and Byron~C Wallace.
\newblock Eraser: A benchmark to evaluate rationalized nlp models.
\newblock In {\em Proceedings of the 58th Annual Meeting of the Association for
  Computational Linguistics}, pages 4443--4458, 2020.

\bibitem{ding-etal-2019-saliency}
Shuoyang Ding, Hainan Xu, and Philipp Koehn.
\newblock Saliency-driven word alignment interpretation for neural machine
  translation.
\newblock In {\em Proceedings of the Fourth Conference on Machine Translation
  (Volume 1: Research Papers)}, pages 1--12, Florence, Italy, August 2019.
  Association for Computational Linguistics.

\bibitem{dosovitskiy2020image}
Alexey Dosovitskiy, Lucas Beyer, Alexander Kolesnikov, Dirk Weissenborn,
  Xiaohua Zhai, Thomas Unterthiner, Mostafa Dehghani, Matthias Minderer, Georg
  Heigold, Sylvain Gelly, et~al.
\newblock An image is worth 16x16 words: Transformers for image recognition at
  scale.
\newblock In {\em International Conference on Learning Representations}, 2020.

\bibitem{ehsan2018rationalization}
Upol Ehsan, Brent Harrison, Larry Chan, and Mark~O Riedl.
\newblock Rationalization: A neural machine translation approach to generating
  natural language explanations.
\newblock In {\em Proceedings of the 2018 AAAI/ACM Conference on AI, Ethics,
  and Society}, pages 81--87, 2018.

\bibitem{gehring2017convolutional}
Jonas Gehring, Michael Auli, David Grangier, Denis Yarats, and Yann~N Dauphin.
\newblock Convolutional sequence to sequence learning.
\newblock In {\em International conference on machine learning}, pages
  1243--1252. PMLR, 2017.

\bibitem{balanced_vqa_v2}
Yash Goyal, Tejas Khot, Douglas Summers{-}Stay, Dhruv Batra, and Devi Parikh.
\newblock Making the {V} in {VQA} matter: Elevating the role of image
  understanding in {V}isual {Q}uestion {A}nswering.
\newblock In {\em Conference on Computer Vision and Pattern Recognition
  (CVPR)}, 2017.

\bibitem{he-etal-2019-towards}
Shilin He, Zhaopeng Tu, Xing Wang, Longyue Wang, Michael Lyu, and Shuming Shi.
\newblock Towards understanding neural machine translation with word
  importance.
\newblock In {\em Proceedings of the 2019 Conference on Empirical Methods in
  Natural Language Processing and the 9th International Joint Conference on
  Natural Language Processing (EMNLP-IJCNLP)}, pages 953--962, Hong Kong,
  China, November 2019. Association for Computational Linguistics.

\bibitem{herman2017promise}
Bernease Herman.
\newblock The promise and peril of human evaluation for model interpretability.
\newblock {\em arXiv preprint arXiv:1711.07414}, 2017.

\bibitem{hochreiter1997long}
Sepp Hochreiter and J{\"u}rgen Schmidhuber.
\newblock Long short-term memory.
\newblock {\em Neural computation}, 9(8):1735--1780, 1997.

\bibitem{holzinger2017we}
Andreas Holzinger, Chris Biemann, Constantinos~S Pattichis, and Douglas~B Kell.
\newblock What do we need to build explainable ai systems for the medical
  domain?
\newblock {\em arXiv preprint arXiv:1712.09923}, 2017.

\bibitem{47088}
Sara Hooker, Dumitru Erhan, Pieter jan Kindermans, and Been Kim.
\newblock Evaluating feature importance estimates.
\newblock {\em arXiv}, 2018.

\bibitem{jacovi2022diagnosing}
Alon Jacovi, Jasmijn Bastings, Sebastian Gehrmann, Yoav Goldberg, and Katja
  Filippova.
\newblock Diagnosing ai explanation methods with folk concepts of behavior.
\newblock {\em arXiv preprint arXiv:2201.11239}, 2022.

\bibitem{jacovi2020towards}
Alon Jacovi and Yoav Goldberg.
\newblock Towards faithfully interpretable nlp systems: How should we define
  and evaluate faithfulness?
\newblock In {\em Proceedings of the 58th Annual Meeting of the Association for
  Computational Linguistics}, pages 4198--4205, 2020.

\bibitem{kalchbrenner2016neural}
Nal Kalchbrenner, Lasse Espeholt, Karen Simonyan, Aaron van~den Oord, Alex
  Graves, and Koray Kavukcuoglu.
\newblock Neural machine translation in linear time.
\newblock {\em arXiv preprint arXiv:1610.10099}, 2016.

\bibitem{kayser2021vil}
Maxime Kayser, Oana-Maria Camburu, Leonard Salewski, Cornelius Emde, Virginie
  Do, Zeynep Akata, and Thomas Lukasiewicz.
\newblock e-vil: A dataset and benchmark for natural language explanations in
  vision-language tasks.
\newblock In {\em Proceedings of the IEEE/CVF International Conference on
  Computer Vision}, pages 1244--1254, 2021.

\bibitem{kenton2019bert}
Jacob Devlin Ming-Wei~Chang Kenton and Lee~Kristina Toutanova.
\newblock Bert: Pre-training of deep bidirectional transformers for language
  understanding.
\newblock In {\em Proceedings of NAACL-HLT}, pages 4171--4186, 2019.

\bibitem{kokhlikyan2020captum}
Narine Kokhlikyan, Vivek Miglani, Miguel Martin, Edward Wang, Bilal Alsallakh,
  Jonathan Reynolds, Alexander Melnikov, Natalia Kliushkina, Carlos Araya, Siqi
  Yan, and Orion Reblitz-Richardson.
\newblock Captum: A unified and generic model interpretability library for
  pytorch, 2020.

\bibitem{lecun1995convolutional}
Yann LeCun, Yoshua Bengio, et~al.
\newblock Convolutional networks for images, speech, and time series.
\newblock {\em The handbook of brain theory and neural networks},
  3361(10):1995, 1995.

\bibitem{lin2014microsoft}
Tsung-Yi Lin, Michael Maire, Serge Belongie, James Hays, Pietro Perona, Deva
  Ramanan, Piotr Doll{\'a}r, and C~Lawrence Zitnick.
\newblock Microsoft coco: Common objects in context.
\newblock In {\em European conference on computer vision}, pages 740--755.
  Springer, 2014.

\bibitem{lundberg2017unified}
Scott~M Lundberg and Su-In Lee.
\newblock A unified approach to interpreting model predictions.
\newblock In {\em Proceedings of the 31st international conference on neural
  information processing systems}, pages 4768--4777, 2017.

\bibitem{maccartney-manning-2008-modeling}
Bill MacCartney and Christopher~D. Manning.
\newblock Modeling semantic containment and exclusion in natural language
  inference.
\newblock In {\em Proceedings of the 22nd International Conference on
  Computational Linguistics (Coling 2008)}, pages 521--528, Manchester, UK,
  August 2008. Coling 2008 Organizing Committee.

\bibitem{marasovic2020natural}
Ana Marasovi{\'c}, Chandra Bhagavatula, Jae sung Park, Ronan Le~Bras, Noah~A
  Smith, and Yejin Choi.
\newblock Natural language rationales with full-stack visual reasoning: From
  pixels to semantic frames to commonsense graphs.
\newblock In {\em Findings of the Association for Computational Linguistics:
  EMNLP 2020}, pages 2810--2829, 2020.

\bibitem{miller2019explanation}
Tim Miller.
\newblock Explanation in artificial intelligence: Insights from the social
  sciences.
\newblock {\em Artificial intelligence}, 267:1--38, 2019.

\bibitem{mittelstadt2019explaining}
Brent Mittelstadt, Chris Russell, and Sandra Wachter.
\newblock Explaining explanations in ai.
\newblock In {\em Proceedings of the conference on fairness, accountability,
  and transparency}, pages 279--288, 2019.

\bibitem{park2018multimodal}
Dong~Huk Park, Lisa~Anne Hendricks, Zeynep Akata, Anna Rohrbach, Bernt Schiele,
  Trevor Darrell, and Marcus Rohrbach.
\newblock Multimodal explanations: Justifying decisions and pointing to the
  evidence.
\newblock In {\em Proceedings of the IEEE conference on computer vision and
  pattern recognition}, pages 8779--8788, 2018.

\bibitem{radford2018improving}
Alec Radford, Karthik Narasimhan, Tim Salimans, Ilya Sutskever, et~al.
\newblock Improving language understanding by generative pre-training.
\newblock 2018.

\bibitem{radford2019language}
Alec Radford, Jeffrey Wu, Rewon Child, David Luan, Dario Amodei, Ilya
  Sutskever, et~al.
\newblock Language models are unsupervised multitask learners.
\newblock {\em OpenAI blog}, 1(8):9, 2019.

\bibitem{rajani2019explain}
Nazneen~Fatema Rajani, Bryan McCann, Caiming Xiong, and Richard Socher.
\newblock Explain yourself! leveraging language models for commonsense
  reasoning.
\newblock In {\em Proceedings of the 57th Annual Meeting of the Association for
  Computational Linguistics}, pages 4932--4942, 2019.

\bibitem{ramesh2021zero}
Aditya Ramesh, Mikhail Pavlov, Gabriel Goh, Scott Gray, Chelsea Voss, Alec
  Radford, Mark Chen, and Ilya Sutskever.
\newblock Zero-shot text-to-image generation.
\newblock In {\em International Conference on Machine Learning}, pages
  8821--8831. PMLR, 2021.

\bibitem{ren2015faster}
Shaoqing Ren, Kaiming He, Ross Girshick, and Jian Sun.
\newblock Faster r-cnn: Towards real-time object detection with region proposal
  networks.
\newblock {\em Advances in neural information processing systems}, 28, 2015.

\bibitem{ribeiro2016should}
Marco~Tulio Ribeiro, Sameer Singh, and Carlos Guestrin.
\newblock " why should i trust you?" explaining the predictions of any
  classifier.
\newblock In {\em Proceedings of the 22nd ACM SIGKDD international conference
  on knowledge discovery and data mining}, pages 1135--1144, 2016.

\bibitem{sammani2022nlx}
Fawaz Sammani, Tanmoy Mukherjee, and Nikos Deligiannis.
\newblock Nlx-gpt: A model for natural language explanations in vision and
  vision-language tasks.
\newblock In {\em Proceedings of the IEEE/CVF Conference on Computer Vision and
  Pattern Recognition}, pages 8322--8332, 2022.

\bibitem{selvaraju2017grad}
Ramprasaath~R Selvaraju, Michael Cogswell, Abhishek Das, Ramakrishna Vedantam,
  Devi Parikh, and Dhruv Batra.
\newblock Grad-cam: Visual explanations from deep networks via gradient-based
  localization.
\newblock In {\em Proceedings of the IEEE international conference on computer
  vision}, pages 618--626, 2017.

\bibitem{shapley1997value}
Lloyd~S Shapley.
\newblock A value for n-person games.
\newblock {\em Classics in game theory}, 69, 1997.

\bibitem{sturmfels2020visualizing}
Pascal Sturmfels, Scott Lundberg, and Su-In Lee.
\newblock Visualizing the impact of feature attribution baselines.
\newblock {\em Distill}, 2020.
\newblock https://distill.pub/2020/attribution-baselines.

\bibitem{sundararajan2020many}
Mukund Sundararajan and Amir Najmi.
\newblock The many shapley values for model explanation.
\newblock In {\em International conference on machine learning}, pages
  9269--9278. PMLR, 2020.

\bibitem{sundararajan2017axiomatic}
Mukund Sundararajan, Ankur Taly, and Qiqi Yan.
\newblock Axiomatic attribution for deep networks.
\newblock In {\em International conference on machine learning}, pages
  3319--3328. PMLR, 2017.

\bibitem{sutskever2014sequence}
Ilya Sutskever, Oriol Vinyals, and Quoc~V Le.
\newblock Sequence to sequence learning with neural networks.
\newblock {\em Advances in neural information processing systems}, 27, 2014.

\bibitem{vaswani2017attention}
Ashish Vaswani, Noam Shazeer, Niki Parmar, Jakob Uszkoreit, Llion Jones,
  Aidan~N Gomez, {\L}ukasz Kaiser, and Illia Polosukhin.
\newblock Attention is all you need.
\newblock {\em Advances in neural information processing systems}, 30, 2017.

\bibitem{wiegreffe2021measuring}
Sarah Wiegreffe, Ana Marasovi{\'c}, and Noah~A Smith.
\newblock Measuring association between labels and free-text rationales.
\newblock In {\em Proceedings of the 2021 Conference on Empirical Methods in
  Natural Language Processing}, pages 10266--10284, 2021.

\bibitem{wiegreffe2020attention}
Sarah Wiegreffe and Yuval Pinter.
\newblock Attention is not not explanation.
\newblock In {\em 2019 Conference on Empirical Methods in Natural Language
  Processing and 9th International Joint Conference on Natural Language
  Processing, EMNLP-IJCNLP 2019}, pages 11--20. Association for Computational
  Linguistics, 2020.

\bibitem{wu2018faithful}
Jialin Wu and Raymond~J Mooney.
\newblock Faithful multimodal explanation for visual question answering.
\newblock {\em arXiv preprint arXiv:1809.02805}, 2018.

\bibitem{xie2019visual}
Ning Xie, Farley Lai, Derek Doran, and Asim Kadav.
\newblock Visual entailment: A novel task for fine-grained image understanding.
\newblock {\em arXiv preprint arXiv:1901.06706}, 2019.

\bibitem{yu2019rethinking}
Mo~Yu, Shiyu Chang, Yang Zhang, and Tommi~S Jaakkola.
\newblock Rethinking cooperative rationalization: Introspective extraction and
  complement control.
\newblock {\em arXiv preprint arXiv:1910.13294}, 2019.

\bibitem{zaidan2007using}
Omar Zaidan, Jason Eisner, and Christine Piatko.
\newblock Using “annotator rationales” to improve machine learning for text
  categorization.
\newblock In {\em Human language technologies 2007: The conference of the North
  American chapter of the association for computational linguistics;
  proceedings of the main conference}, pages 260--267, 2007.

\bibitem{zellers2019recognition}
Rowan Zellers, Yonatan Bisk, Ali Farhadi, and Yejin Choi.
\newblock From recognition to cognition: Visual commonsense reasoning.
\newblock In {\em Proceedings of the IEEE/CVF conference on computer vision and
  pattern recognition}, pages 6720--6731, 2019.

\bibitem{zhang2019bertscore}
Tianyi Zhang, Varsha Kishore, Felix Wu, Kilian~Q Weinberger, and Yoav Artzi.
\newblock Bertscore: Evaluating text generation with bert.
\newblock {\em arXiv preprint arXiv:1904.09675}, 2019.

\bibitem{zhou2016learning}
Bolei Zhou, Aditya Khosla, Agata Lapedriza, Aude Oliva, and Antonio Torralba.
\newblock Learning deep features for discriminative localization.
\newblock In {\em Proceedings of the IEEE conference on computer vision and
  pattern recognition}, pages 2921--2929, 2016.

\end{thebibliography}
\cleardoublepage

\appendix
\fancyhead[LO,RE]{}                      

\fancyhead[LE,RO]{\it Appendix: Additional Results}                
 \chapter{Additional Results}\label{app:nomenclature}

\section{Automatic Explanation Metrics Results}
\label{appx:automatic-expl-metrics}

\begin{table}[h!!]
    \centering
    \scriptsize
    \begin{tabular}{lrrrrrrrrr}
    \toprule
    Model & BLEU1  & BLEU2  & BLEU3  & BLEU4  & METEOR & ROUGE-L & CIDEr & SPICE & BERT \\
    \midrule
    e-UG  & 0.2943 & 0.1940 & 0.1327 & 0.0933 & 0.1925 & 0.2758 & \textbf{0.8419} & 0.3392 & 0.8156 \\
    NQ    & 0.3047 & 0.2019 & 0.1382 & \textbf{0.0972} & 0.1934 & 0.2776 & 0.8308 & 0.3349 & 0.8143 \\
    NA    & \textbf{0.3074} & \textbf{0.2028} & \textbf{0.1385} & 0.0969 & \textbf{0.1958} & \textbf{0.2784} & 0.8409 & \textbf{0.3426} & \textbf{0.8158} \\
    OU    & 0.2982 & 0.1966 & 0.1339 & 0.0937 &  0.1934 & 0.2782 & 0.8310 & 0.3397 & 0.8157 \\
    NU    & 0.3041 & 0.1998 & 0.1365 & 0.0956 & 0.1915 & 0.2703 & 0.7915 & 0.3251 & 0.8119\\
    OA    & 0.1504 & 0.0722 & 0.0384 & 0.0230 & 0.0875 & 0.1601 & 0.1270 & 0.0703 & 0.7502\\
    \bottomrule
    \end{tabular}
    \caption{e-SNLI-VE automatic explanation metrics results. Best results in bold.}
    \label{tab:my_label}
\end{table}

\begin{table}[h!!]
    \centering
    \scriptsize
    \begin{tabular}{lrrrrrrrrr}
    \toprule
    Model &  BLEU1 &  BLEU2 &  BLEU3 &  BLEU4 &  METEOR &  ROUGE-L &  CIDEr &  SPICE &   BERT \\
    \midrule
    e-UG & 0.5541 & 0.4030 & 0.2908 & 0.2092 &  0.2068 &   0.4338 & 0.6667 & 0.1884 & 0.8607 \\
    NQ   & 0.5504 & 0.3965 & 0.2834 & 0.2011 &  0.2094 &   0.4296 & 0.6248 & 0.1873 & 0.8607 \\
    NA   & \textbf{0.5729} & \textbf{0.4212} & \textbf{0.3072} & \textbf{0.2256} &  \textbf{0.2210} &   \textbf{0.4505} & \textbf{0.7311} & \textbf{0.2003} & \textbf{0.8664} \\
    OU   & 0.5457 & 0.3955 & 0.2859 & 0.2072 &  0.2066 &   0.4312 & 0.6552 & 0.1826 & 0.8595 \\
    NU   & 0.5597 & 0.4127 & 0.3011 & 0.2185 &  0.2163 &   0.4441 & 0.6823 & 0.1924 & 0.8661 \\
    OA   & 0.5098 & 0.3360 & 0.2387 & 0.1717 &  0.1797 &   0.3837 & 0.5305 & 0.1454 & 0.8343 \\
    \bottomrule
    \end{tabular}
    \caption{VQA-X automatic explanation metrics results. Best results in bold.}
    \label{tab:my_label}
\end{table}

\pagebreak

\section{Faithfulness Metrics Correlation}
\label{appx:fmetrics-correlation}

\begin{table}[h!!!!!]
\centering
\begin{adjustbox}{max width=\textwidth}

\begin{tabular}{lrrrrrrr}
\toprule
\textbf{e-UG (default)} &  NLP-Suff. &  NLP-Comp. &  IMG-Suff. &  IMG-Comp &  $S_F$-NLP &  $S_F$-IMG &  $S_F$-Overall \\
\midrule
NLP-Suff.     &      1.000 &      0.015 &      0.069 &     0.006 &      0.042 &     -0.005 &          0.029 \\
NLP-Comp.     &      0.015 &      1.000 &      0.059 &     0.109 &      0.039 &      0.042 &          0.057 \\
IMG-Suff.     &      0.069 &      0.059 &      1.000 &    -0.149 &     -0.111 &     -0.044 &         -0.114 \\
IMG-Comp      &      0.006 &      0.109 &     -0.149 &     1.000 &      0.111 &      0.059 &          0.124 \\
$S_F$-NLP     &      0.042 &      0.039 &     -0.111 &     0.111 &      1.000 &     -0.010 &          0.763 \\
$S_F$-IMG     &     -0.005 &      0.042 &     -0.044 &     0.059 &     -0.010 &      1.000 &          0.639 \\
$S_F$-Overall &      0.029 &      0.057 &     -0.114 &     0.124 &      0.763 &      0.639 &          1.000 \\
\bottomrule
\end{tabular}
\end{adjustbox}

\bigskip

\begin{adjustbox}{max width=\textwidth}
\begin{tabular}{lrrrrrrr}
\toprule
\textbf{NQ} &  NLP-Suff. &  NLP-Comp. &  IMG-Suff. &  IMG-Comp &  $S_F$-NLP &  $S_F$-IMG &  $S_F$-Overall \\
\midrule
NLP-Suff.     &      1.000 &     -0.049 &      0.084 &     0.064 &      0.011 &      0.061 &          0.048 \\
NLP-Comp.     &     -0.049 &      1.000 &      0.106 &     0.080 &      0.009 &      0.083 &          0.061 \\
IMG-Suff.     &      0.084 &      0.106 &      1.000 &    -0.119 &     -0.064 &     -0.057 &         -0.085 \\
IMG-Comp      &      0.064 &      0.080 &     -0.119 &     1.000 &      0.011 &      0.043 &          0.036 \\
$S_F$-NLP     &      0.011 &      0.009 &     -0.064 &     0.011 &      1.000 &      0.020 &          0.761 \\
$S_F$-IMG     &      0.061 &      0.083 &     -0.057 &     0.043 &      0.020 &      1.000 &          0.664 \\
$S_F$-Overall &      0.048 &      0.061 &     -0.085 &     0.036 &      0.761 &      0.664 &          1.000 \\
\bottomrule
\end{tabular}
\end{adjustbox}

\caption{Pearson correlation coefficient of the faithfulness metrics on the VQA-X dataset for the default and No-additional-question model. There is no correlation ($r < 0.2$) between comprehensiveness/sufficiency results and the similarity-based $S_F$ metrics.}
\end{table}

\begin{table}[h!!!!!]
\centering
\begin{adjustbox}{max width=\textwidth}

\begin{tabular}{lrrrrrrr}
\toprule
\textbf{e-UG (default)} &  NLP-Suff. &  NLP-Comp. &  IMG-Suff. &  IMG-Comp &  $S_F$-NLP &  $S_F$-IMG &  $S_F$-Overall \\
\midrule
NLP-Suff.     &      1.000 &      0.007 &      0.049 &     0.072 &      0.133 &     -0.012 &          0.100 \\
NLP-Comp.     &      0.007 &      1.000 &     -0.055 &    -0.020 &     -0.057 &     -0.135 &         -0.124 \\
IMG-Suff.     &      0.049 &     -0.055 &      1.000 &     0.075 &     -0.081 &      0.027 &         -0.049 \\
IMG-Comp      &      0.072 &     -0.020 &      0.075 &     1.000 &     -0.020 &      0.035 &          0.004 \\
$S_F$-NLP     &      0.133 &     -0.057 &     -0.081 &    -0.020 &      1.000 &      0.013 &          0.815 \\
$S_F$-IMG     &     -0.012 &     -0.135 &      0.027 &     0.035 &      0.013 &      1.000 &          0.590 \\
$S_F$-Overall &      0.100 &     -0.124 &     -0.049 &     0.004 &      0.815 &      0.590 &          1.000 \\
\bottomrule
\end{tabular}
\end{adjustbox}

\bigskip

\begin{adjustbox}{max width=\textwidth}
\begin{tabular}{lrrrrrrr}
\toprule
\textbf{NQ} &  NLP-Suff. &  NLP-Comp. &  IMG-Suff. &  IMG-Comp &  $S_F$-NLP &  $S_F$-IMG &  $S_F$-Overall \\
\midrule
NLP-Suff.     &      1.000 &     -0.090 &      0.010 &     0.038 &      0.185 &     -0.009 &          0.129 \\
NLP-Comp.     &     -0.090 &      1.000 &     -0.031 &     0.031 &     -0.071 &     -0.007 &         -0.056 \\
IMG-Suff.     &      0.010 &     -0.031 &      1.000 &     0.039 &     -0.109 &     -0.017 &         -0.091 \\
IMG-Comp      &      0.038 &      0.031 &      0.039 &     1.000 &     -0.043 &      0.001 &         -0.031 \\
$S_F$-NLP     &      0.185 &     -0.071 &     -0.109 &    -0.043 &      1.000 &      0.014 &          0.739 \\
$S_F$-IMG     &     -0.009 &     -0.007 &     -0.017 &     0.001 &      0.014 &      1.000 &          0.684 \\
$S_F$-Overall &      0.129 &     -0.056 &     -0.091 &    -0.031 &      0.739 &      0.684 &          1.000 \\
\bottomrule
\end{tabular}
\end{adjustbox}

\caption{Pearson correlation coefficient of the faithfulness metrics correlation on the e-SNLI-VE dataset for the default and No-additional-question model. Similarly to the other dataset, no correlation ($r < 0.2$) between comprehensiveness/ sufficiency results and the similarity-based $S_F$ metrics are observed.}
\end{table}

\cleardoublepage

\fancyhead[LE,RO]{\it Appendix: Qualitative Examples}            
\chapter{Qualitative Examples of Feature Attribution and Model Inference}\label{app:examples}

This section includes qualitative examples of five task results produced by the default e-UG, No-additional-question, No-answer, and Uniter-only variants. The examples have been randomly selected from tasks correctly answered by the default e-UG variant. Attributions on the visual and linguistic inputs are visualized, in addition to the predicted label, generated explanation and faithfulness scores. Note: Due to a data loss caused by hardware failure, we cannot present Uniter-only examples, as we were unfortunately unable to reproduce the results in time for the thesis deadline.

\section{e-SNLI-VE}
\label{appx:qulaitative-examples-esnlive}

\begin{figure}[h!!!!]
    \centering
    \includegraphics[width=\textwidth, trim={2.5cm 0.0cm 2.5cm 0.0cm}, clip]{"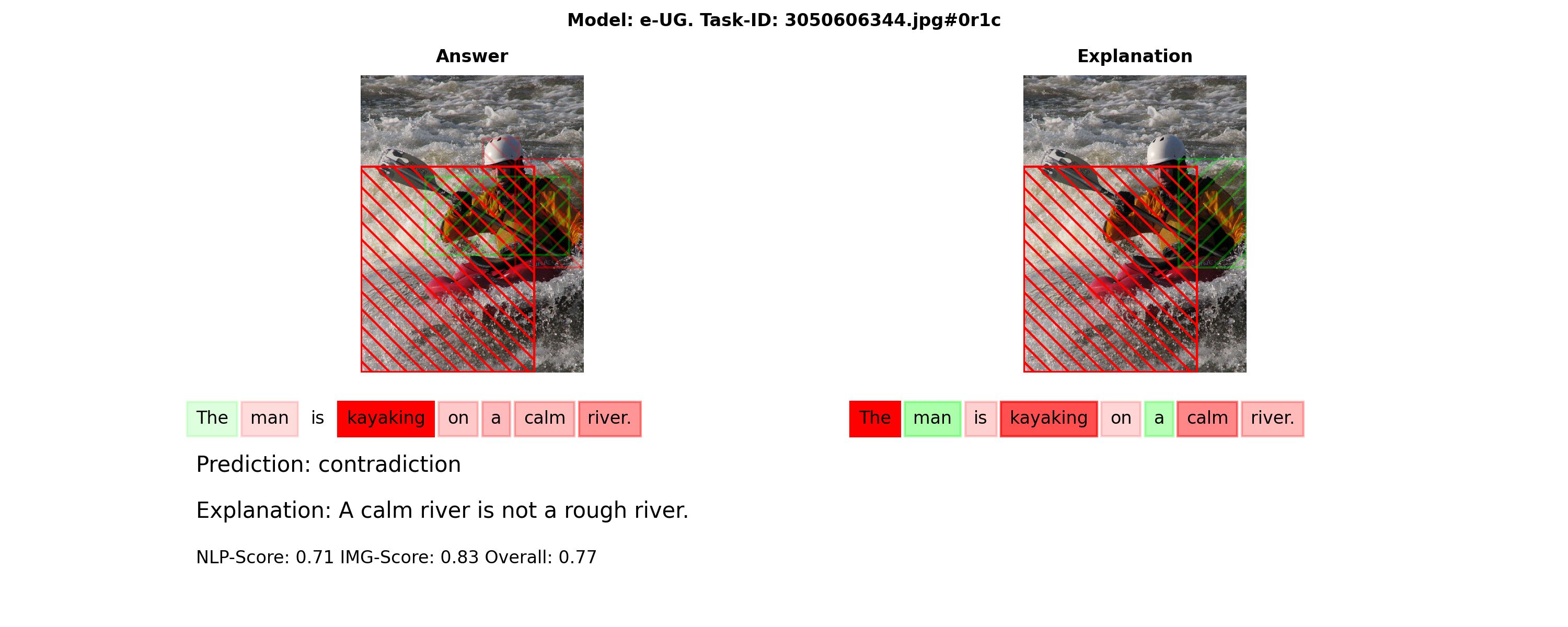"}
    \includegraphics[width=\textwidth, trim={2.5cm 0.0cm 2.5cm 0.0cm}, clip]{"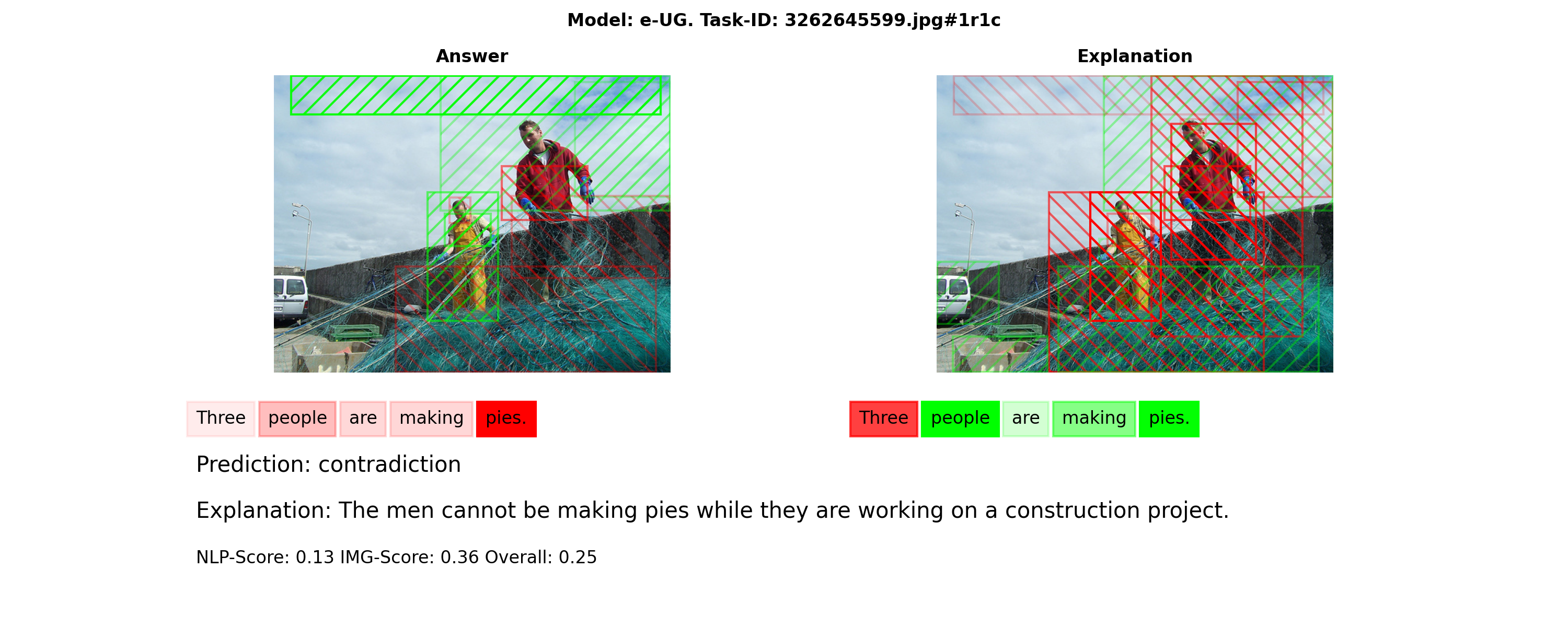"}
    \includegraphics[width=\textwidth, trim={2.5cm 0.0cm 2.5cm 0.0cm}, clip]{"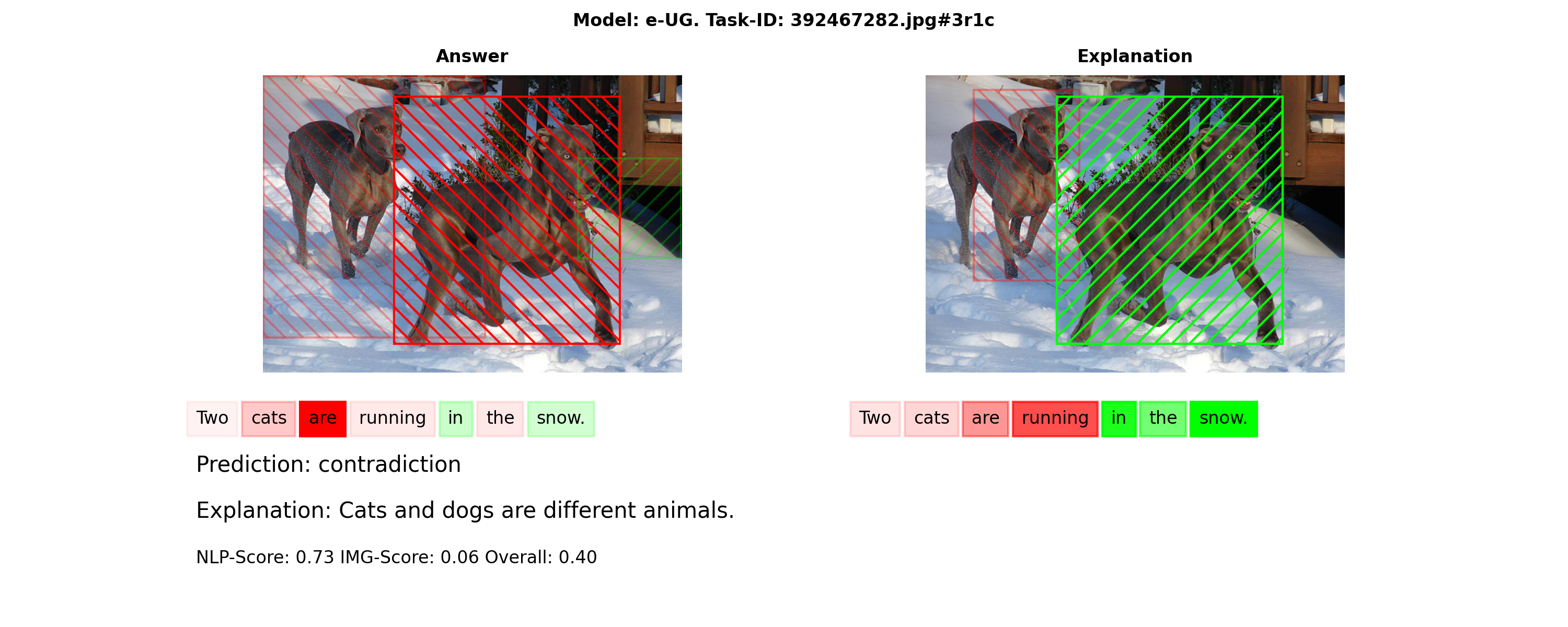"}
    \caption{e-SNLI-VE: e-UG (default) examples (1/2)}
    \label{fig:my_label}
\end{figure}

\begin{figure}[h!!!!]
    \centering
    \includegraphics[width=\textwidth, trim={2.5cm 0.0cm 2.5cm 0.0cm}, clip]{"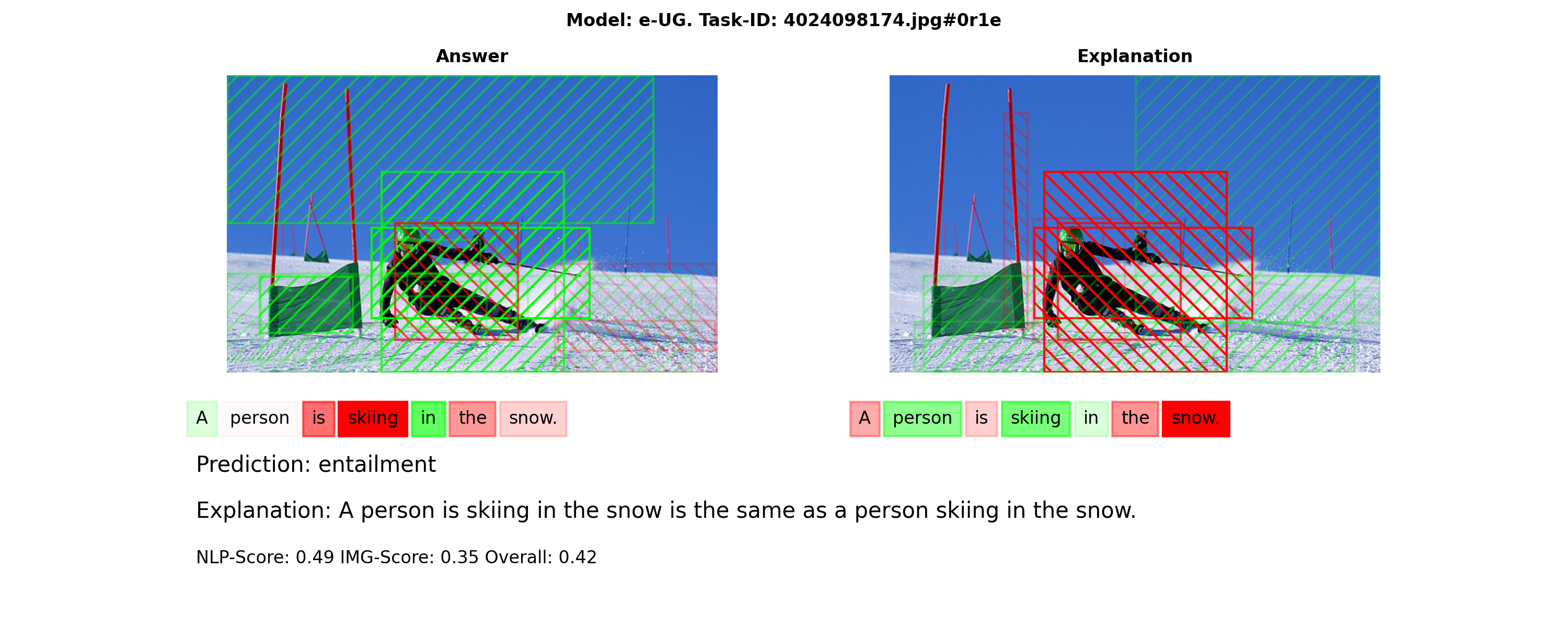"}
    \includegraphics[width=\textwidth, trim={1cm 0.0cm 1cm 0.0cm}, clip]{"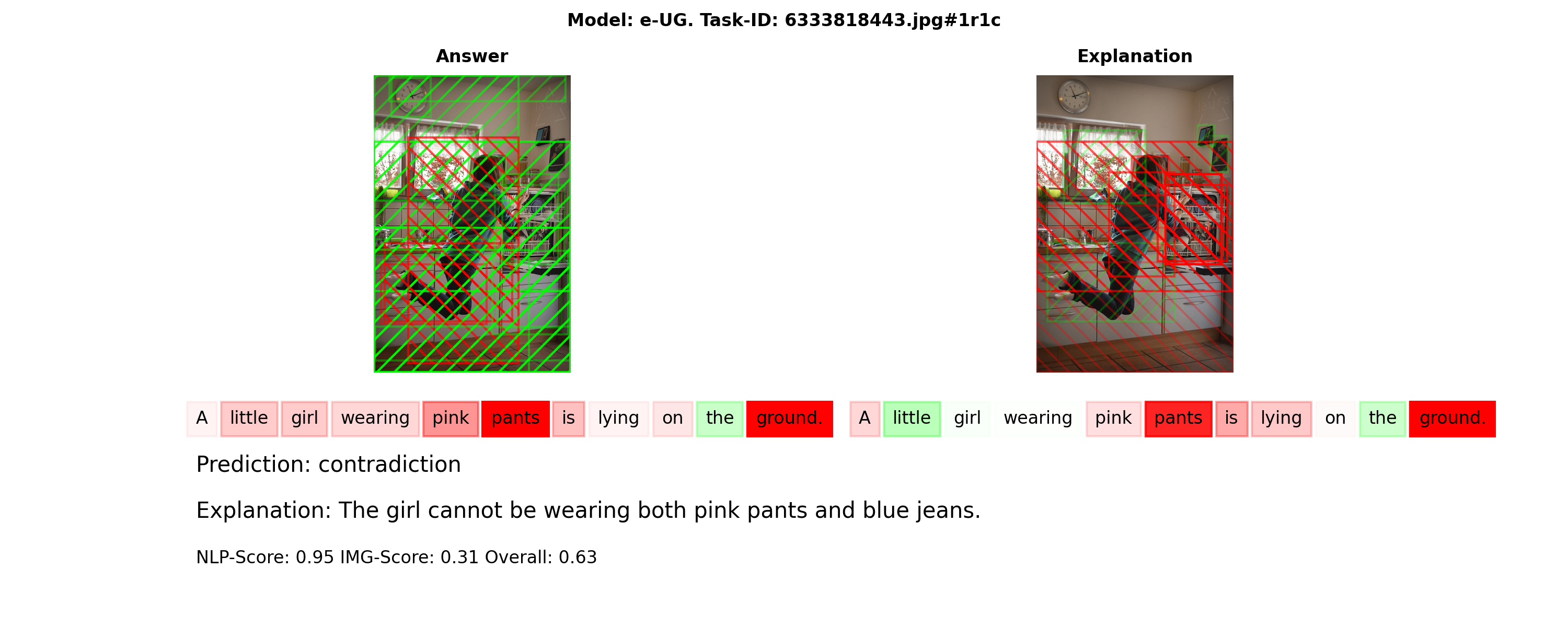"}
    \caption{e-SNLI-VE: e-UG (default) examples (2/2)}
    \label{fig:my_label}
\end{figure}

\begin{figure}[h!!!!]
    \centering
    \includegraphics[width=\textwidth, trim={2.5cm 0.0cm 2.5cm 0.0cm}, clip]{"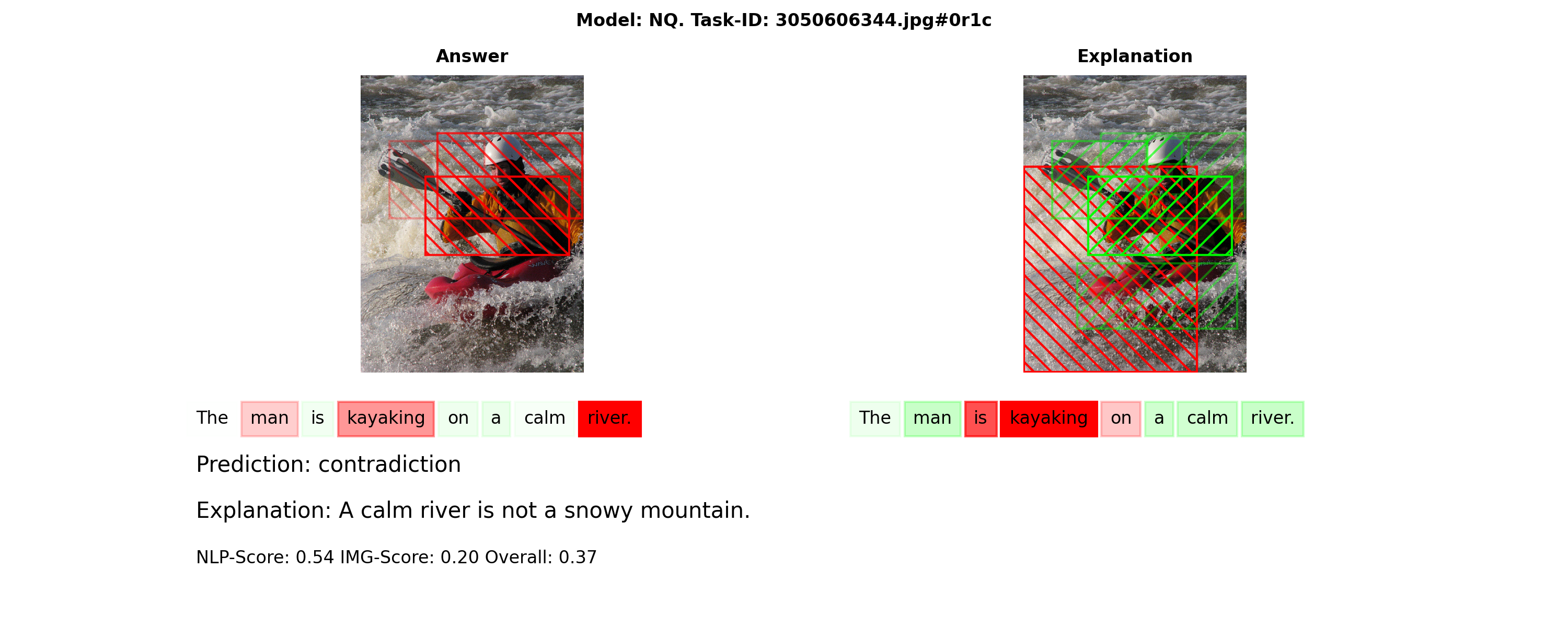"}
    \includegraphics[width=\textwidth, trim={2.5cm 0.0cm 2.5cm 0.0cm}, clip]{"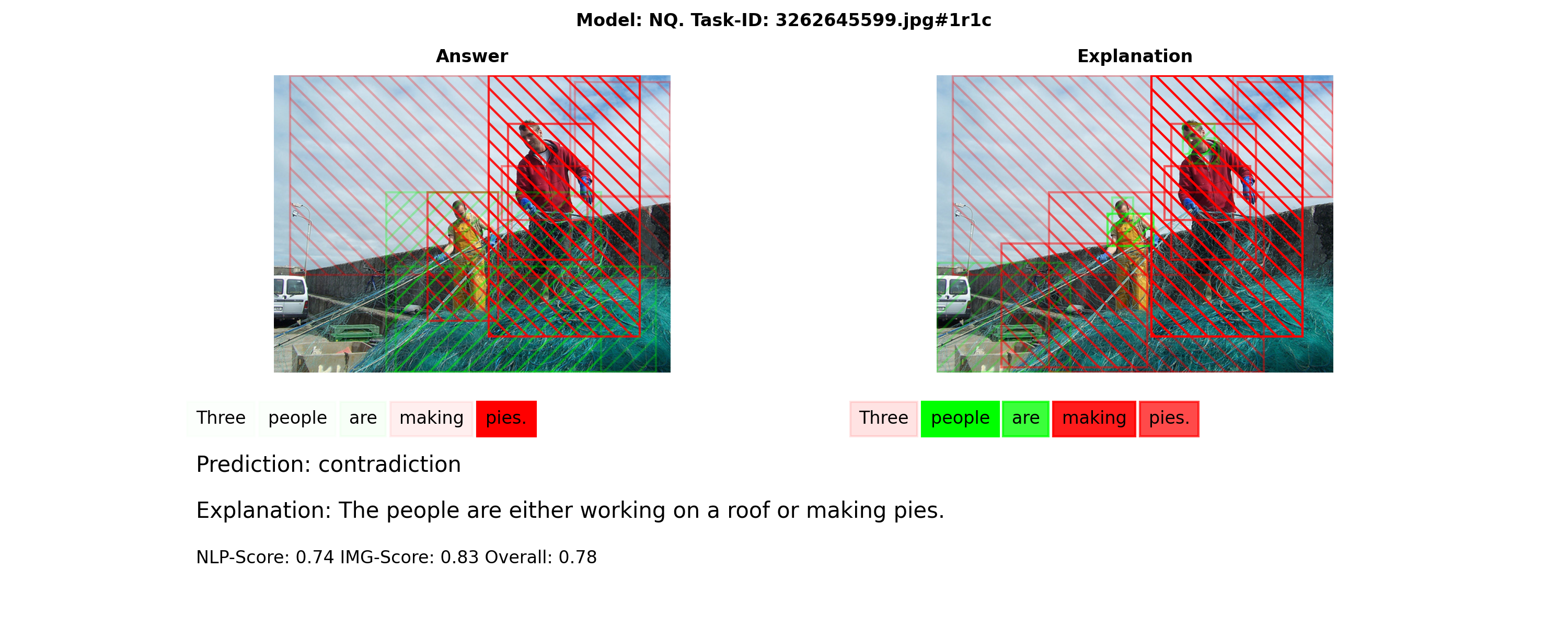"}
    \includegraphics[width=\textwidth, trim={2.5cm 0.0cm 2.5cm 0.0cm}, clip]{"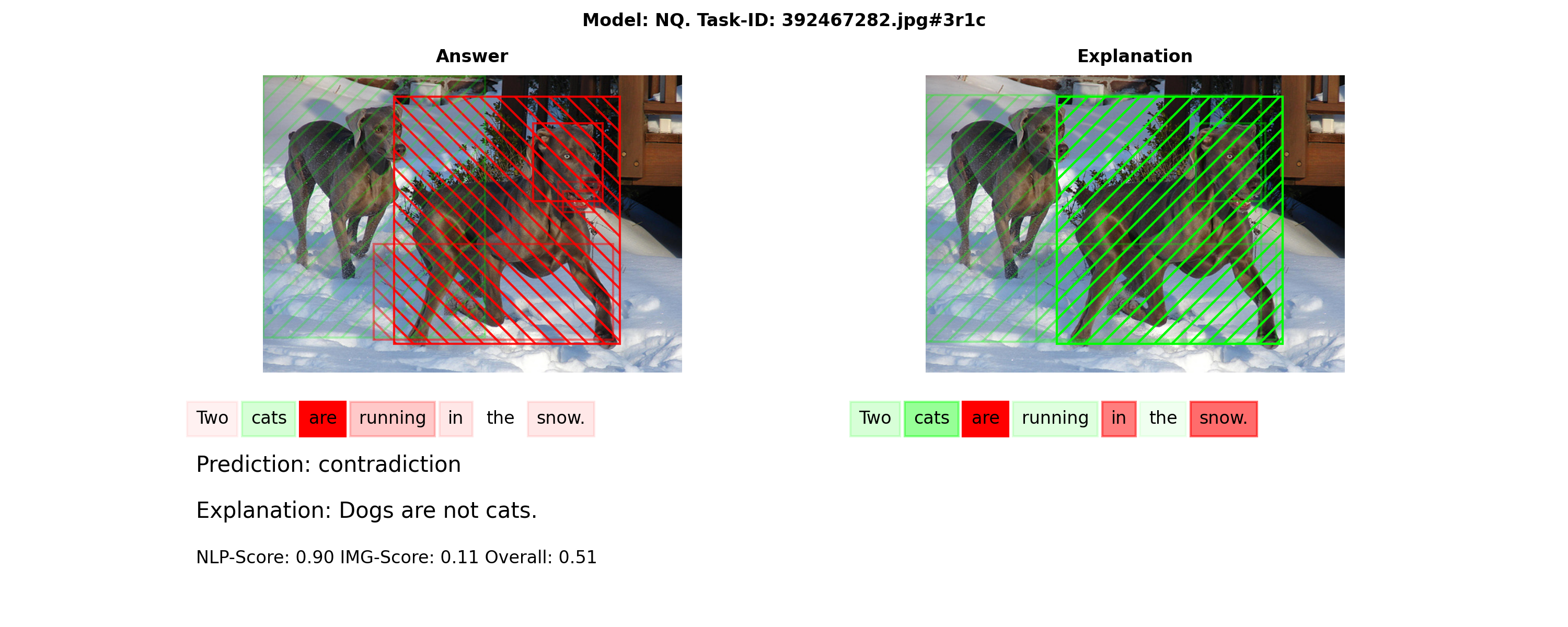"}
    \caption{e-SNLI-VE: No-additional-question (NQ) examples (1/2)}
    \label{fig:my_label}
\end{figure}

\begin{figure}[h!!!!]
    \centering
    \includegraphics[width=\textwidth, trim={2.5cm 0.0cm 2.5cm 0.0cm}, clip]{"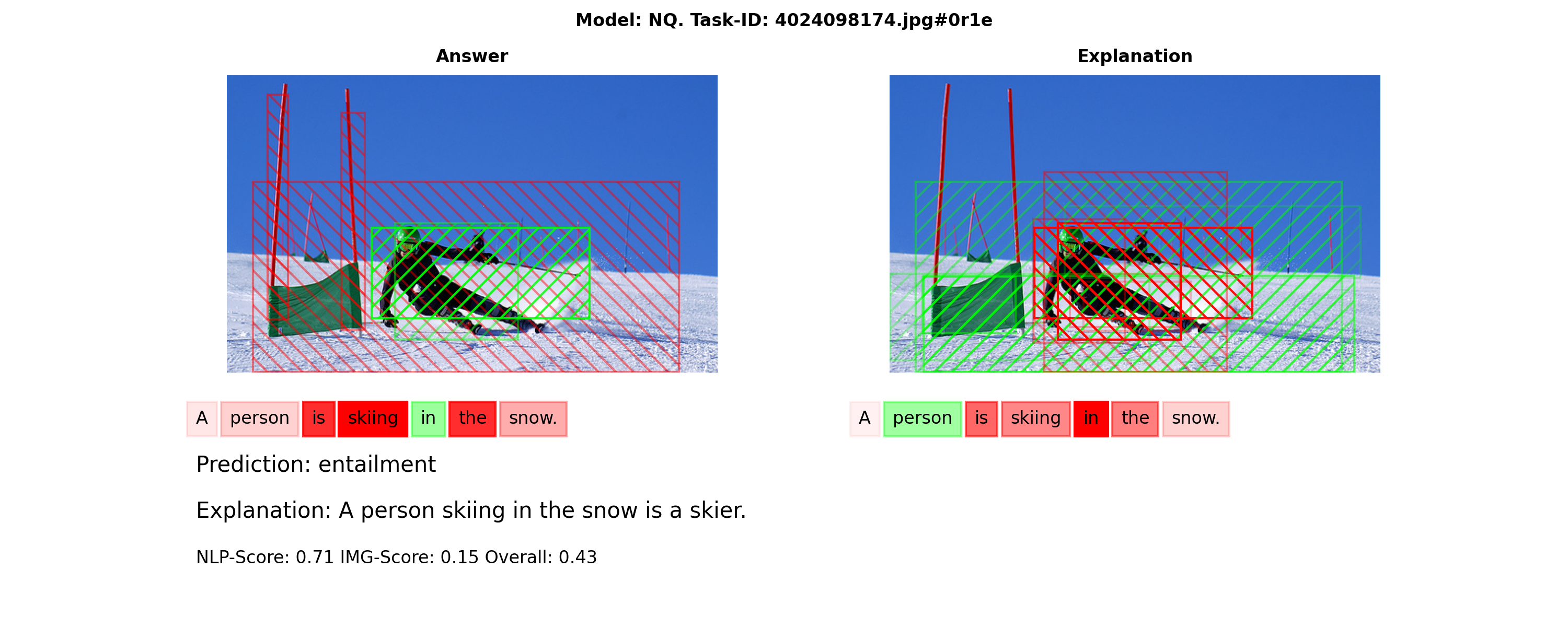"}
    \includegraphics[width=\textwidth, trim={1cm 0.0cm 1cm 0.0cm}, clip]{"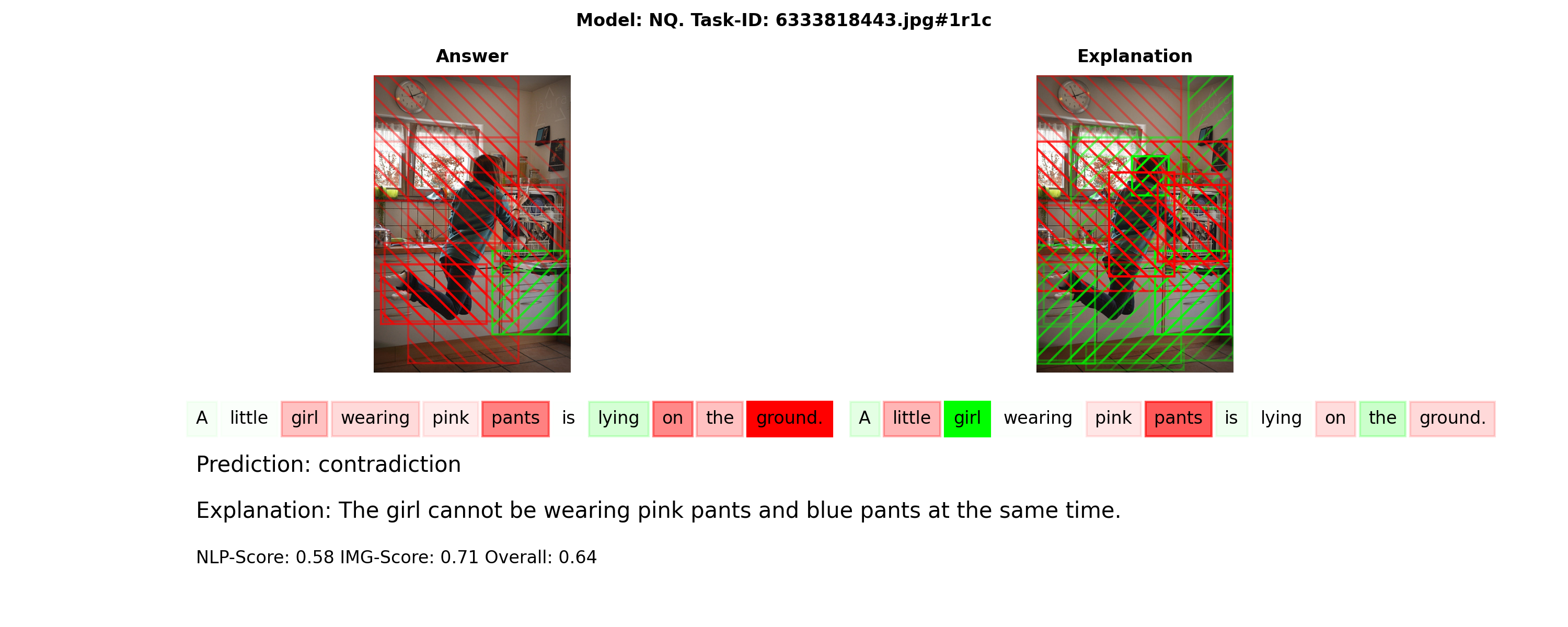"}
    \caption{e-SNLI-VE: No-additional-question (NQ) examples (2/2)}
    \label{fig:my_label}
\end{figure}

\begin{figure}[h!!!!]
    \centering
    \includegraphics[width=\textwidth, trim={2.5cm 0.0cm 2.5cm 0.0cm}, clip]{"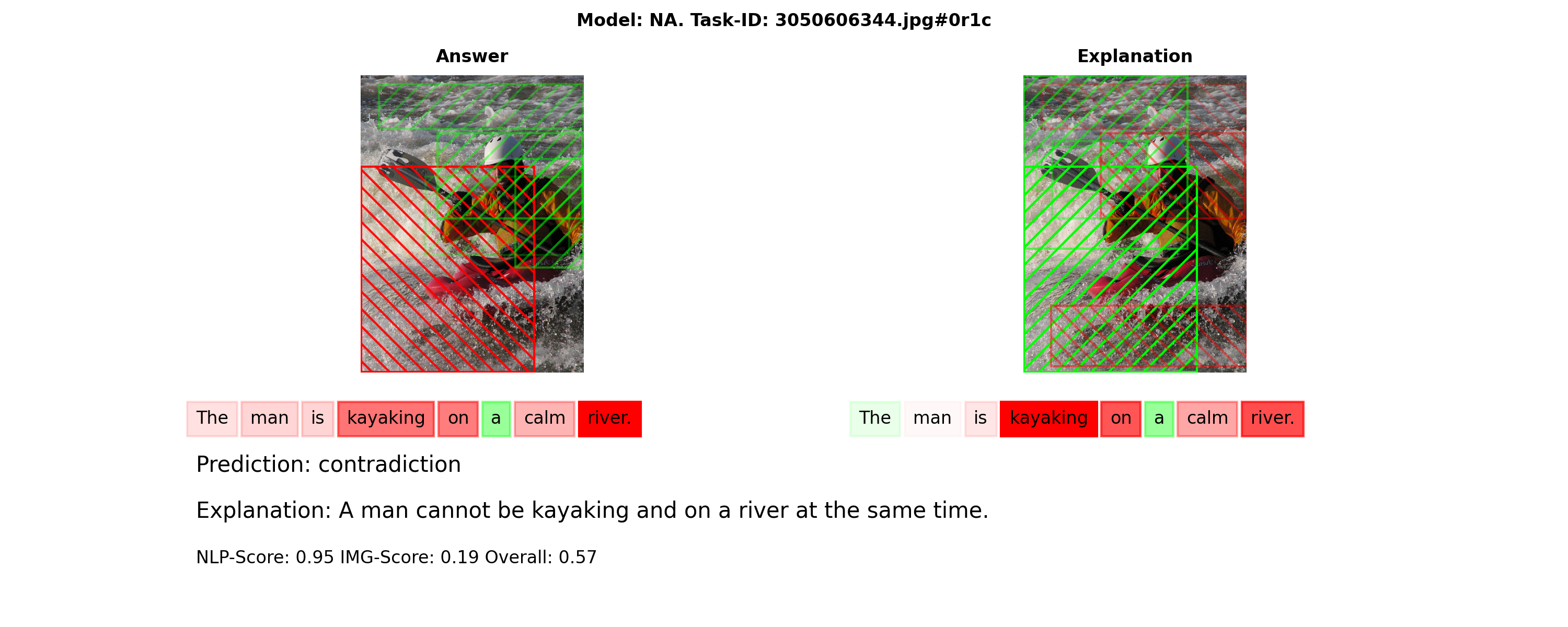"}
    \includegraphics[width=\textwidth, trim={2.5cm 0.0cm 2.5cm 0.0cm}, clip]{"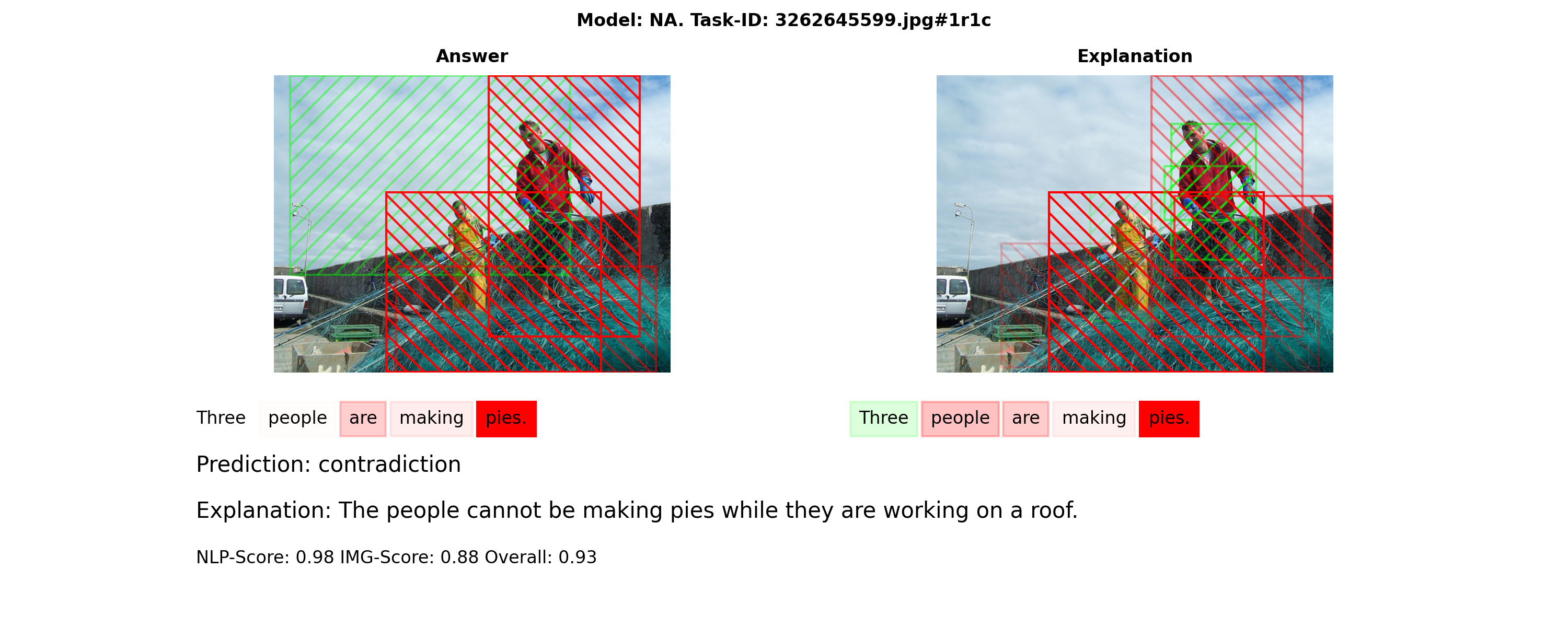"}
    \includegraphics[width=\textwidth, trim={2.5cm 0.0cm 2.5cm 0.0cm}, clip]{"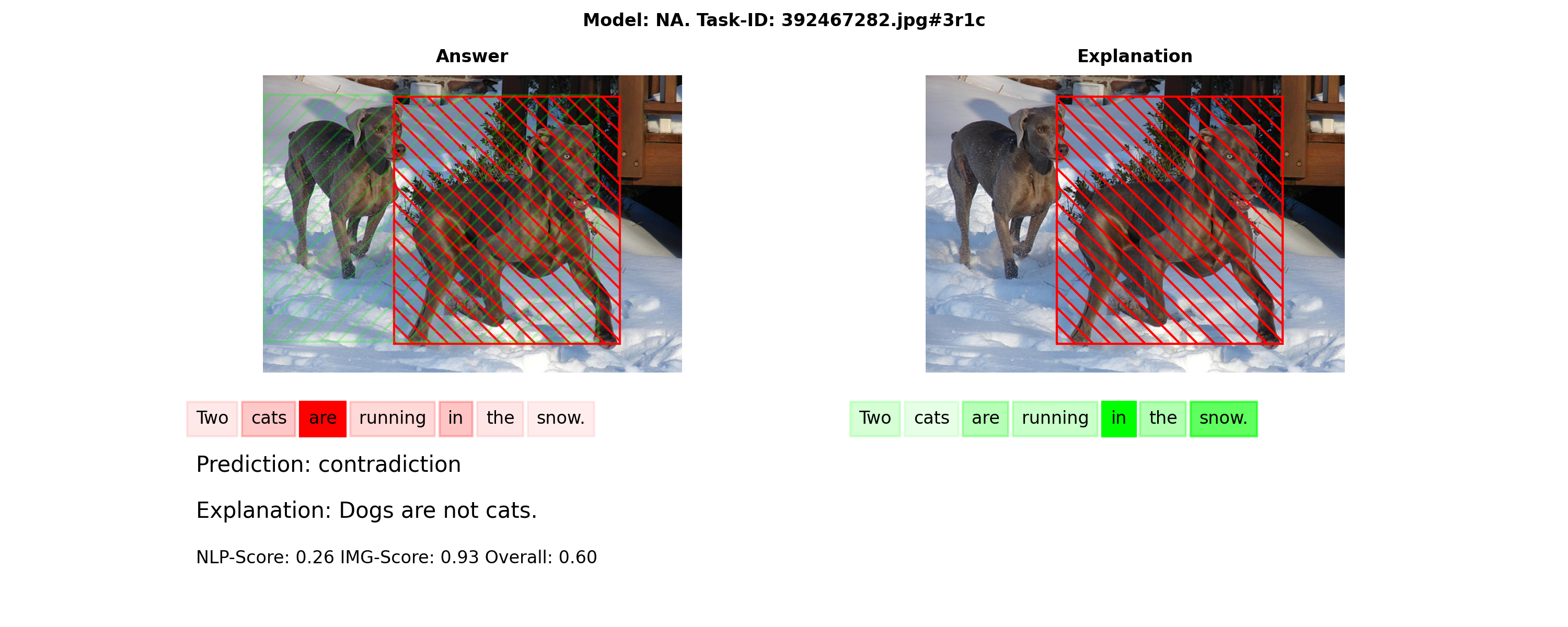"}
    \caption{e-SNLI-VE: No-answer (NA) examples (1/2)}
    \label{fig:my_label}
\end{figure}

\begin{figure}[h!!!!]
    \centering
    \includegraphics[width=\textwidth, trim={2.5cm 0.0cm 2.5cm 0.0cm}, clip]{"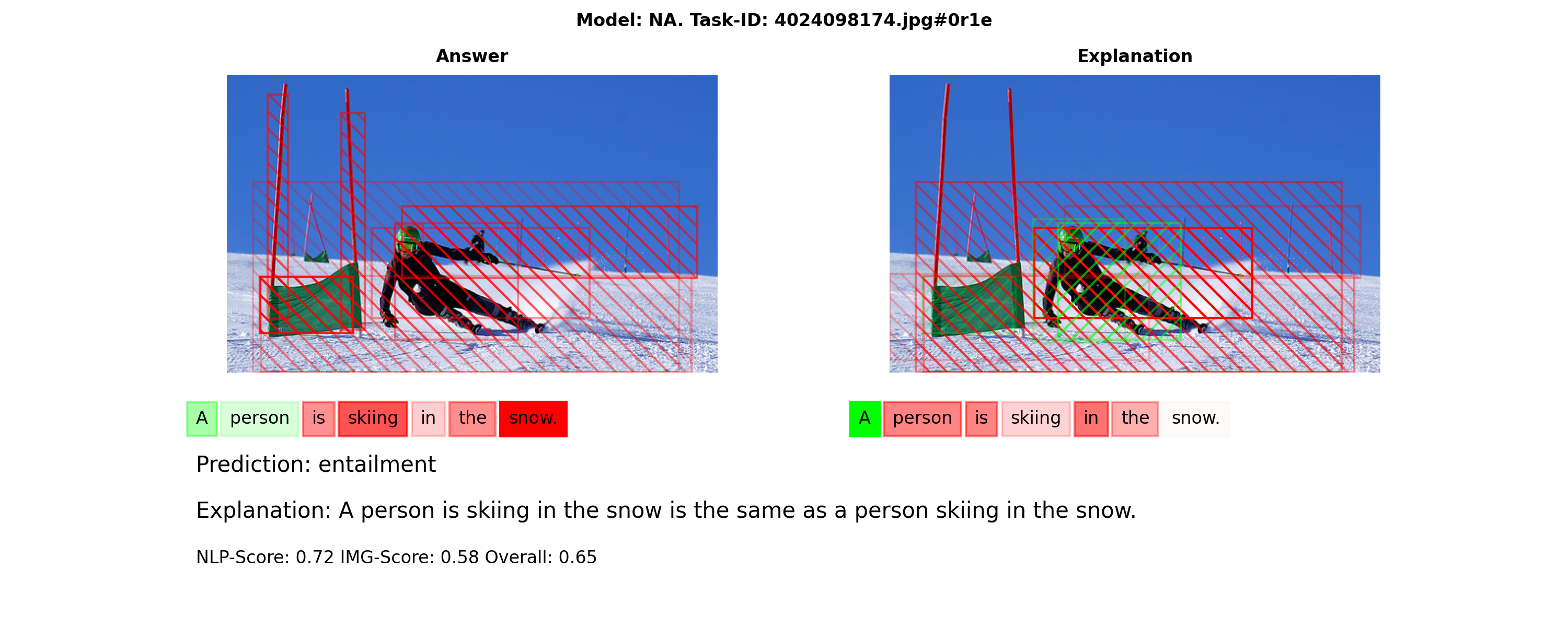"}
    \includegraphics[width=\textwidth, trim={1cm 0.0cm 1cm 0.0cm}, clip]{"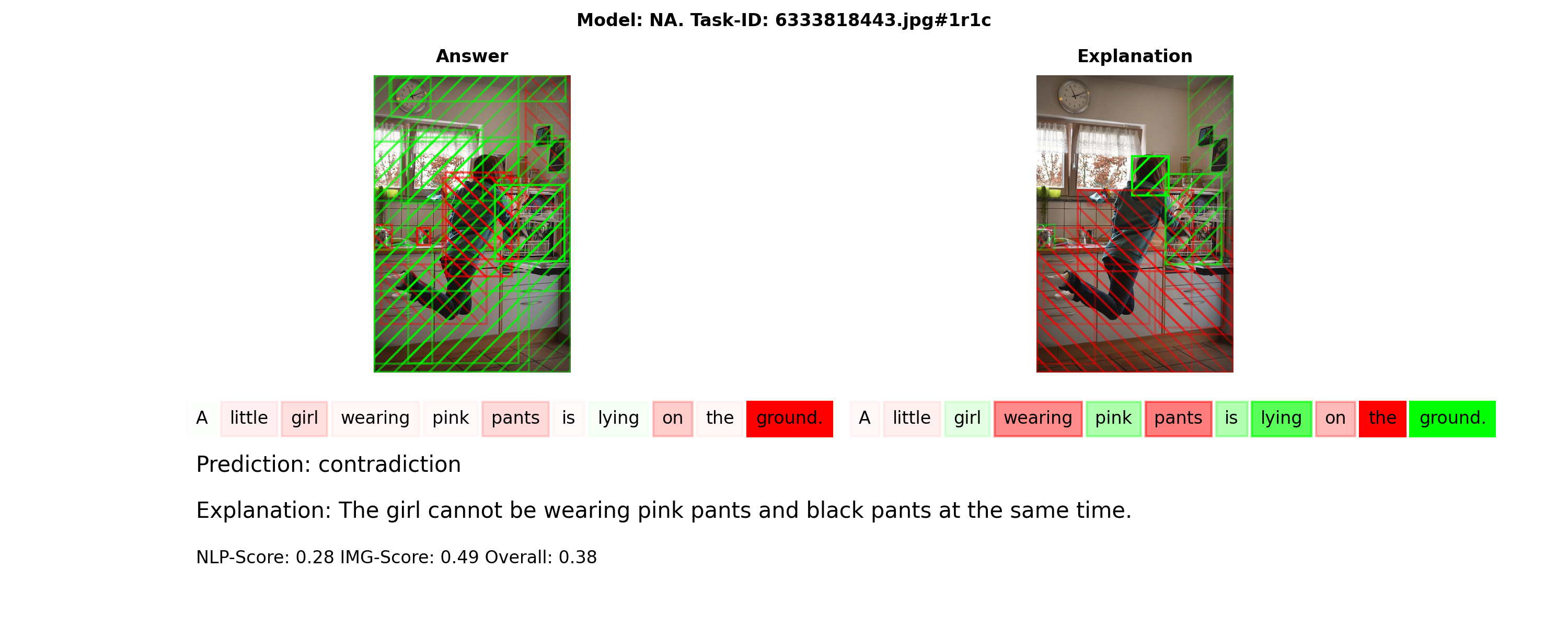"}
    \caption{e-SNLI-VE: No-answer (NA) examples (2/2)}
    \label{fig:my_label}
\end{figure}

\clearpage

\section{VQA-X}
\label{appx:qulaitative-examples-vqax}

\begin{figure}[h!!!!]
    \centering
    \includegraphics[width=\textwidth, trim={2.5cm 0.0cm 2.5cm 0.0cm}, clip]{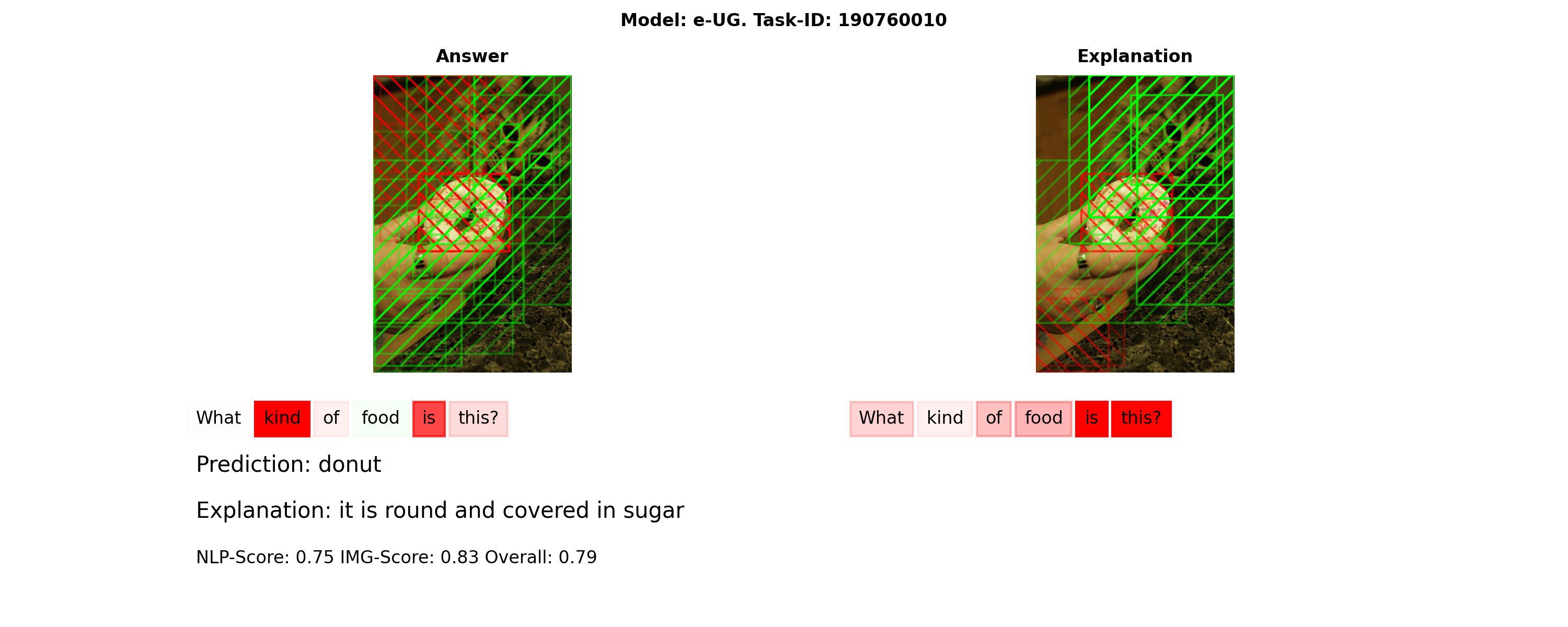}
    \includegraphics[width=\textwidth, trim={2.5cm 0.0cm 2.5cm 0.0cm}, clip]{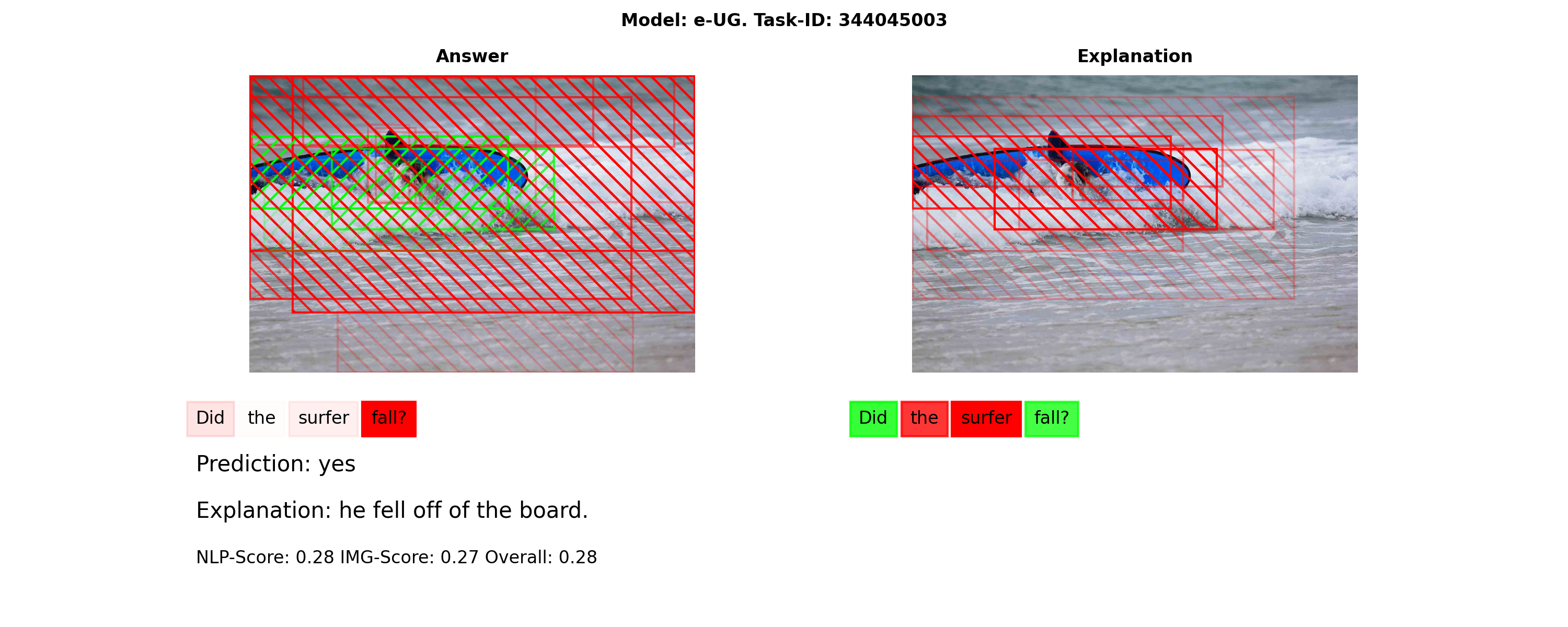}
    \includegraphics[width=\textwidth, trim={2.5cm 0.0cm 2.5cm 0.0cm}, clip]{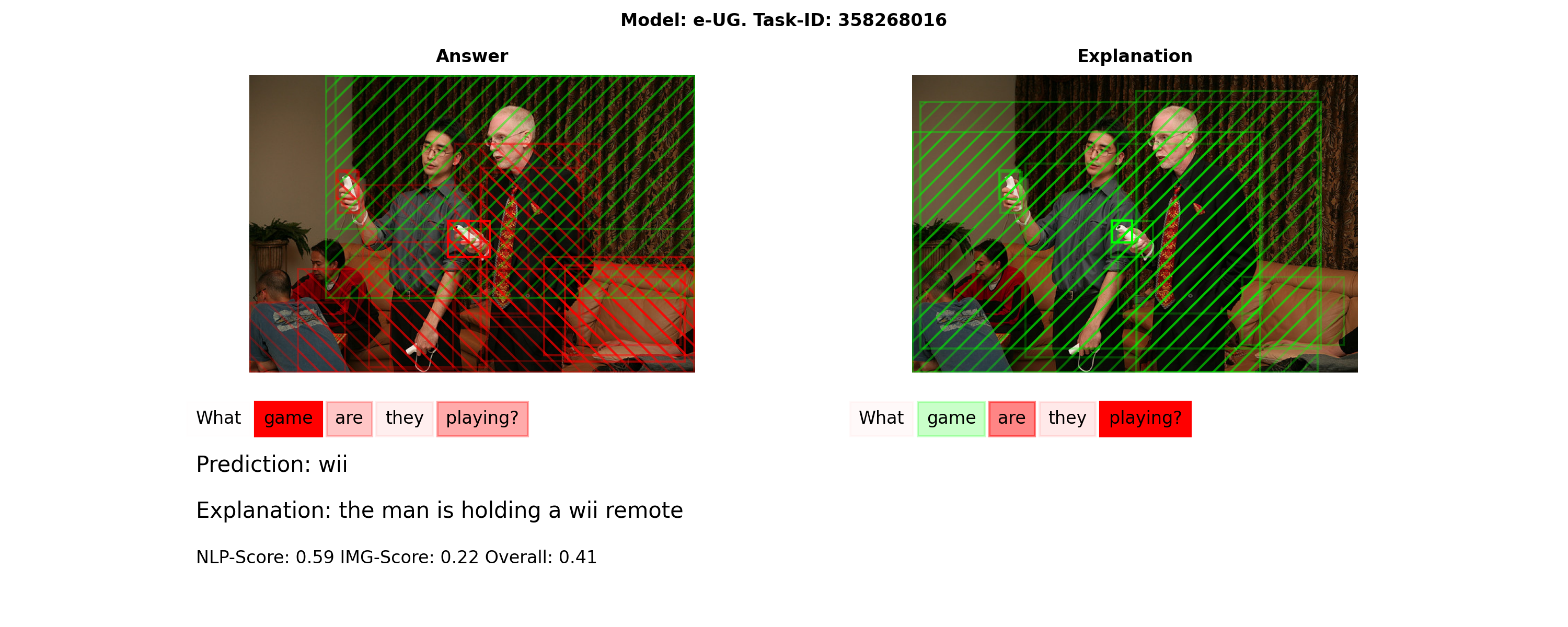}
    \caption{VQA-X: e-UG (default) examples (1/2)}
    \label{fig:my_label}
\end{figure}

\begin{figure}[h!!!!]
    \centering
    \includegraphics[width=\textwidth, trim={2.5cm 0.0cm 2.5cm 0.0cm}, clip]{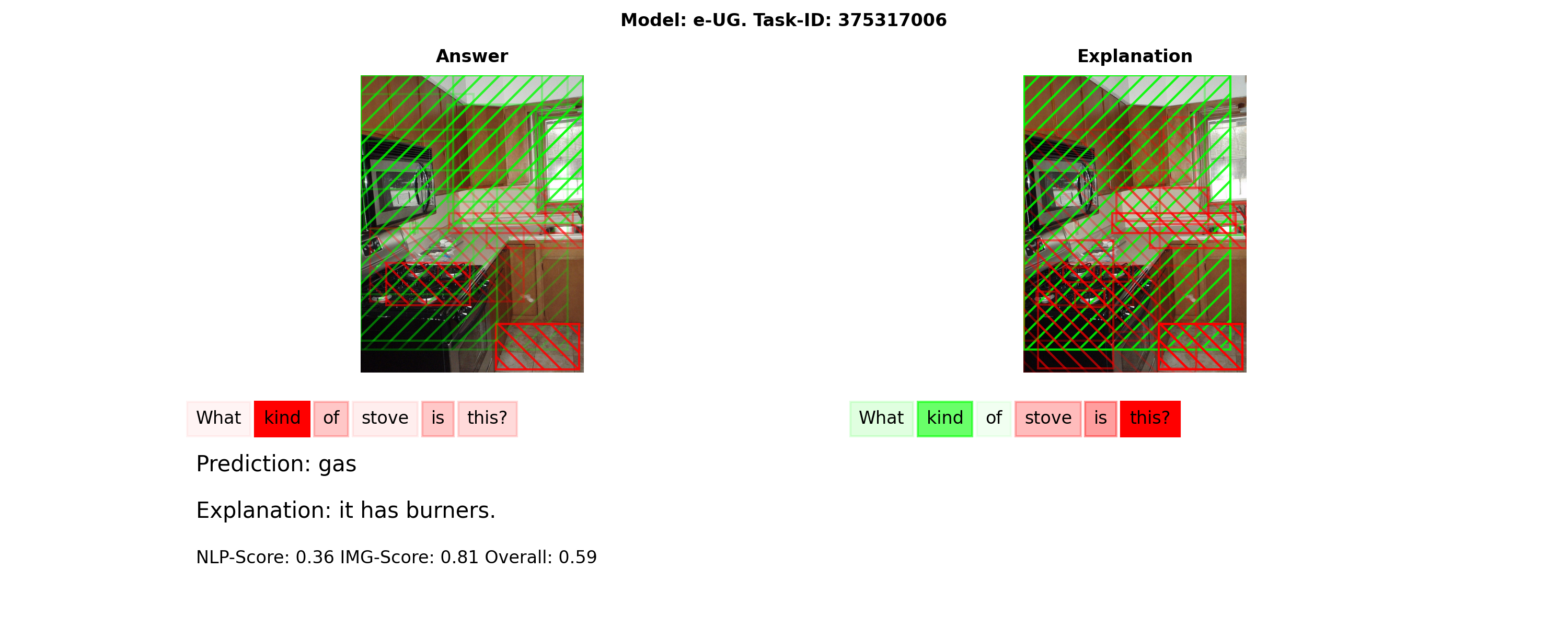}
    \includegraphics[width=\textwidth, trim={1cm 0.0cm 1cm 0.0cm}, clip]{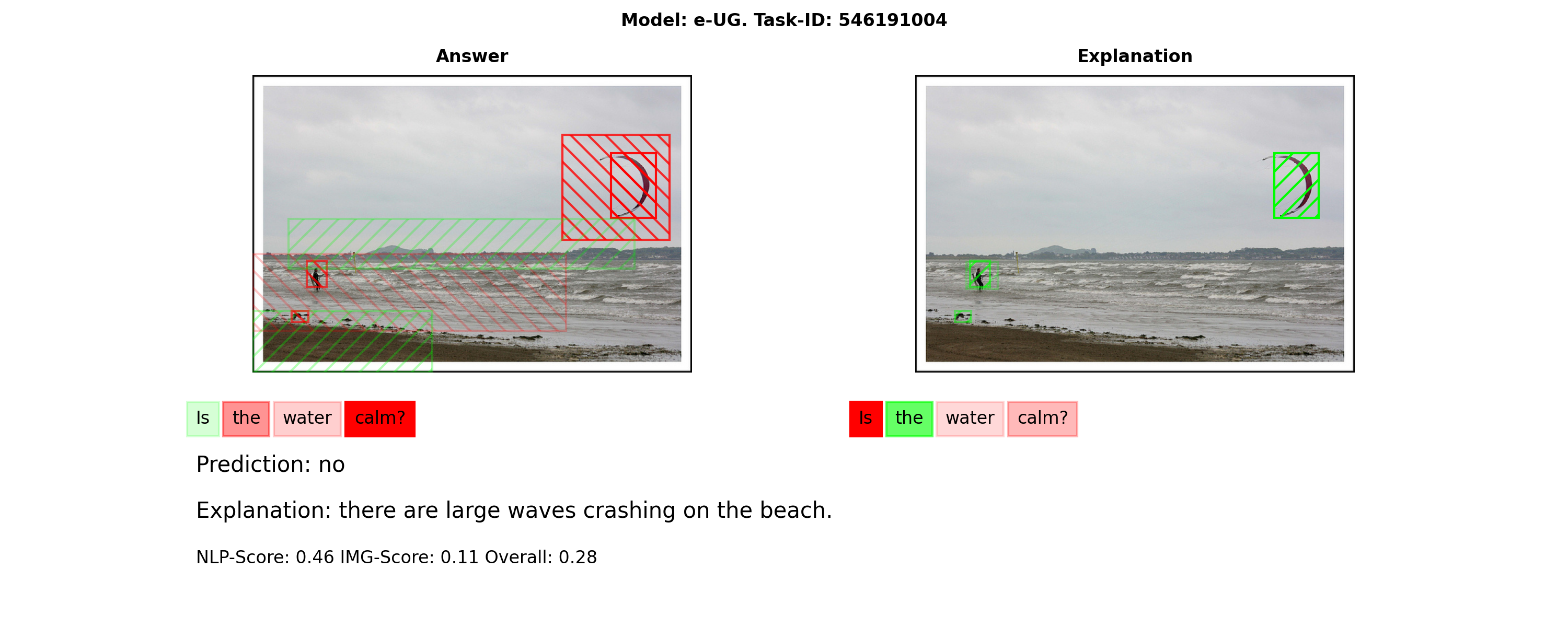}
    \caption{VQA-X: e-UG (default) examples (2/2)}
    \label{fig:my_label}
\end{figure}

\begin{figure}[h!!!!]
    \centering
    \includegraphics[width=\textwidth, trim={2.5cm 0.0cm 2.5cm 0.0cm}, clip]{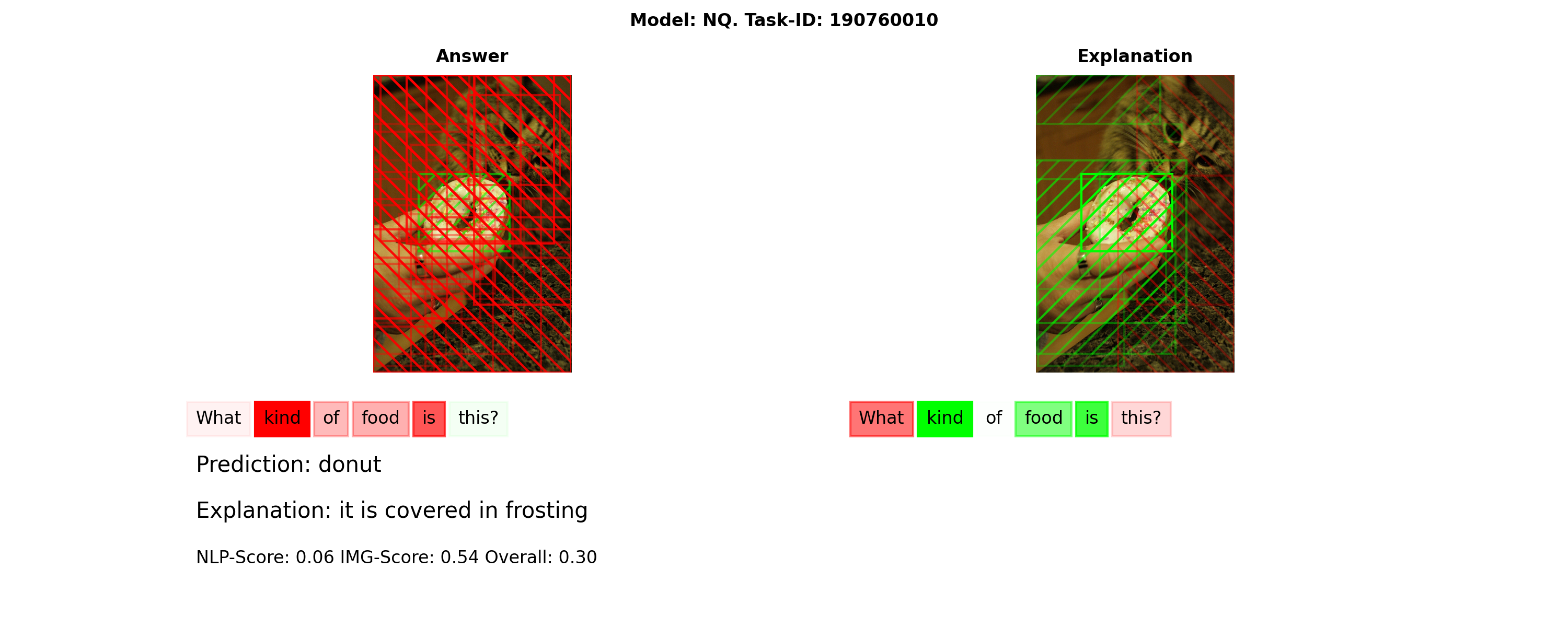}
    \includegraphics[width=\textwidth, trim={2.5cm 0.0cm 2.5cm 0.0cm}, clip]{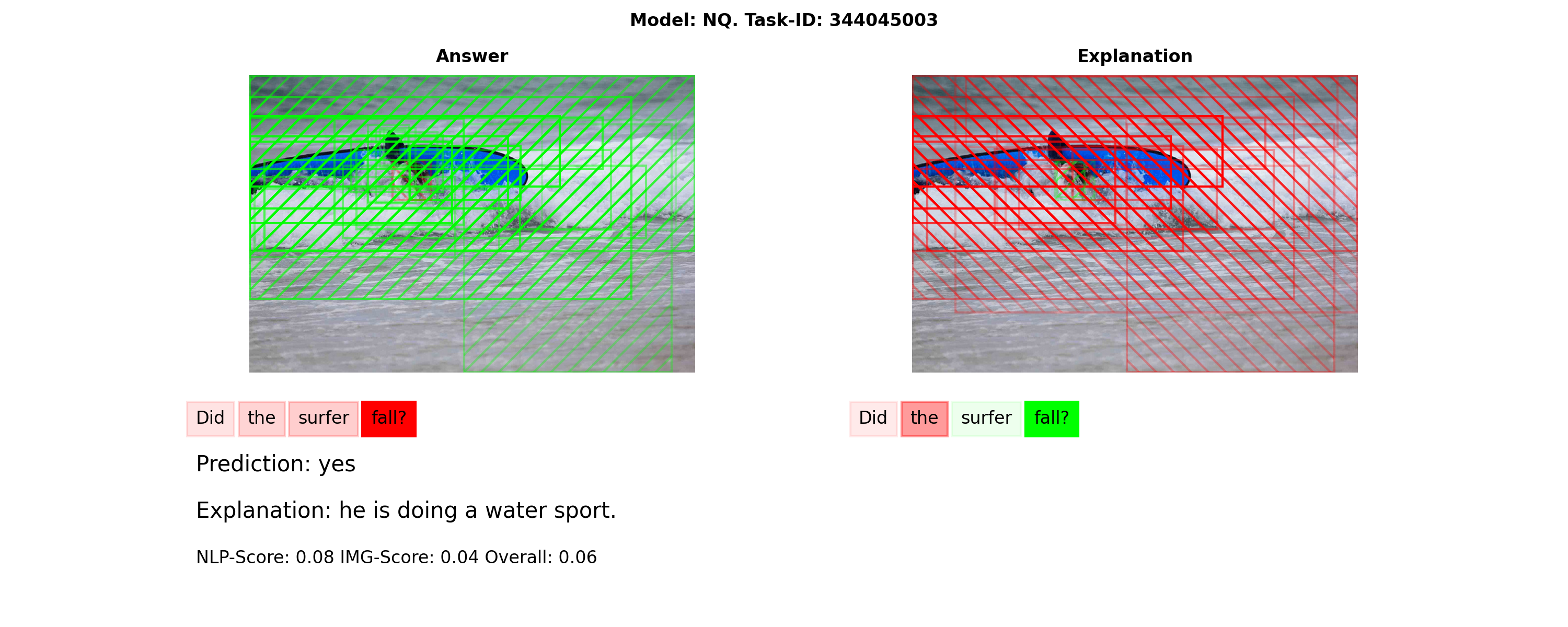}
    \includegraphics[width=\textwidth, trim={2.5cm 0.0cm 2.5cm 0.0cm}, clip]{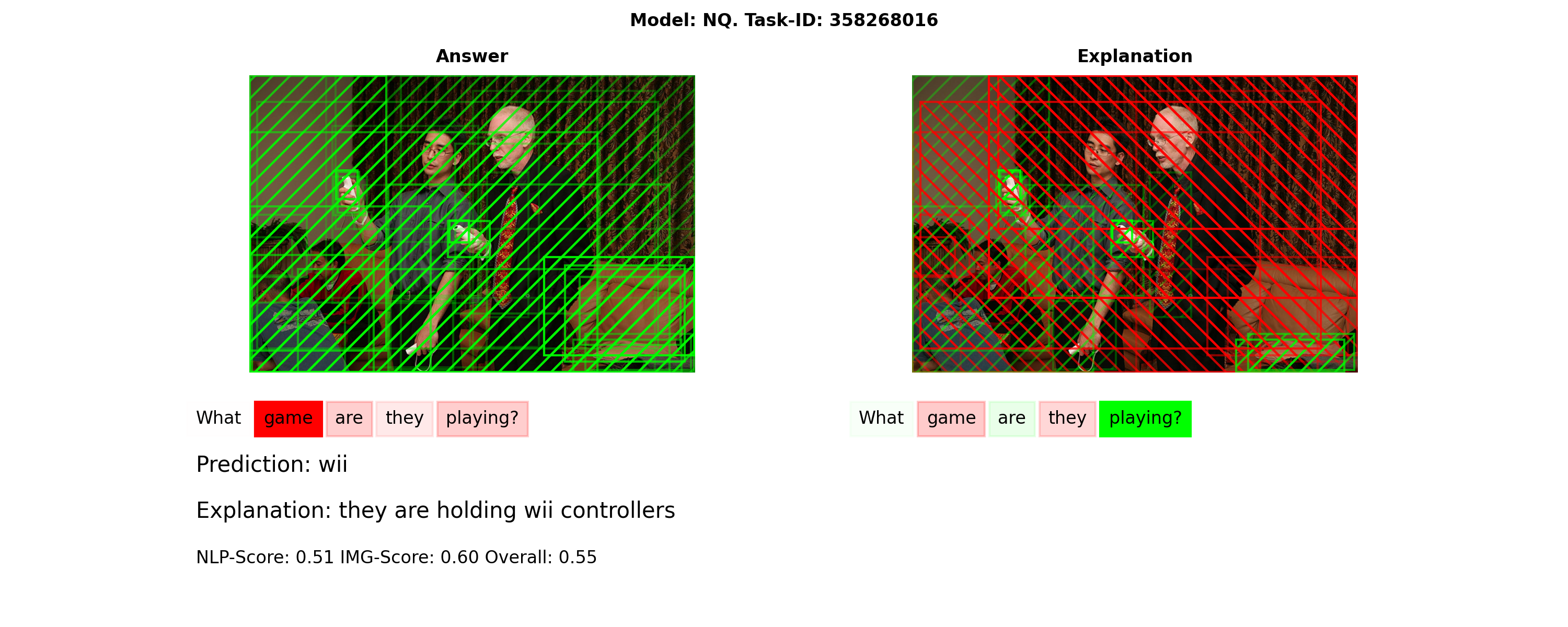}
    \caption{VQA-X: No-additional-question (NQ) examples (1/2)}
    \label{fig:my_label}
\end{figure}

\begin{figure}[h!!!!]
    \centering
    \includegraphics[width=\textwidth, trim={2.5cm 0.0cm 2.5cm 0.0cm}, clip]{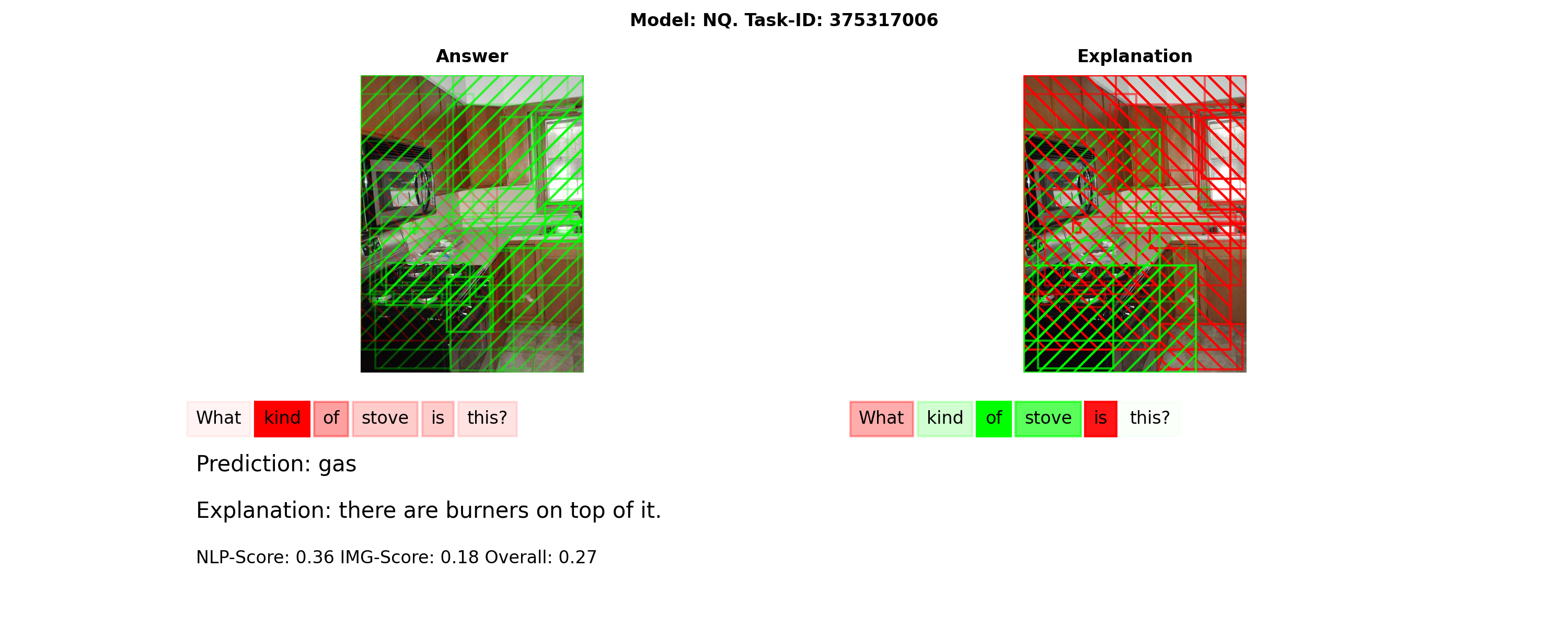}
    \includegraphics[width=\textwidth, trim={1cm 0.0cm 1cm 0.0cm}, clip]{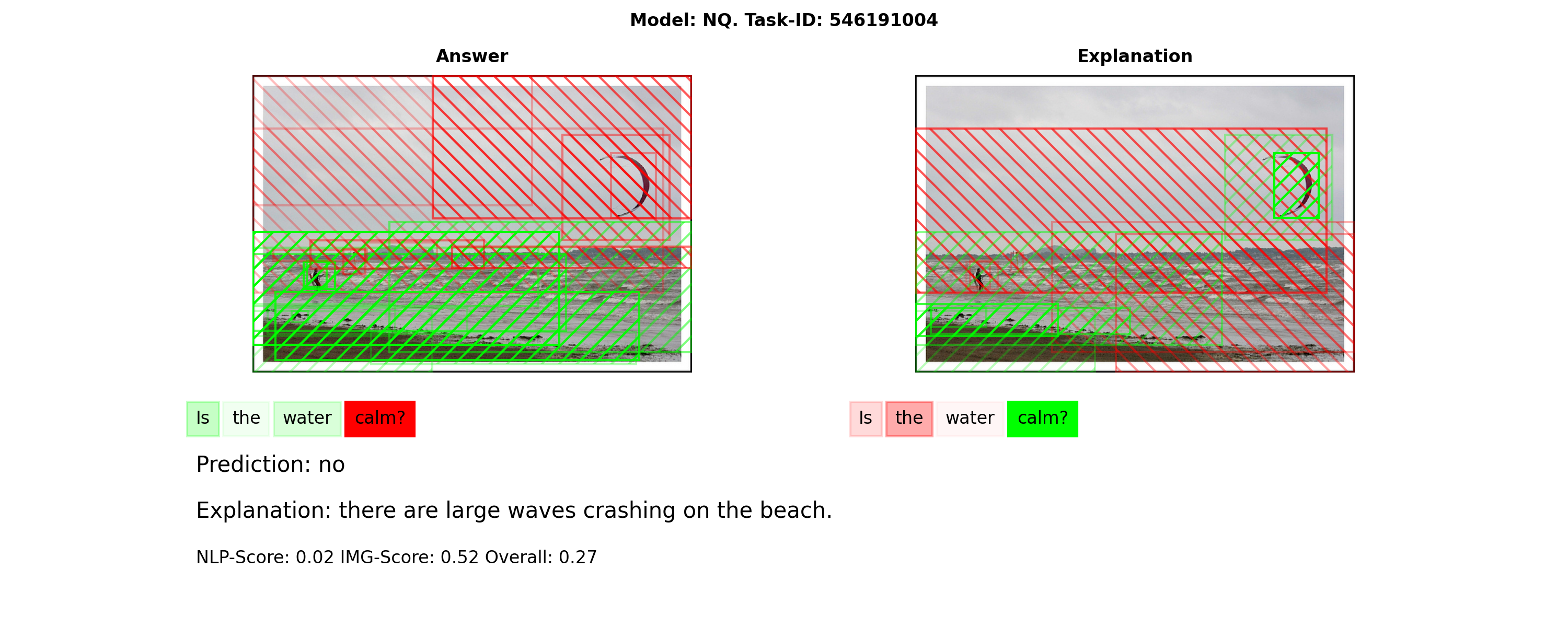}
    \caption{VQA-X: No-additional-question (NQ) examples (2/2)}
    \label{fig:my_label}
\end{figure}

\begin{figure}[h!!!!]
    \centering
    \includegraphics[width=\textwidth, trim={2.5cm 0.0cm 2.5cm 0.0cm}, clip]{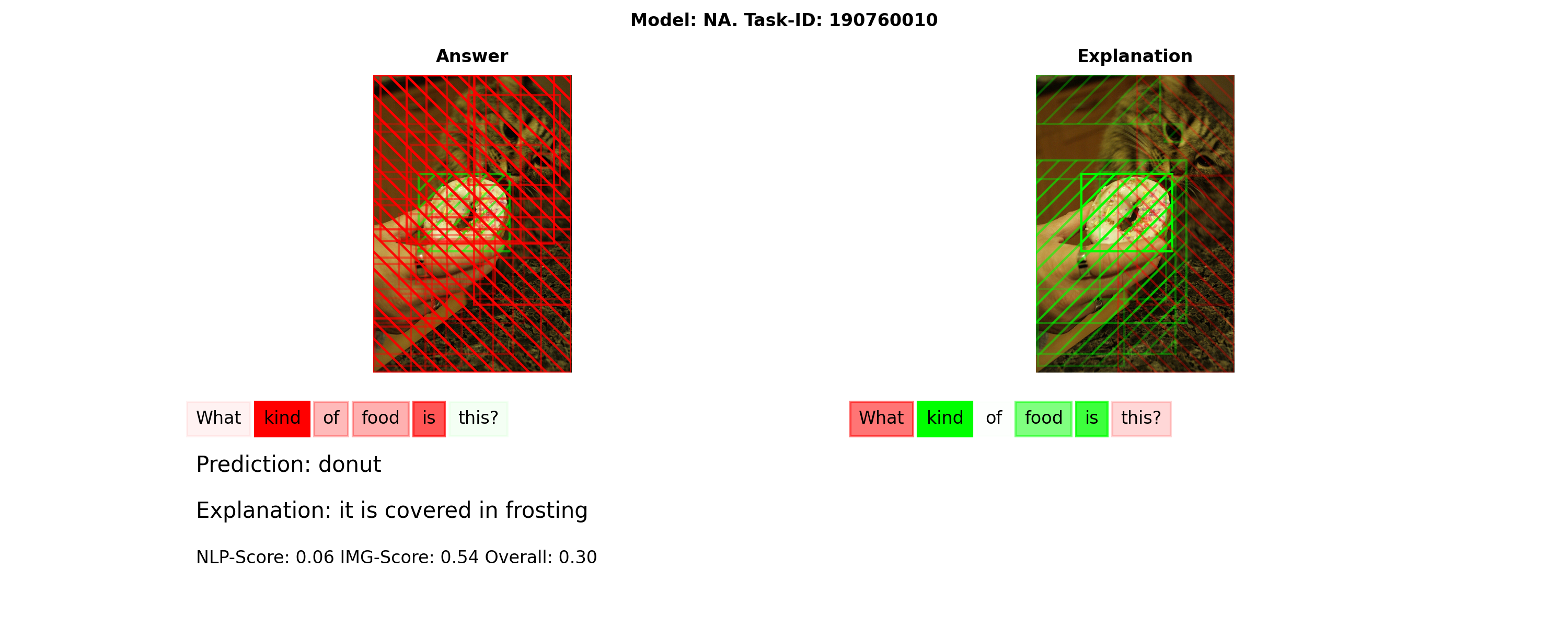}
    \includegraphics[width=\textwidth, trim={2.5cm 0.0cm 2.5cm 0.0cm}, clip]{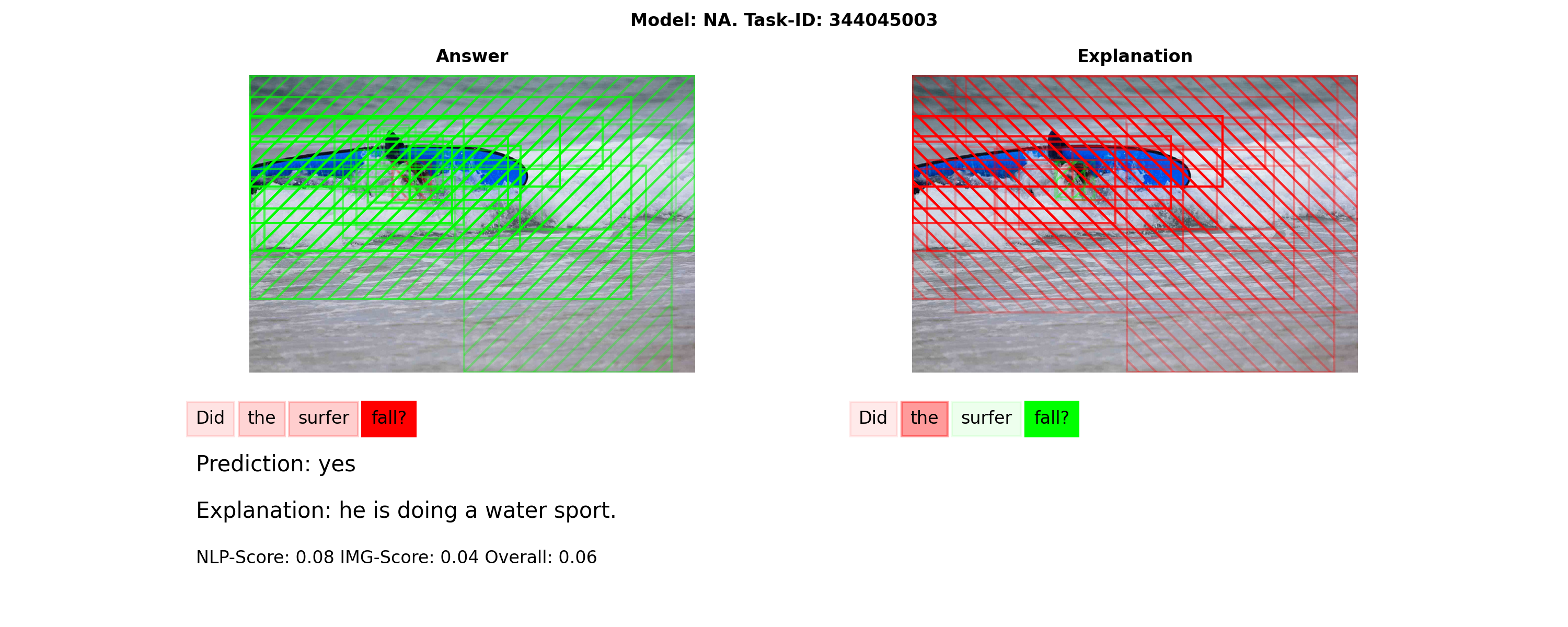}
    \includegraphics[width=\textwidth, trim={2.5cm 0.0cm 2.5cm 0.0cm}, clip]{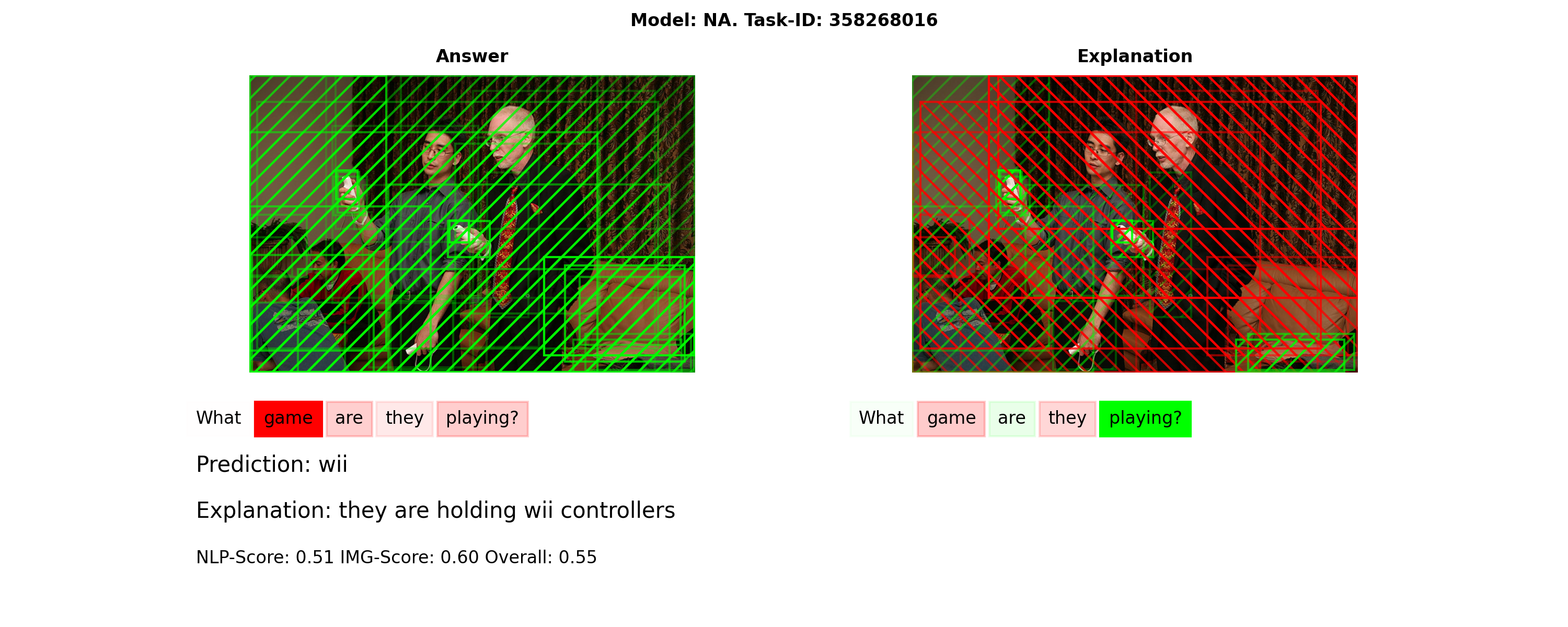}
    \caption{VQA-X: No-answer (NA) examples (1/2)}
    \label{fig:my_label}
\end{figure}

\begin{figure}[h!!!!]
    \centering
    \includegraphics[width=\textwidth, trim={2.5cm 0.0cm 2.5cm 0.0cm}, clip]{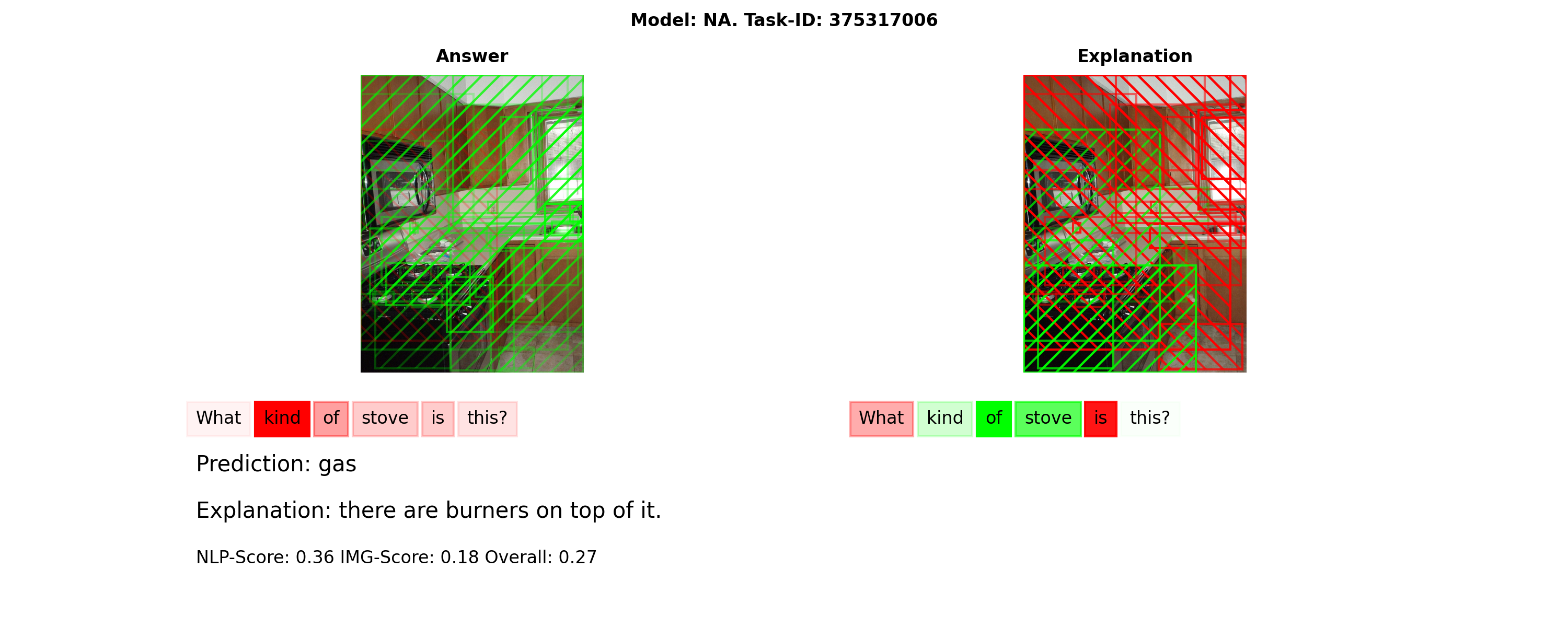}
    \includegraphics[width=\textwidth, trim={1cm 0.0cm 1cm 0.0cm}, clip]{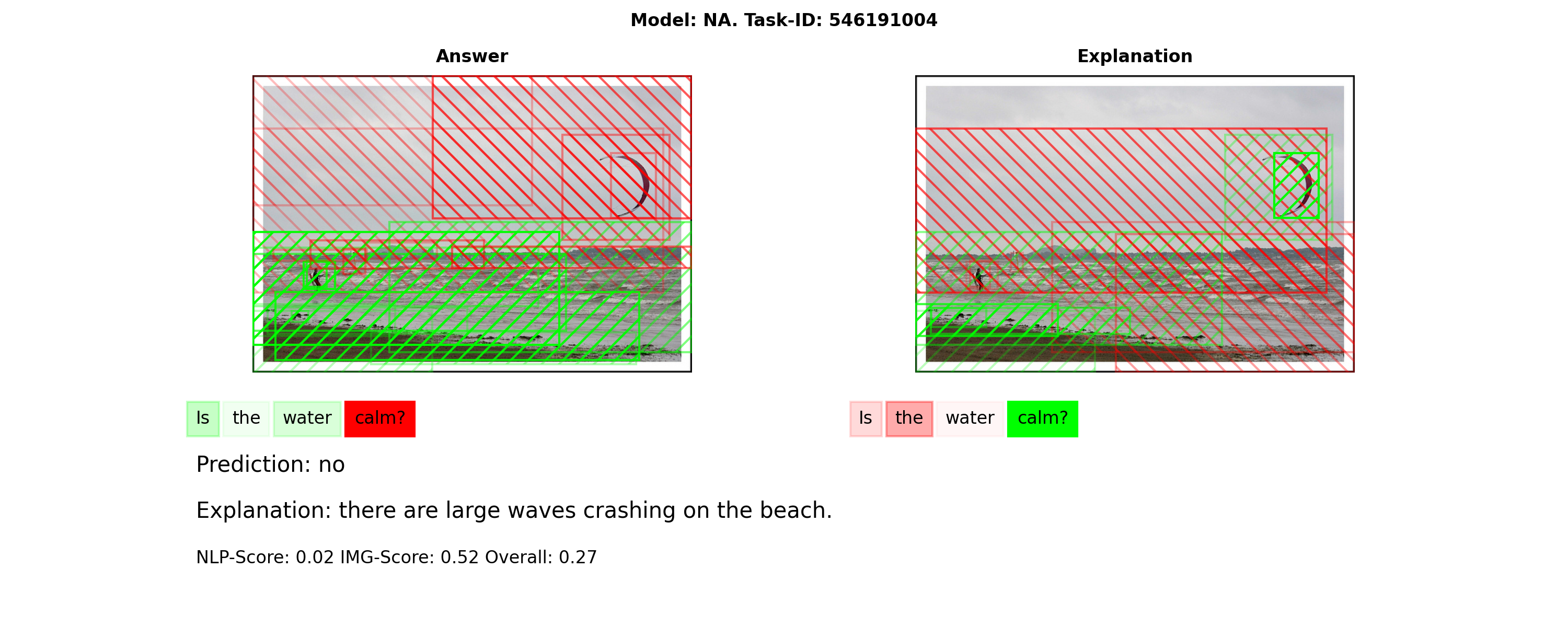}
    \caption{VQA-X: No-answer (NA) examples (2/2)}
    \label{fig:my_label}
\end{figure}

\begin{figure}[h!!!!]
    \centering
    \includegraphics[width=\textwidth, trim={2.5cm 0.0cm 2.5cm 0.0cm}, clip]{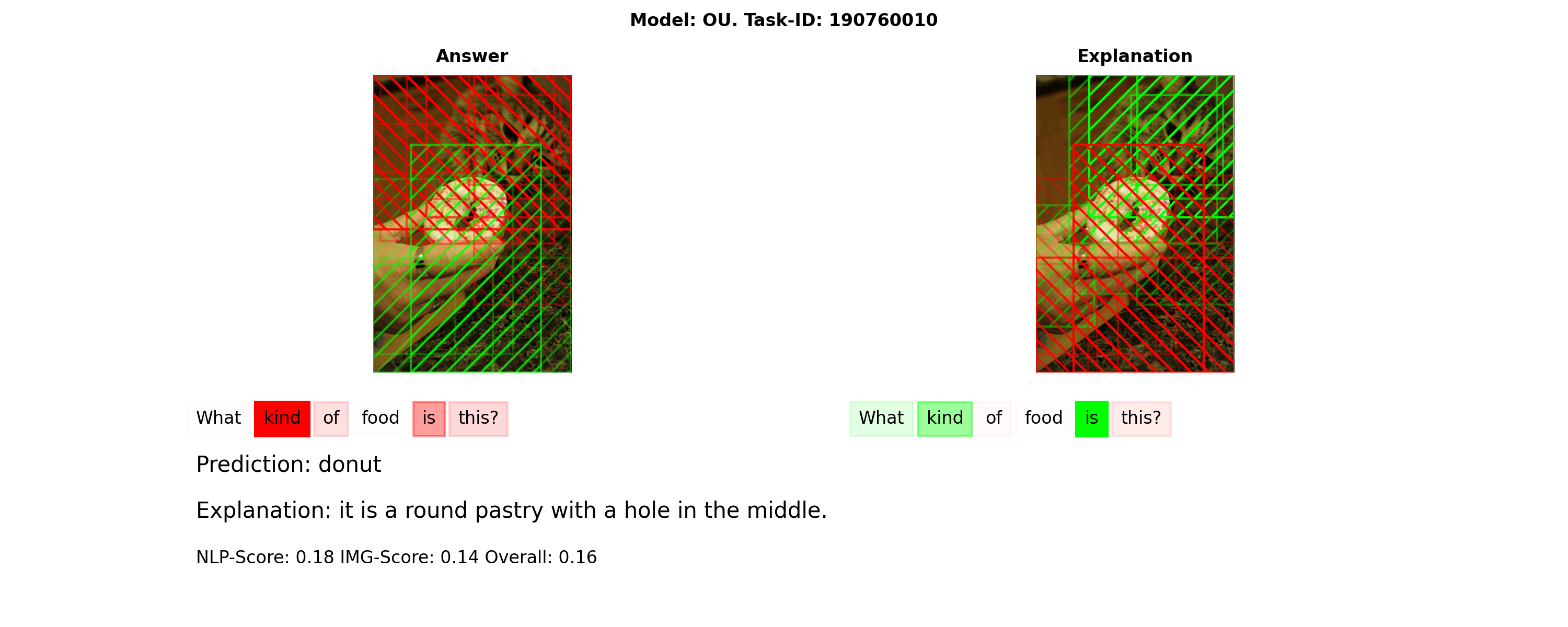}
    \includegraphics[width=\textwidth, trim={2.5cm 0.0cm 2.5cm 0.0cm}, clip]{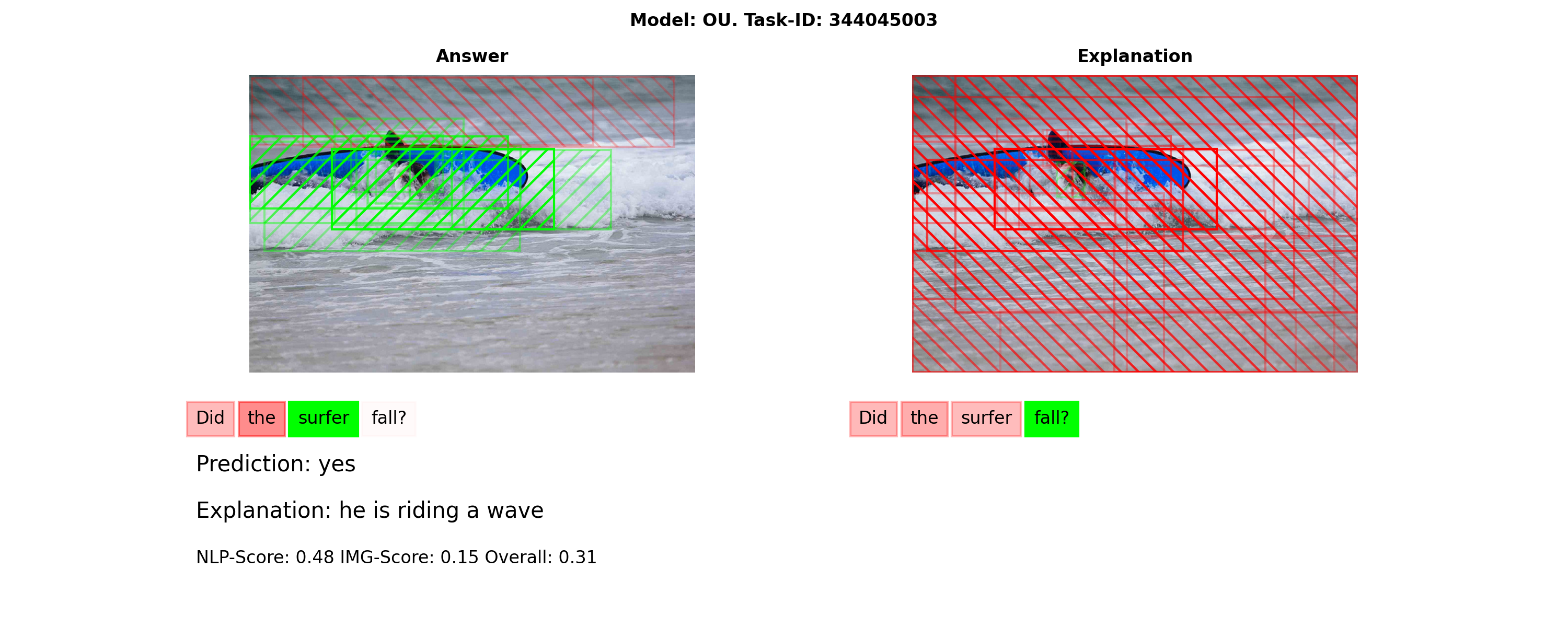}
    \includegraphics[width=\textwidth, trim={2.5cm 0.0cm 2.5cm 0.0cm}, clip]{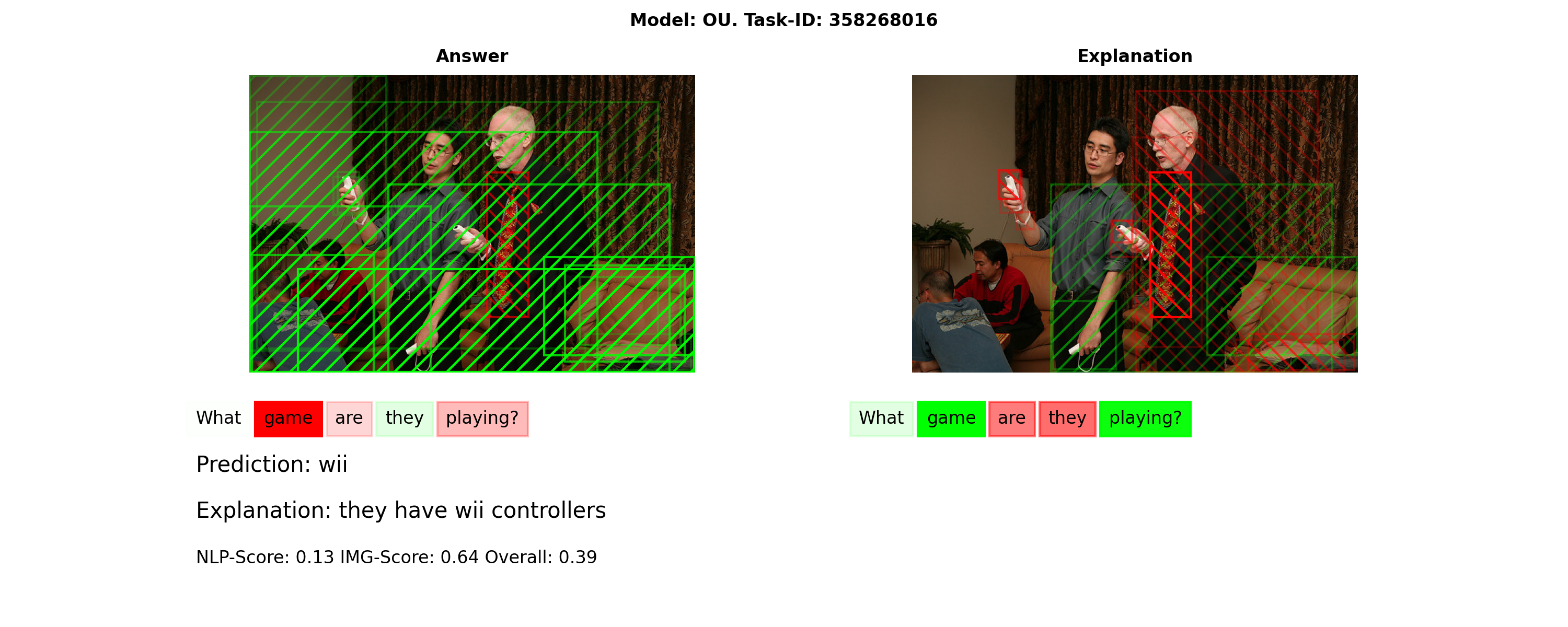}
    \caption{VQA-X: Only-Uniter (OU) examples (1/2)}
    \label{fig:my_label}
\end{figure}

\begin{figure}[h!!!!]
    \centering
    \includegraphics[width=\textwidth, trim={2.5cm 0.0cm 2.5cm 0.0cm}, clip]{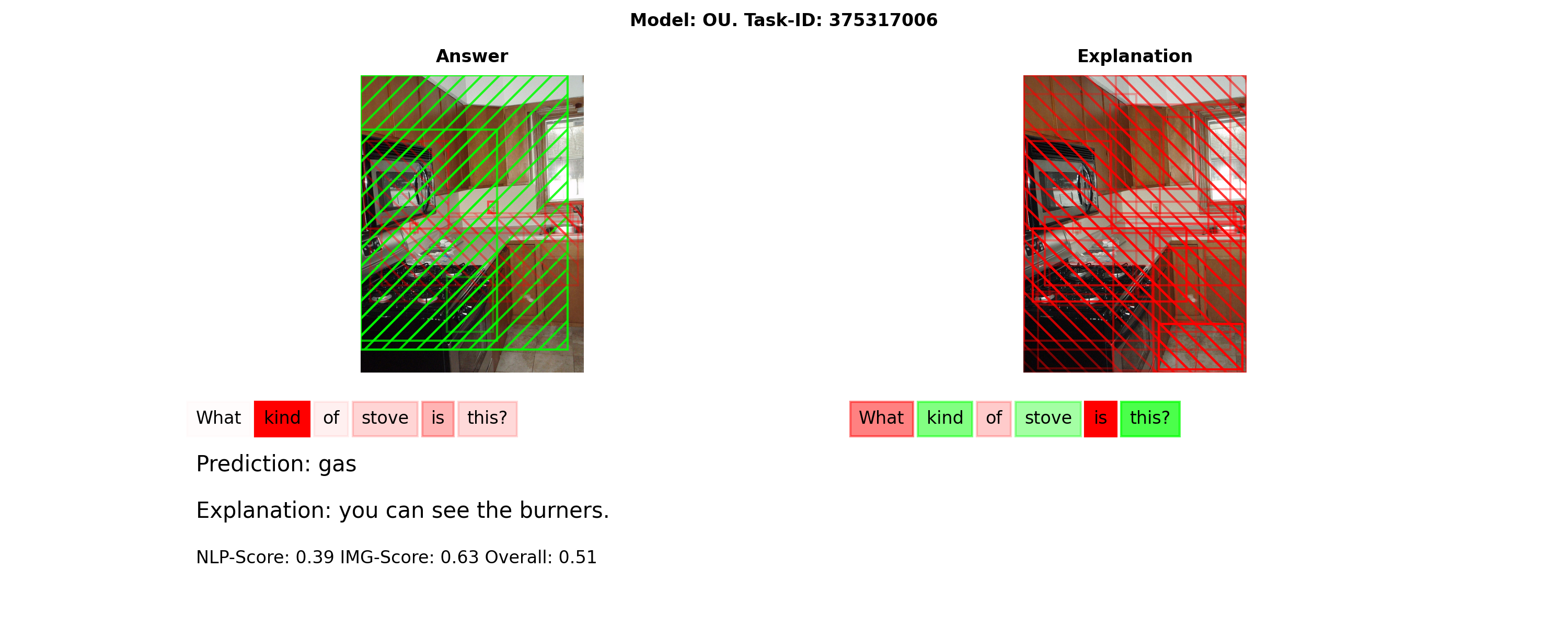}
    \includegraphics[width=\textwidth, trim={1cm 0.0cm 1cm 0.0cm}, clip]{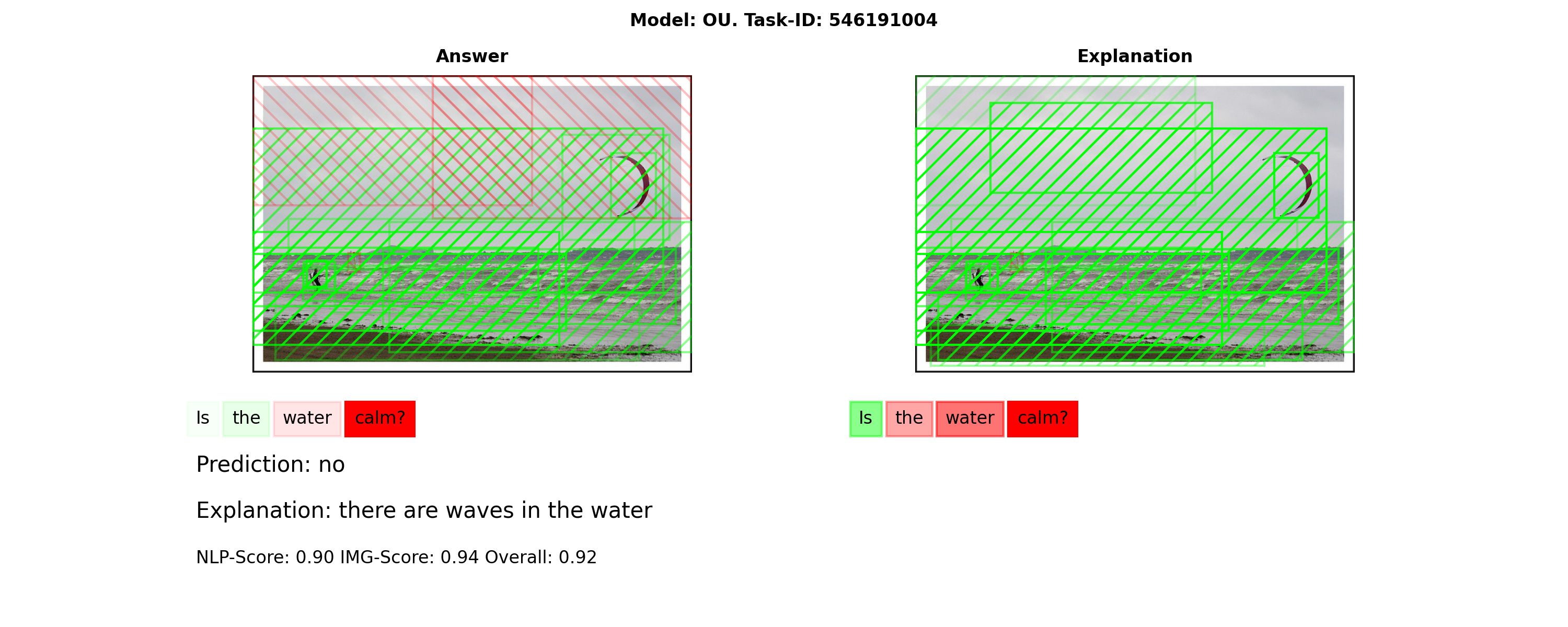}
    \caption{VQA-X: Only-Uniter (OU) examples (2/2)}
    \label{fig:my_label}
\end{figure}
\cleardoublepage

\fancyhead[LE,RO]{\it List of Figures}
\listoffigures
\cleardoublepage
\fancyhead[LE,RO]{\it List of Tables}
\listoftables
\cleardoublepage

\vspace{2cm}
\chapter*{Erkl\"arung der Urheberschaft}
\label{sec:urheber}
\fancyhead[LE]{\it Erkl\"arung der Urheberschaft}
Ich versichere an Eides statt, dass ich die \trtype{} im Studiengang \trcourseofstudies{} selbstst\"andig verfasst und keine anderen als die angegebenen Hilfsmittel -- insbesondere keine im Quellenverzeichnis nicht benannten Internet-Quellen -- benutzt habe. Alle Stellen, die w\"ortlich oder sinngem\"a{\ss} aus Ver\"offentlichungen entnommen wurden, sind als solche kenntlich gemacht. Ich versichere weiterhin, dass ich die Arbeit vorher nicht in einem anderen Pr\"ufungsverfahren eingereicht habe und die eingereichte schriftliche Fassung der auf dem elektronischen Speichermedium entspricht.

\vspace{4cm}
\noindent Ort, Datum \hfill Unterschrift

\newpage
\thispagestyle{empty}
\hspace{1cm}
\newpage

\vspace{2cm}
\chapter*{Erkl\"arung zur Ver\"offentlichung}
\label{sec:urheber}
\fancyhead[LE]{\it Erkl\"arung zur Ver\"offentlichung}
Ich erkl\"are mein Einverst\"andnis mit der Einstellung dieser \trtype{} in den Bestand der Bibliothek.

\vspace{4cm}
\noindent Ort, Datum \hfill Unterschrift

\newpage
\thispagestyle{empty}
\hspace{1cm}
\newpage

\end{document}